%% file: main.tex
\documentclass[mnsc]{informs3}
\OneAndAHalfSpacedXI
\usepackage[OT1]{fontenc}

\usepackage{amsthm}
\usepackage{microtype}

\usepackage{bm}
\usepackage{mathtools,booktabs,algpseudocode,placeins,subcaption}
\usepackage[ruled]{algorithm}
\usepackage{graphicx}
\usepackage{xcolor}
\usepackage{tikz}
\usetikzlibrary{arrows.meta,positioning,calc,fit}
\usepackage{pgfplots}
\pgfplotsset{compat=1.18}
\usepackage[authoryear,round]{natbib}
\bibpunct[, ]{(}{)}{,}{a}{}{,}
\def\bibfont{\small}

\usepackage[hidelinks,hypertexnames=false]{hyperref}
\providecommand{\doi}[1]{doi:\,\href{https://doi.org/#1}{\nolinkurl{#1}}}

\colorlet{red}{black}
\colorlet{blue}{black}
\definecolor{ORAccent}{RGB}{35,83,116}
\definecolor{ORAccentLight}{RGB}{232,239,244}
\definecolor{ORGray}{RGB}{92,92,92}
\definecolor{ORLightGray}{RGB}{238,238,238}

\allowdisplaybreaks
\newcommand{\preprintpageheader}{%
  \vbox{\hbox to \textwidth{\hspace{2.4pt}{\fontsize{10}{10}\selectfont\thepage}\hfil}%
    \vskip 4pt\hrule width \textwidth height 0.5pt}%
}
\let\setoddRH\preprintpageheader
\let\setevenRH\preprintpageheader

\theoremstyle{plain}
\newtheorem{theorem}{Theorem}
\newtheorem{lemma}{Lemma}
\newtheorem{proposition}{Proposition}
\newtheorem{corollary}{Corollary}
\newtheorem{assumption}{Assumption}
\theoremstyle{definition}
\newtheorem{definition}{Definition}
\theoremstyle{remark}
\newtheorem{remark}{Remark}
\renewcommand{\Halmos}{\qedhere}

\newcommand{\StartElectronicCompanion}{%
  \clearpage
  \setcounter{section}{0}
  \setcounter{subsection}{0}
  \renewcommand{\thesection}{\Alph{section}}
  \makeECnumber{equation}%
  \makeECnumber{figure}%
  \makeECnumber{table}%
  \makeECnumber{theorem}%
  \makeECnumber{lemma}%
  \makeECnumber{proposition}%
  \makeECnumber{corollary}%
  \makeECnumber{assumption}%
  \makeECnumber{definition}%
  \makeECnumber{remark}%
}
\newcommand{\makeECnumber}[1]{%
  \setcounter{#1}{0}%
  \expandafter\renewcommand\csname the#1\endcsname{EC.\arabic{#1}}%
}

\algrenewcommand\algorithmicrequire{\textbf{Input:}}
\algrenewcommand\algorithmicensure{\textbf{Output:}}

\newcommand{\R}{\mathbb{R}}
\newcommand{\E}{\mathbb{E}}
\newcommand{\Prob}{\mathbb{P}}

\newcommand{\thetastar}{\theta^\star}
\newcommand{\thetatilde}{\widetilde\theta}
\newcommand{\norm}[1]{\left\lVert #1 \right\rVert}
\newcommand{\abs}[1]{\left\lvert #1 \right\rvert}

\newcommand{\ip}[2]{\left\langle #1, #2 \right\rangle}
\newcommand{\Diag}{\mathrm{Diag}}
\newcommand{\diag}{\mathrm{diag}}

\newcommand{\KL}{\mathrm{KL}}

\newcommand{\Reg}{\mathrm{Reg}}

\newcommand{\Kcal}{\mathcal{K}}
\newcommand{\St}{\mathcal{S}}

\newcommand{\Ecal}{\mathcal{E}}
\newcommand{\Fcal}{\mathcal{F}}

\newcommand{\Bcal}{\mathcal{B}}

\newcommand{\Otilde}{\widetilde{O}}
\newcommand{\Omegatilde}{\widetilde{\Omega}}
\renewcommand{\argmax}{\mathop{\mathrm{arg\ max}}}
\renewcommand{\argmin}{\mathop{\mathrm{arg\ min}}}
\newcommand{\one}[1]{\mathbf{1}\!\left\{#1\right\}}

\hypersetup{
  pdftitle={The Revenue Cost of Inference in Online Assortment Optimization},
  pdfauthor={Jierui (Jerry) Zuo and Hanzhang Qin},
  pdfkeywords={assortment optimization, choice bandits, experimental design}
}
\begin{document}

\RUNAUTHOR{}
\RUNTITLE{The Revenue Cost of Inference in Online Assortment Optimization}
\TITLE{The Revenue Cost of Inference\\ in Online Assortment Optimization}
\ARTICLEAUTHORS{%
  \AUTHOR{Jierui (Jerry) Zuo}
  \AFF{Michael G. Foster School of Business\\ University of Washington}
  \AUTHOR{Hanzhang Qin}
  \AFF{Department of Industrial Systems Engineering and Management\\ National University of Singapore}
}
\ABSTRACT{%
A seller running an assortment experiment must earn revenue while collecting data to compare future offers. For a regular finite-dimensional parametric choice model and a finite assortment family, we characterize the oracle frontier between expected revenue loss and the largest local first-order variance constant across prespecified revenue comparisons. The oracle benchmark remains finite at zero cost exactly when revenue-maximizing assortments provide information in every target-gradient direction. When costly offers can supply the missing information, we derive the exact leading coefficient of the inverse-budget variance law and the limiting optimal spending shares. A pilot-based procedure reserves reference offers to preserve information and tightens the estimated budget constraint to account for uncertainty. We give conditions for the implemented design's variance criterion to converge to the stabilized oracle value at a fixed budget and reference share. Simulations with synthetic instances and Expedia-calibrated markets show that allocating offers across the full assortment family can improve the precision of revenue comparisons under a common opportunity-cost budget. Additional experiments assess pilot spending, budget protection, and sensitivity to choice-model misspecification.
}
\maketitle

\section{Introduction}\label{sec:intro}
\subsection{Overview of the problem}\label{subsec:intro-overview}
Online retailers must decide which products to display while learning how those choices affect revenue. At Alibaba, \citet{FeldmanEtAl2022} compared the platform's existing product-selection method with an assortment rule based on a multinomial logit (MNL) model. Their week-long randomized field experiment covered more than 10 million customer visits. When both methods used the same 25 features, the MNL-based rule generated RMB~5.17 per visit, compared with RMB~4.04 for the existing method. Such experiments make the revenue consequences of a display measurable. They also expose some customers to the less profitable option while the comparison is being made.

Hotel platforms face a similar need for precise comparisons. Expedia Group uses interleaving tests, which mix results from competing hotel rankings, to measure relative customer preference \citep{WoznicaEtAl2026}. Figure~\ref{fig:expedia-preference-workflow} summarizes how the company turns clicks and bookings into preference estimates with confidence intervals. These estimates help select candidates for A/B tests of business outcomes. Infrequent bookings make small improvements difficult to detect with limited traffic.

\begingroup
\setlength{\intextsep}{8pt plus 2pt minus 2pt}
\setlength{\textfloatsep}{10pt plus 2pt minus 2pt}
\begin{figure}[htbp]
\centering
\input{figures/expedia_preference_learning_tikz}
\caption{Learning customer preferences in Expedia Group's experimentation process. Rankings $A$ and $B$ are compared using customer choices. \textit{Source:} Authors' schematic redrawing based on Figure~4 and the screening workflow in \citet{WoznicaEtAl2026}.}
\label{fig:expedia-preference-workflow}
\end{figure}
\endgroup

These experiments inform future displays. We study expected revenue differences between specified assortments, using a choice model to connect customer preferences to revenue. We ask how to estimate these differences while limiting revenue lost during the experiment, with presentation held fixed.

Consider replacing a familiar product with an alternative that the retailer seldom displays. Current sales may reveal little about the alternative, so evaluating the change can require lower-revenue offers. Other comparisons can be learned from profitable displays already in use. We ask which comparisons need additional offers and how to allocate them.

We specify the comparisons, called target contrasts, before collecting data. The seller reserves experimental traffic and uses an existing choice bandit policy on the remaining arrivals. The design chooses how much traffic to reserve and which assortments to offer.

For regular finite-dimensional parametric choice models over a finite assortment family, we characterize an oracle frontier between revenue opportunity cost and local contrast precision. The allocation problem is a semidefinite program (SDP), and shared demand parameters allow offers to inform comparisons involving other assortments. We characterize when revenue-maximizing offers provide enough information for local first-order inference. When costly offers can supply the missing information, we identify the exact leading coefficient in the inverse-budget variance law. We also characterize how spending is optimally divided among these offers as the budget approaches zero.

We select a stabilized allocation under unknown demand using reference offers and a tightened cost constraint.

\subsection{Related Literature}
Choice models link the offered assortment to customer choices and expected revenue \citep{Train2009}. Assortment optimization has been studied under MNL demand \citep{RusmevichientongShenShmoys2010,SaureZeevi2013}, nested logit models \citep{DavisGallegoTopaloglu2014,ChenShiWangZhou2021}, and the Exponomial Choice model \citep{AlptekinogluSemple2016}. Neural choice models also capture dependence on the offered assortment \citep{WangCaiLiTalluri2023,AouadDesir2022,KoLi2023,WangLiTalluri2023}. In our design problem, the choice model determines both revenue and the information available for estimating the target contrasts.

Assortment bandit policies learn demand while minimizing cumulative regret. UCB and Thompson sampling methods for MNL bandits are developed in \cite{AgrawalAvadhanulaGoyalZeevi2016,AgrawalAvadhanulaGoyalZeevi2017,AgrawalAvadhanulaGoyalZeevi2019}, with broader context in \citet{Agrawal2019}. Further work sharpens regret guarantees \citep{WangChenZhou2018}, studies contextual MNL bandits \citep{OhIyengar2019,OhIyengar2021,PerivierGoyal2022,LeeOh2024}, and considers cascading assortments \citep{ChoiUdwaniOh2024}, general value functions \citep{ZhangLuo2024}, and nonlinear or neural utilities \citep{HwangKimOh2025}. We use an existing policy on the rounds outside the planned experiment and allocate the experimental offers for contrast precision.

Bandit experimental design studies how to earn rewards while collecting information for estimation or for identifying the best action. \citet{SimchiLeviWang2025} characterize the tradeoff between regret and the precision of estimated reward differences. Related work connects regret minimization with best-arm identification \citep{TanZhongCheung2021,QinRusso2024} and studies regret and estimation error in combinatorial bandits \citep{XieCaoXu2026}. Our main frontier fixes the demand instance and asks how much revenue must be forgone to estimate specified revenue contrasts. Their gradients select the information directions that matter. Theorems~\ref{thm:free-inference} and~\ref{thm:exact-frontier} characterize when revenue-maximizing assortments cover these directions and identify the exact small-budget variance coefficient and limiting optimal spending shares. Section~\ref{sec:tradeoff} gives a complementary worst-case rate analysis on an MNL benchmark.

For MNL models, \citet{LeeOh2026OptimalDesign} develop optimal design methods over combinatorial assortment families, using certified optimization and a polynomial-time surrogate, and apply them to best-assortment identification. Our criterion minimizes the largest local variance constant across prespecified revenue contrasts under an opportunity-cost constraint. We focus in particular on the singular zero-budget endpoint and characterize the exact cost of information missing from revenue-maximizing assortments.

Inference must account for adaptive data collection. \citet{HadadHirshbergZhanWagerAthey2021} develop policy-evaluation intervals, \citet{CookMishlerRamdas2024} study efficient inference under adaptive assignment, and \citet{GuoXu2026} examine misspecified contextual bandit models. We condition post-pilot inference on the pilot-selected schedule. Appendix~\ref{sec:extensions} gives direct inference under general stationary choice laws.

A single linearized contrast gives a classical design criterion \citep{Elfving1952}. Multiple-contrast design uses information-matrix methods \citep{KieferWolfowitz1959,KieferWolfowitz1960,Pukelsheim2006}, conic formulations \citep{Sagnol2011}, and integer allocation methods \citep{PukelsheimRieder1992}. \citet{MetelkinaPronzato2017} formulate a compound design criterion balancing information for parameter estimation against response-based regret in covariate-adaptive treatment allocation. Their framework allows shared parameters and a regret constraint, includes a fixed reference mixture, and discusses criteria for treatment differences. \citet{Pronzato2009} studies regularization of singular contrast-optimal designs. As the revenue-loss budget approaches zero, the allocation concentrates on revenue-maximizing assortments. When the targets require additional information, optimal spending is determined by information per unit of opportunity cost in the directions that revenue-maximizing assortments do not inform to first order. Theorem~\ref{thm:exact-frontier} identifies the exact leading variance coefficient and characterizes how optimal designs divide that spending across offers. For MNL demand, Corollary~\ref{cor:cost-information-rule} gives a rule for choosing which incumbent product to replace when designing an offer that isolates one missing product.

\subsection{Contributions}\label{subsec:intro-contributions}
We develop an assortment experimentation framework for regular finite-dimensional parametric choice models over a finite assortment family. The paper makes three contributions.

First, we characterize \emph{how much revenue a seller must forgo to estimate specified revenue differences precisely}. Precision is measured by a first-order approximation to the large-sample variance at a fixed demand parameter, using how revenue differences respond to small parameter changes. Theorem~\ref{thm:free-inference} identifies when revenue-maximizing assortments cover all information directions needed for the comparisons. When extra information is needed and available, Theorem~\ref{thm:exact-frontier} gives the exact inverse-budget variance coefficient and limiting optimal spending shares as the budget approaches zero. Spending depends on information per unit of forgone revenue in the missing directions, which can come from offers outside the target comparisons. For MNL demand with a unique revenue-maximizing assortment at capacity, Corollary~\ref{cor:cost-information-rule} gives a replacement rule for offers that isolate one missing product.

Second, we develop \emph{a way to implement the design when demand is unknown}. A pilot sample estimates demand. We reserve traffic for reference offers that jointly provide information in every parameter direction. A pilot certificate bounds cost-estimation error and determines how much to tighten the budget. When this certificate is valid and the tightened design is selected, Theorem~\ref{thm:end-to-end-overlay} bounds average post-pilot revenue loss by the budget plus a rounding allowance for integer offer counts. This allowance decreases inversely with post-pilot sample size. As pilot estimates approach the true demand parameters and tightening vanishes, the executed variance criterion approaches its oracle value under known demand, provided an allocation with the reserved reference share fits strictly within the budget. The budget and reference share stay fixed. Pilot cost enters total regret separately. Corollary~\ref{cor:end-to-end-inference} gives a Gaussian approximation for post-pilot estimates conditional on the pilot-selected schedule under bounded MNL demand.

Third, we characterize \emph{how increasing experimental traffic changes revenue loss and estimation accuracy under MNL demand as the horizon grows}. We combine MNL-UCB, a policy that learns demand while seeking revenue, with offers that display one product at a time and divide planned traffic evenly across products. Theorem~\ref{thm:mnloverlay} gives worst-case bounds on revenue regret and estimation error. On a separate two-product benchmark with one product per offer, Corollary~\ref{cor:frontier} gives matching upper and lower polynomial exponents. These exponents describe the powers of the horizon governing regret growth and error decay.

A three-target synthetic illustration isolates the role of the assortment family in joint precision. Oracle calculations in four Expedia-calibrated markets exhibit both free- and costly-inference regimes. Joint simulations compare one common design for four contrasts in each market with target-restricted designs and pooled single-target designs. Simulations of sixteen separately optimized revenue comparisons assess post-pilot precision at matched design opportunity cost, empirical pilot reuse, and sensitivity to inference regularization under a common budget ceiling. Pilot-size experiments assess total opportunity cost, certificate feasibility, and post-pilot precision.

\section{Online Assortment Experimentation}\label{sec:problem}
We consider a seller who offers an assortment at each round and observes the customer's choice. The seller seeks to earn revenue during the selling horizon and to estimate prespecified revenue contrasts from the choice data.

\subsection{Stationary Assortment-Choice Environment}\label{subsec:model}
There are $N$ products indexed by $[N]:=\{1,\ldots,N\}$ and a family $\St\subseteq2^{[N]}$ of feasible assortments. Let $K:=\max_{S\in\St}|S|$ denote the largest feasible assortment size. The selling horizon has $T$ rounds. At each round $t=1,\ldots,T$, the seller offers an assortment $S_t\in\St$ and observes the customer's choice $Y_t\in S_t\cup\{0\}$, where $0$ denotes the outside option or no purchase. The seller may use past observations and randomization when choosing $S_t$.

Let $\Fcal_t:=\sigma(S_1,Y_1,\ldots,S_t,Y_t)$ denote the information from offers and observed choices through round $t$. The information available after choosing $S_t$ and before observing $Y_t$ is $\Fcal_{t-1}^{+}:=\sigma(\Fcal_{t-1},S_t)$.

We assume conditional stationarity: there is a fixed, unknown choice law $p^\star$ such that, almost surely,
\begin{equation}
\Prob(Y_t=i\mid\Fcal_{t-1}^{+})
=
p^\star(i\mid S_t),
\qquad i\in S_t\cup\{0\}.
\label{eq:model}
\end{equation}
For a given assortment, the customer's choice probabilities remain the same at every round.

\subsection{Parametric Choice Models and the MNL Specialization}\label{subsec:parametric-mnl}
For the allocation problem and the likelihood analysis, the stationary choice law belongs to a differentiable finite-dimensional family $\{p_\theta:\theta\in\Theta\subseteq\mathbb R^d\}$. The true parameter $\thetastar\in\Theta$ satisfies $p^\star(\cdot\mid S)=p_{\thetastar}(\cdot\mid S)$ for every $S\in\St$. An assortment is informative when its observed choices help distinguish nearby demand parameters. Its Fisher information matrix $\mathcal I_S(\theta)$ measures this ability in different parameter directions. A target contrast's gradient describes how the revenue difference changes in those same directions. Together, these objects determine the local precision of an allocation.

Our primary specialization is MNL demand. Each product
$i\in[N]$ has attraction $v_i>0$, the outside option is normalized to
$v_0=1$, and, for an offered set $S$,
\begin{equation*}
p_\theta(i\mid S)=\frac{v_i}{1+\sum_{j\in S}v_j},\quad i\in S,
\qquad
p_\theta(0\mid S)=\frac{1}{1+\sum_{j\in S}v_j}.
\end{equation*}
We use the log-attraction parameterization $\theta_i:=\log v_i$, so
$v_i=e^{\theta_i}$. The attraction $v_i$ measures product $i$'s appeal relative to no purchase. The log-attraction expresses this same quantity on a logarithmic scale.

For $y\in S\cup\{0\}$, let
$\ell(\theta;y,S):=-\log p_\theta(y\mid S)$.
Let $\theta_S=(\theta_i)_{i\in S}$ denote the parameters of the offered
products. With $p_S=(p_\theta(i\mid S))_{i\in S}$, we have
\begin{align*}
\frac{\partial}{\partial\theta_i}\log p_\theta(y\mid S)
&=\one{y=i}-p_\theta(i\mid S),\qquad i\in S,\\
\nabla_{\theta_S}^2\ell(\theta;y,S)
&=\diag(p_S)-p_Sp_S^\top.
\end{align*}
The Hessian does not depend on the observed choice. We obtain the
$N\times N$ Fisher information matrix $\mathcal I_S(\theta)$ by placing
this Hessian on the rows and columns indexed by $S$ and setting all
remaining entries to zero. The design in Section~\ref{subsec:contrast-design}
uses this matrix.

\begin{assumption}[Bounded attractions]\label{ass:mnlbounded}
There exist constants $0<v_{\min}\le v_{\max}<\infty$ such that
$v_i\in[v_{\min},v_{\max}]$ for every item. We take the parameter space to be
\[
\Theta=[\log v_{\min},\log v_{\max}]^N.
\]
\end{assumption}
These bounds limit how rare or dominant a product can be within the fixed catalog. They provide the probability and information bounds used in the uniform guarantees below.

\subsection{Revenue, Regret, and Inferential Targets}\label{subsec:revenue-regret}
Each product $i\in[N]$ has known normalized revenue $r_i\in[0,1]$,
and $r_0:=0$. Under the stationary choice law, the expected revenue
from offering assortment $S$ to one customer is
$R^\star(S):=\sum_{i\in S}r_i p^\star(i\mid S)$.
In the parametric model, the expected revenue at parameter $\theta$ is
$R_\theta(S):=\sum_{i\in S}r_i p_\theta(i\mid S)$.
Let $S_\theta^\star$ denote a feasible assortment that maximizes
this expected revenue, so
$S_\theta^\star\in\argmax_{S\in\St}R_\theta(S)$.
When $p^\star=p_{\thetastar}$, we have
$R^\star(S)=R_{\thetastar}(S)$.
Under MNL,
\begin{equation*}
R_\theta(S)=\frac{\sum_{i\in S}r_i e^{\theta_i}}
{1+\sum_{j\in S}e^{\theta_j}}
\in[0,1].
\end{equation*}
For a policy $\pi$, cumulative regret at horizon $T$ is
\begin{equation*}
\Reg_\pi(T,\thetastar):=\sum_{t=1}^T\{R_{\thetastar}(S^\star_{\thetastar})-R_{\thetastar}(S_t)\}.
\end{equation*}
This quantity is the cumulative pseudo-regret. Each term compares the expected revenue of the best assortment with that of the offered assortment. The total can still be random because the policy uses past observations and may randomize its offers.

\paragraph{Inferential targets.}\label{subsec:targets}
Fix a finite contrast index set $\Kcal$. Each index $k=(a,b)\in\Kcal$
specifies two assortments $S_a,S_b\in\St$ whose expected revenues will be
compared. Under a stationary law, the model-free target is
$\Delta^{(a,b)}:=R^\star(S_a)-R^\star(S_b)$.
Under the parametric specialization, the same target at the true parameter is
$\Delta_R^{(a,b)}:=R_{\thetastar}(S_a)-R_{\thetastar}(S_b)$.
When the data-generating parameter is a generic $\theta$, the same contrast is denoted $\Delta_{R,\theta}^{(a,b)}:=R_\theta(S_a)-R_\theta(S_b)$.

Every target contrast lies in $[-1,1]$. We take contrast estimates to lie
in this interval, since projection onto it cannot increase absolute error.

\subsection{Planned Traffic and the Two Design Levers}\label{subsec:joint-objective}
Before the horizon begins, the seller reserves a set $I_T\subseteq[T]$ of
$n_T$ rounds for experimental offers. We call these planned rounds. An online
assortment policy, called the base policy, selects assortments on all remaining
rounds to earn revenue.

At parameter $\theta$, offering assortment $S$ instead of a revenue-maximizing
assortment incurs the expected revenue loss
$d_S(\theta):=R_\theta(S_\theta^\star)-R_\theta(S)$.
We refer to this loss as the opportunity cost of the offer. The experimental
allocation determines which assortments are offered on planned rounds. For a
fixed allocation, a budget limits its average opportunity cost per offer.
When a pilot is used to select the allocation, this budget applies to the
post-pilot offers. The overall regret bound also includes the pilot cost.

\paragraph{Allocation versus scale.}\label{subsec:two-levers}
The allocation and the number of offers jointly determine the cost and precision
of the experiment. Let $q=(q_S)_{S\in\St}$ denote the allocation, where
$q_S$ is the fraction of experimental offers assigned to assortment $S$.
These fractions satisfy $q_S\geq0$ and $\sum_{S\in\St}q_S=1$. At a fixed
parameter $\theta$, the average opportunity cost per offer is
$C_\theta(q):=\sum_{S\in\St}q_Sd_S(\theta)$.
Section~\ref{sec:overlay} defines $V_\theta(q)$, the largest first-order variance
constant across the target contrasts. For $m$ offers allocated according to $q$,
the total expected opportunity cost is $mC_\theta(q)$. The first-order standard
error benchmark for the least precise target contrast is
$\sqrt{V_\theta(q)/m}$. Here $m$ denotes a generic sample size, whereas $n_T$
counts all planned rounds, including any pilot observations.

Section~\ref{sec:overlay} chooses an allocation under a given opportunity-cost
budget. Section~\ref{sec:implementation} uses pilot data to learn and implement
that allocation when the choice model parameters are unknown.
Section~\ref{sec:tradeoff} establishes worst-case regret and estimation-error
rates as the number of planned rounds varies.
\long\def\RateWiseCriteria{%
\subsection{Rate-Wise Criteria}\label{subsec:rate-criteria}
Fix an instance class $\Theta_0\subseteq\Theta$. For a policy $\pi$ and contrast estimator $\widehat\Delta$, define the worst-case objectives
\begin{align*}
\mathcal R_T(\pi)&:=\sup_{\theta\in\Theta_0}\E_\theta[\Reg_\pi(T,\theta)],\\
\mathcal E_T(\widehat\Delta)&:=\sup_{\theta\in\Theta_0}\max_{(a,b)\in\Kcal}\E_\theta\abs{\widehat\Delta^{(a,b)}-\Delta_{R,\theta}^{(a,b)}}.
\end{align*}
The first criterion measures expected cumulative regret at the worst instance
in $\Theta_0$. The second measures the largest expected absolute error among
the target contrasts, again at the worst instance.

We compare how regret grows and estimation error decreases with the selling
horizon, ignoring logarithmic factors and holding problem parameters fixed.
Within a fixed base-policy class, instance class, and target family, a schedule
is rate-wise nondominated if no other schedule improves one of these polynomial
rates without worsening the other. Here $\Otilde(\cdot)$ suppresses logarithmic
factors in $T$, $N$, and $\delta^{-1}$.
}%

\section{The Oracle Frontier between Revenue and Precision}\label{sec:overlay}
At a fixed demand parameter, we minimize the largest local first-order variance constant across the target contrasts under an opportunity-cost budget. Varying the budget defines the oracle frontier for a regular finite-dimensional parametric choice model over a finite assortment family. Section~\ref{sec:implementation} develops a pilot-based implementation under unknown demand.

\long\def\BlackBoxOverlaySection{%
\subsection{Black-Box Overlay Architecture}\label{subsec:blackbox-overlay}
We use a base choice bandit policy $\Bcal$ to select assortments on the $T-n_T$ base-policy rounds. A separate experimental design rule selects the assortments on the $n_T$ planned rounds. The base policy uses only observations from its preceding base-policy rounds. Its round counter increases by one on each of these rounds. A planned round leaves both its counter and its observation history unchanged. Keeping the histories separate lets us use the base policy's existing regret guarantee and analyze the experimental design on its own. In the pilot-based design, we estimate the contrasts from the post-pilot sample.

Conditional stationarity in \eqref{eq:model} lets the base policy's regret guarantee carry over to its own rounds. The next theorem initially charges each planned round the maximum possible loss. The budgeted design will refine this charge for post-pilot offers in Section~\ref{subsec:end-to-end-guarantees}. Figure~\ref{fig:pilot-safe-pipeline} shows the implementation, and Algorithm~\ref{alg:overlay} in Appendix~\ref{app:overlay-algorithm} gives the general procedure.

\FloatBarrier
\begingroup
\setlength{\intextsep}{8pt plus 2pt minus 2pt}
\setlength{\textfloatsep}{10pt plus 2pt minus 2pt}
\begin{figure}[!htbp]
\centering
\input{figures/pilot_safe_pipeline_tikz}
\caption{Two-track implementation. Base-policy and planned rounds are
interleaved, with separate observation histories. The base policy updates
only from its own observations. Pilot observations are used to tighten the
opportunity cost budget and select a stabilized allocation, which is rounded
before deployment. Contrast inference uses only post-pilot observations
and conditions on the pilot-selected schedule.}
\label{fig:pilot-safe-pipeline}
\end{figure}
\endgroup

For an instance $\theta$, let $B_{\Bcal}(m,\theta)$ be an upper bound on the expected pseudo-regret of the base policy after $m$ of its own rounds.

Let $\pi_{\rm overlay}$ denote the policy induced by
Algorithm~\ref{alg:overlay}, and write
$\Reg_{\rm overlay}(T,\theta)
:=
\Reg_{\pi_{\rm overlay}}(T,\theta)$.
When a high-probability guarantee is used,
$B_{\Bcal}^{\rm hp}(m,\theta;\delta)$ denotes a pseudo-regret bound
that holds with probability at least $1-\delta$.

\begin{theorem}[Black-box overlay guarantee]\label{thm:overlay}
Suppose rewards are normalized to $[0,1]$ and the conditional-stationarity condition in \eqref{eq:model} holds. For any deterministic set of planned rounds $I_T$ and any planned design rule, Algorithm~\ref{alg:overlay} satisfies
\begin{equation*}
\E_\theta[\Reg_{\rm overlay}(T,\theta)]
\le n_T+B_{\Bcal}(T-n_T,\theta).
\end{equation*}
If the base-policy guarantee holds with probability at least $1-\delta$, the corresponding overlay bound holds with the same probability after adding the deterministic term $n_T$.
\end{theorem}

\begin{proof}
Every planned round contributes at most one unit of pseudo-regret. At each base-policy round, conditional stationarity in \eqref{eq:model} gives the customer's choice the same conditional distribution as in an uninterrupted run of $\Bcal$. The base policy also selects its assortment using only its own preceding observations. Its trajectory over the $T-n_T$ base-policy rounds therefore has the same distribution as an uninterrupted run of that length under the same environment. Adding the regret from the planned rounds gives the stated bound. The same decomposition applies on the base policy's high-probability guarantee event.
\end{proof}

The bound separates base-policy performance from the choice of experimental offers. We next use the opportunity-cost budget to choose those offers.
}%

\subsection{Revenue Cost and Statistical Uncertainty}\label{subsec:contrast-design}
Each allocation trades off revenue opportunity cost and the precision of the target comparisons. We evaluate these quantities at the same parameter and first describe how the target revenue differences respond to small parameter changes.

Throughout this subsection, $\St$ and $\Kcal$ are finite.
For $k=(a,b)\in\Kcal$, write
\[
\Delta_k(\theta):=\Delta_{R,\theta}^{(a,b)},
\qquad
g_k(\theta):=\nabla_\theta\Delta_k(\theta).
\]
The gradient $g_k(\theta)$ describes how the $k$th revenue difference changes with the demand parameters. Collect the target contrasts in the vector
$\boldsymbol{\Delta}(\theta):=(\Delta_k(\theta))_{k\in\Kcal}$.
For an assortment $S$, let $\mathcal I_S(\theta)$ be
the Fisher information generated by offering $S$, and define
$M(q,\theta):=\sum_{S\in\St}q_S\mathcal I_S(\theta)$.
Let $\Delta(\St)$ denote the set of nonnegative allocations whose components
sum to one. For $q\in\Delta(\St)$, $M(q,\theta)$ is the expected Fisher
information from one experimental offer. We use the opportunity costs
$d_S(\theta)$ and $C_\theta(q)$ defined in Section~\ref{subsec:joint-objective}.

For the log-attraction MNL parameterization defined in
Section~\ref{subsec:parametric-mnl}, we use $p_S(\theta)$ for the
$N$-dimensional choice probability vector with entries $p_\theta(i\mid S)$
for $i\in S$ and zeros outside $S$. Then
\begin{equation}
\mathcal I_S(\theta)
=\Diag\bigl(p_S(\theta)\bigr)-p_S(\theta)p_S(\theta)^\top,
\qquad
\{\nabla_\theta R_\theta(S)\}_i
=\one{i\in S}p_\theta(i\mid S)\{r_i-R_\theta(S)\}.
\label{eq:mnl-design-inputs}
\end{equation}
Equation~\eqref{eq:mnl-design-inputs} gives the information and
revenue-gradient inputs to the design program. We compare allocations using
the largest first-order variance constant across the target contrasts. For
allocations whose information range contains every target gradient, define
\begin{equation*}
V_\theta(q):=
\max_{k\in\Kcal}g_k(\theta)^\top M(q,\theta)^\dagger g_k(\theta).
\end{equation*}
Here $M^\dagger$ denotes the Moore--Penrose pseudoinverse, which computes the inverse on the directions with information. Each quadratic form is a local variance constant for one contrast, and $V_\theta(q)$ takes the largest. For $m$ experimental offers, $V_\theta(q)/m$ is the variance benchmark for the least precise contrast. It uses a local approximation in which sampling error decreases with the inverse square root of sample size. If a target gradient lies outside $\operatorname{Range}\{M(q,\theta)\}$, we set $V_\theta(q)=+\infty$: the allocation leaves a relevant first-order direction uncovered. For example, under MNL demand, choices from $\{1,2\}$ reveal nothing about the attraction of product 3. A comparison that changes with that attraction requires additional offers. The three-product example below works through this information requirement.

The next subsection finds the smallest variance constant attainable under each opportunity-cost budget.

\subsection{Frontier Value, Shape, and Shadow Price}
At a fixed model parameter $\bar\theta$, we choose an allocation under an
opportunity cost budget $\eta\ge0$ per experimental offer. We minimize a
scalar upper bound $\tau$ on the local variance constants of all target
contrasts:
\begin{equation}
\begin{aligned}
F_{\bar\theta}(\eta):=\min_{q,\tau}\quad &\tau\\
\text{s.t.}\quad
&q\ge0,\qquad \sum_{S\in\St}q_S=1,
\qquad \sum_{S\in\St}q_Sd_S(\bar\theta)\le\eta,\\
&\begin{bmatrix}
M(q,\bar\theta)&g_k(\bar\theta)\\
g_k(\bar\theta)^\top&\tau
\end{bmatrix}\succeq0,
\qquad k\in\Kcal.
\end{aligned}
\label{eq:c-optimal-sdp}
\end{equation}
The first two constraints make $q$ a valid allocation of experimental offers.
The cost constraint limits its average opportunity cost to $\eta$. By the
generalized Schur complement, each matrix constraint ensures that the target
gradient lies in the range of $M(q,\bar\theta)$ and that its local variance
constant is at most $\tau$. The optimal value $F_{\bar\theta}(\eta)$ is
therefore the lowest possible value of $V_{\bar\theta}(q)$ under this budget.
Let $\mathcal Q^\star(\bar\theta,\eta)$ denote the set of optimal
allocations, and let $q^\star(\bar\theta,\eta)$ be any member when this set
is nonempty.

Proposition~\ref{prop:design-sdp} in Appendix~\ref{app:sdp-representation}
gives the formal equivalence.

\begin{proposition}[Shape and shadow price of the oracle frontier]\label{prop:frontier-shape}
For fixed $\theta$, $F_\theta(\eta)$ is nonincreasing and convex in
$\eta$. Under Slater's condition, if $\lambda^\star(\eta)$ is an optimal
multiplier on the budget constraint, then
$-\lambda^\star(\eta)\in\partial F_\theta(\eta)$.
Here $\partial F_\theta(\eta)$ denotes the subdifferential
of $F_\theta$ at $\eta$.
\end{proposition}

\begin{proof}
See Corollary~\ref{cor:stabilized-shadow-price} and
Appendix~\ref{app:equivalence-pricing}.
\end{proof}

A larger budget can improve precision by allowing allocations with lower
variance constants. Convexity means that the marginal reduction in the optimal
variance constant does not increase as the budget grows. At a differentiable
point, $\lambda^\star(\eta)$ gives this reduction per unit of additional
budget. This is the shadow price of the revenue budget.

\subsection{Free and Costly First-Order Inference}
At zero opportunity cost, an allocation can use only revenue-maximizing assortments. First-order inference is free when these offers supply information for every target comparison. The pilot and reference costs of implementing a design under unknown demand are addressed in the next section.

\begin{samepage}
\begin{theorem}[Free-inference span characterization]\label{thm:free-inference}
Fix $\theta$, let
\[
\mathcal O(\theta):=\{S\in\St:d_S(\theta)=0\},
\qquad
M_{\mathcal O}(\theta):=\sum_{S\in\mathcal O(\theta)}\mathcal I_S(\theta).
\]
Then
\begin{equation}
F_\theta(0)<\infty
\quad\Longleftrightarrow\quad
g_k(\theta)\in\operatorname{Range}\{M_{\mathcal O}(\theta)\}
\quad\text{for every }k\in\Kcal.
\label{eq:zero-cost-characterization}
\end{equation}
Equivalently, a distribution supported on revenue-maximizing assortments yields
a finite first-order variance benchmark for every target contrast.
\end{theorem}
\end{samepage}

The span condition concerns precision near the fixed parameter. To identify the contrast from the observed offers, demand parameters that produce the same observed choice law must also imply the same revenue comparison. Appendix~\ref{app:design-efficiency} separates the local information bound from the conditions for attaining it.

\begin{proof}
See Appendix~\ref{app:zero-regret-proof}.
\end{proof}

When the span condition fails, the missing information must come from offers with positive opportunity cost. A smaller budget limits how often these offers can be used. Provided that some allocation supplies the missing information, the next theorem identifies the exact leading variance coefficient and the limiting optimal spending shares.

All quantities below are evaluated at the fixed parameter $\theta$.
Let $W=\operatorname{Null}(M_{\mathcal O})$, and let $P_W$ denote
orthogonal projection onto $W$. Write $h_k=P_Wg_k$ for the part of
the target gradient not informed by revenue-maximizing offers.
For each offer with $d_S>0$, define its information per unit of opportunity
cost in these missing directions by
\[
A_S=\left.\frac{P_W\mathcal I_SP_W}{d_S}\right|_W.
\]
For nonnegative $y=(y_S)_{S:d_S>0}$, where $y_S$ is the fraction of the
opportunity-cost budget assigned to $S$, define
\[
\Psi_\theta(y)=\max_{k\in\Kcal}
h_k^\top\left(\sum_{S:d_S>0}y_SA_S\right)^\dagger h_k.
\]
The quadratic form is infinite if its target is outside the matrix range.
Projected vectors and matrices are represented in orthonormal coordinates on $W$.
Below, $\operatorname{dist}(y,\mathcal Y)$ denotes Euclidean distance from
$y$ to the set $\mathcal Y$.

\begin{theorem}[Exact frontier and optimal costly allocations]\label{thm:exact-frontier}
Fix $\theta$ with $F_\theta(0)=\infty$ and $F_\theta(\bar\eta)<\infty$ for some $\bar\eta>0$. Define
\begin{equation}
L_\theta=\min_{y\ge0:\,\sum_Sy_S\le1}\Psi_\theta(y),
\qquad
\mathcal Y_\theta^\star=\operatorname{argmin}_{y\ge0:\,\sum_Sy_S\le1}
\Psi_\theta(y).
\label{eq:projected-cost-design}
\end{equation}
Then $0<L_\theta<\infty$, the minimum is attained, and every minimizer
satisfies $\sum_Sy_S=1$. Moreover,
\begin{equation}
\lim_{\eta\downarrow0}\eta F_\theta(\eta)=L_\theta.
\label{eq:exact-small-budget}
\end{equation}
For any choice of optimal original designs $q_\eta^\star$, set
$y_{\eta,S}=d_Sq_{\eta,S}^\star/\eta$ for $d_S>0$. Then
\begin{equation}
\operatorname{dist}(y_\eta,\mathcal Y_\theta^\star)\to0,
\qquad
\frac{C_\theta(q_\eta^\star)}{\eta}\to1.
\label{eq:cost-allocation-limit}
\end{equation}
Every accumulation point belongs to $\mathcal Y_\theta^\star$.
If this set contains a single vector, $y_\eta$ converges to that vector.
\end{theorem}

\begin{proof}
See Appendix~\ref{app:exact-frontier-proof}.
\end{proof}

The projection removes information already available from revenue-maximizing
offers. The remaining design problem values each costly offer by its
information per unit of forgone revenue. Since the positive revenue gaps
are fixed, the same limit-set conclusion holds for $q_{\eta,S}^\star/\eta$
after dividing cost shares by $d_S$. 

\begin{corollary}[Inverse-budget law]\label{cor:costly-inference}
Under the assumptions of Theorem~\ref{thm:exact-frontier}, for all sufficiently small $\eta>0$,
$L_\theta/2\le\eta F_\theta(\eta)\le 2L_\theta$.
\end{corollary}
\begin{proof}
This follows from $\eta F_\theta(\eta)\to L_\theta\in(0,\infty)$.
\end{proof}

\begin{corollary}[Information per cost in one dimension]\label{cor:one-dimensional-cost}
If $\dim W=1$, let $a_S$ denote the scalar $A_S$. Then
$L_\theta=(\max_k h_k^2)/(\max_{S:d_S>0}a_S)$,
and optimal solutions of the projected problem assign positive cost shares
only to offers maximizing $a_S$.
\end{corollary}

Figure~\ref{fig:free-costly-geometry} shows how the target gradients relate
to the information supplied by revenue-maximizing assortments.

\begingroup
\setlength{\intextsep}{8pt plus 2pt minus 2pt}
\setlength{\textfloatsep}{10pt plus 2pt minus 2pt}
\begin{figure}[!htbp]
\centering
\input{figures/free_costly_geometry_tikz}
\caption{Information-span characterization of the two local regimes. In the
free regime, every target gradient lies in the information span of
revenue-maximizing assortments. In the costly regime, at least one target gradient
lies outside this span and requires positive-opportunity-cost traffic.}
\label{fig:free-costly-geometry}
\end{figure}
\endgroup

The exact small-budget law also relates precision to the total expected opportunity
cost of $m$ experimental offers. Call $\theta$ \emph{costly} for $\Kcal$ if
$F_\theta(0)=+\infty$ but $F_\theta(\bar\eta)<\infty$ for some $\bar\eta>0$.
For an optimal allocation $q_\eta^\star\in\mathcal Q^\star(\theta,\eta)$, write
$D_m^\star(\theta,\eta):=mC_\theta(q_\eta^\star)$
for this total cost. On a fixed costly instance, the optimal allocation asymptotically
uses the full budget, so $C_\theta(q_\eta^\star)/\eta\to1$ as $\eta\downarrow0$. Thus
\[
\sqrt{\frac{F_\theta(\eta)}{m}}
=(1+o(1))\sqrt{\frac{L_\theta}{D_m^\star(\theta,\eta)}}
\qquad\text{as }\eta\downarrow0.
\]
On a fixed costly instance at small budgets, the optimal local standard error benchmark decreases at the inverse square root of total opportunity cost. Corollary~\ref{cor:small-budget-sacrifice} gives this relation. This relation concerns the local benchmark at each budget. The learned implementation below keeps its budget and reference share fixed.

\paragraph{What the span condition means under MNL demand.}
Let $U(\theta):=\bigcup_{S\in\mathcal O(\theta)}S$ be the set of products that appear in at least one revenue-maximizing assortment. With positive attractions and an outside option, $\mathcal I_S(\theta)$ is positive definite on the coordinates indexed by $S$ and zero on the other coordinates. Indeed, for a vector $x$ supported on $S$, its quadratic form is the variance of a random variable that takes value $x_i$ when product $i$ is chosen and zero when the outside option is chosen. Every such outcome has positive probability, so this variance is positive whenever $x\ne0$. It follows that
\[
\operatorname{Range}\{M_{\mathcal O}(\theta)\}
=\{x\in\mathbb R^N:x_i=0\text{ for }i\notin U(\theta)\}.
\]
Thus first-order inference is free exactly when every target gradient is zero outside $U(\theta)$. The condition concerns the gradients. Contributions from the two revenues in a contrast can cancel for some products.

\begin{corollary}[An MNL assortment selection rule]\label{cor:cost-information-rule}
Under MNL demand, suppose a single target has
$h=\alpha e_j\ne0$, where product $j$ is absent from every revenue-maximizing
assortment. If a maximizer of
\[
\frac{p_\theta(j\mid S)\{1-p_\theta(j\mid S)\}}{d_S}
\]
contains no other product absent from all revenue-maximizing assortments,
placing all cost share on that offer is projected-optimal. The coefficient is
$\alpha^2$ divided by the displayed maximum.
For the family of all nonempty assortments of size at most $K$, suppose the
revenue-maximizing assortment $B$ is unique. With spare capacity, the best
offer isolating $j$ is $B\cup\{j\}$. If $|B|=K$, it replaces one product
$i\in B$. With
$V_B=\sum_{i\in B}v_i$ and $R^\star=R_\theta(B)$, select $i$ to maximize
\begin{equation}
\frac{v_j(1+V_B-v_i)}
{(1+V_B-v_i+v_j)
 \{v_i(r_i-R^\star)-v_j(r_j-R^\star)\}}.
\label{eq:mnl-exchange-score}
\end{equation}
When this isolating offer attains the unrestricted maximum, it is the
optimal first-order costly experiment for every contrast with missing
component $\alpha e_j$.
The condition is automatic when $j$ is the only missing product.
\end{corollary}

The rule can apply even when several information directions are
missing: a single contrast may need only one of them. Joint targets are
governed by the full projected program~\eqref{eq:projected-cost-design}.
The exchange score balances the incumbent's revenue contribution against
the incoming product's contribution and the dilution of its choice feedback.

For example, consider three products with attractions $(1,1,1)$, revenues $(1,1,1/4)$, and all nonempty assortments of size at most two. The unique revenue-maximizing assortment is $\{1,2\}$, with revenue $2/3$. Suppose the seller wants to compare the two singleton offers. The contrast $R(\{1\})-R(\{2\})$ has gradient $(1/4,-1/4,0)$ in log-attraction coordinates. Observations of $\{1,2\}$ supply information in both required directions, so this comparison has a finite zero-cost variance benchmark. The useful experimental assortment is outside the target pair.

Now suppose the seller wants to evaluate adding product 3 to the offer of product 1. The gradient of $R(\{1,3\})-R(\{1\})$ is $(-1/18,0,-1/18)$. Choices from $\{1,2\}$ reveal no information about the third coordinate. Offers containing product 3 are therefore needed, and all such feasible offers have positive opportunity cost. Singleton offers identify all three attractions, so the positive-budget feasibility condition holds. The best missing-coordinate information per cost is $8/9$, attained by $\{1,3\}$ and $\{2,3\}$, whereas the singleton $\{3\}$ gives $6/13$. Corollary~\ref{cor:cost-information-rule} therefore gives $L_\theta=1/288$. Indeed, allocating $4\eta$ to $\{1,3\}$ and the remainder to $\{1,2\}$ has cost $\eta$ and variance constant $1/(288\eta)+1/24$ for $0<\eta<1/4$.

A unique optimal experiment can lie outside the target pair. An offer with a lower opportunity cost can also provide less information per unit of that cost (see Appendix~\ref{app:oracle-illustrations}).

\FloatBarrier

\section{Learning and Implementing a Stabilized Design}\label{sec:implementation}
Section~\ref{sec:overlay} characterizes the oracle frontier at a fixed demand parameter. We now use pilot data to select a stabilized allocation under unknown demand, with the budget and reference share held fixed. The implementation accounts for uncertainty in estimated opportunity costs, retains full-rank information for likelihood inference, and converts allocation fractions into offer counts.

The planned set $I_T$ contains $n_T=p_T+m_T$ rounds: $p_T$ pilot observations select the design, and $m_T$ post-pilot observations estimate the target contrasts. The budget $\eta$ applies to the average opportunity cost of the post-pilot offers. Pilot costs enter total regret separately, alongside the cost of the base-policy rounds. We write $\widehat q_T$ for the selected fractional allocation and $\overline q_T$ for its rounded empirical allocation.

\subsection{Black-Box Overlay Architecture}\label{subsec:blackbox-overlay}
We use a base choice bandit policy $\Bcal$ to select assortments on the $T-n_T$ base-policy rounds. A separate experimental design rule selects the assortments on the $n_T$ planned rounds. The base policy uses only observations from its preceding base-policy rounds. Its round counter increases by one on each of these rounds. A planned round leaves both its counter and its observation history unchanged. Keeping the histories separate lets us use the base policy's existing regret guarantee and analyze the experimental design on its own. In the pilot-based design, we estimate the contrasts from the post-pilot sample.

Conditional stationarity in \eqref{eq:model} lets the base policy's regret guarantee carry over to its own rounds. The next theorem initially charges each planned round the maximum possible loss. The budgeted design will refine this charge for post-pilot offers in Section~\ref{subsec:end-to-end-guarantees}. Figure~\ref{fig:pilot-safe-pipeline} shows the implementation, and Algorithm~\ref{alg:overlay} in Appendix~\ref{app:overlay-algorithm} gives the general procedure.

\FloatBarrier
\begingroup
\setlength{\intextsep}{8pt plus 2pt minus 2pt}
\setlength{\textfloatsep}{10pt plus 2pt minus 2pt}
\begin{figure}[!htbp]
\centering
\input{figures/pilot_safe_pipeline_tikz}
\caption{Two-track implementation. Base-policy and planned rounds are
interleaved, with separate observation histories. The base policy updates
only from its own observations. Pilot observations are used to tighten the
opportunity cost budget and select a stabilized allocation, which is rounded
before deployment. Contrast inference uses only post-pilot observations
and conditions on the pilot-selected schedule.}
\label{fig:pilot-safe-pipeline}
\end{figure}
\endgroup

For an instance $\theta$, let $B_{\Bcal}(m,\theta)$ be an upper bound on the expected pseudo-regret of the base policy after $m$ of its own rounds.

Let $\pi_{\rm overlay}$ denote the policy induced by
Algorithm~\ref{alg:overlay}, and write
$\Reg_{\rm overlay}(T,\theta)
:=
\Reg_{\pi_{\rm overlay}}(T,\theta)$.
When a high-probability guarantee is used,
$B_{\Bcal}^{\rm hp}(m,\theta;\delta)$ denotes a pseudo-regret bound
that holds with probability at least $1-\delta$.

\begin{theorem}[Black-box overlay guarantee]\label{thm:overlay}
Suppose rewards are normalized to $[0,1]$ and the conditional-stationarity condition in \eqref{eq:model} holds. For any deterministic set of planned rounds $I_T$ and any planned design rule, Algorithm~\ref{alg:overlay} satisfies
\begin{equation*}
\E_\theta[\Reg_{\rm overlay}(T,\theta)]
\le n_T+B_{\Bcal}(T-n_T,\theta).
\end{equation*}
If the base-policy guarantee holds with probability at least $1-\delta$, the corresponding overlay bound holds with the same probability after adding the deterministic term $n_T$.
\end{theorem}

\begin{proof}
Every planned round contributes at most one unit of pseudo-regret. At each base-policy round, conditional stationarity in \eqref{eq:model} gives the customer's choice the same conditional distribution as in an uninterrupted run of $\Bcal$. The base policy also selects its assortment using only its own preceding observations. Its trajectory over the $T-n_T$ base-policy rounds therefore has the same distribution as an uninterrupted run of that length under the same environment. Adding the regret from the planned rounds gives the stated bound. The same decomposition applies on the base policy's high-probability guarantee event.
\end{proof}

The bound separates base-policy performance from the choice of experimental offers. We next use the opportunity-cost budget to choose those offers.

\subsection{Pilot-Stabilized Operational Implementation}\label{subsec:operational-design}
\paragraph{Stabilization.}
The oracle criterion needs information about the target comparisons. Our implementation fits all demand parameters and uses a Gaussian approximation supported by full-rank information. We therefore reserve a fixed share of offers for a reference design that covers every parameter direction. This stabilizes the fit and can add opportunity cost.

Fix $\epsilon\in(0,1)$ and a reference design
$q^{\rm ref}\in\Delta(\St)$, and define
\begin{equation*}
\mathcal D_\epsilon^{\rm ref}
:=
\{(1-\epsilon)\widetilde q+\epsilon q^{\rm ref}:
  \widetilde q\in\Delta(\St)\}.
\end{equation*}
The mixture reserves an $\epsilon$ share for $q^{\rm ref}$, and $\widetilde q$ determines the allocation of the remaining share.
The stabilized oracle value is
\begin{equation}
\tau_\epsilon^\star(\theta,\eta)
:=
\min_{\substack{
q\in\mathcal D_\epsilon^{\rm ref}\\
\sum_S q_S d_S(\theta)\le\eta}}
V_\theta(q),
\label{eq:stabilized-oracle}
\end{equation}
and $\mathcal Q_\epsilon^\star(\theta,\eta)$ denotes its optimizer set.
If $M(q^{\rm ref},\theta)\succ0$, every stabilized design is full rank.
Write $\eta_{\rm ref}=C_\theta(q^{\rm ref})$. For
$\eta>\epsilon\eta_{\rm ref}$, put
$\eta'=(\eta-\epsilon\eta_{\rm ref})/(1-\epsilon)$. Then
\begin{equation}
F_\theta(\eta)\le\tau_\epsilon^\star(\theta,\eta)
\le\frac{F_\theta(\eta')}{1-\epsilon}.
\label{eq:stabilized-oracle-comparison}
\end{equation}
The lower bound follows from restricting the feasible designs. The upper
bound mixes an ideal design at budget $\eta'$ with the reference design,
accounting for its cost before allocating the remaining traffic.
The reference allocation uses $\epsilon\eta_{\rm ref}$ of the per-offer budget. On a fixed costly instance with $\eta_{\rm ref}>0$, let $\epsilon=\epsilon(\eta)$ satisfy $\epsilon(\eta)\eta_{\rm ref}=o(\eta)$. Then $\epsilon(\eta)\to0$ and $\eta'/\eta\to1$. Applying Theorem~\ref{thm:exact-frontier} to both bounds in~\eqref{eq:stabilized-oracle-comparison} gives $\eta\tau_{\epsilon(\eta)}^\star(\theta,\eta)\to L_\theta$. Section~\ref{subsec:end-to-end-guarantees} establishes convergence of the learned variance criterion to the stabilized oracle value at a fixed budget and reference share.

\paragraph{Pilot budget tightening.}
Let $\widehat\theta_{p_T}$ be the choice model parameter estimate obtained from the $p_T$ pilot observations. The estimated opportunity costs may understate their true values. We use a nonnegative margin $\gamma_T$, computed from the pilot data, to account for this error. The pilot certificate bounds this underestimation simultaneously for all assortments on an event $\mathcal E_{p,T}$:
\begin{equation}
d_S(\thetastar)
\le
d_S(\widehat\theta_{p_T})+\gamma_T,
\qquad S\in\St.
\label{eq:generic-pilot-certificate}
\end{equation}
The margin $\gamma_T$ is measured in opportunity-cost units:
subtracting it from $\eta$ protects the true per-design-round budget against
pilot estimation error. Whenever the tightened problem is feasible, select
\[
\widehat q_T
\in
\mathcal Q_\epsilon^\star
\!\left(
\widehat\theta_{p_T},
\eta-\gamma_T
\right).
\]
Then
$\sum_S\widehat q_{T,S}d_S(\thetastar)\le\eta$
on $\mathcal E_{p,T}$.
The Euclidean and confidence-set constructions of $\gamma_T$, together with
the consistency conditions for $\widehat q_T$, are given in
Appendix~\ref{app:sdp-technical}. The same margin that protects the cost constraint can rule out useful allocations. An imprecise pilot leaves less budget for the post-pilot design, which may then be infeasible. For the MNL specialization, Proposition~\ref{prop:box-certificate} constructs a margin from an attraction confidence box. The pilot-size experiments compare this margin with the Lipschitz margin and assess precision and fallback behavior.

\paragraph{Reference-first rounding.}
The selected allocation specifies fractions of the post-pilot offers. We convert these fractions into integer counts before deployment. Rounding small weights to zero could remove assortments from the reference design and destroy full-rank information. We therefore reserve one offer for each assortment used by the reference design, then round the allocation of the remaining offers. Let $s_{\rm ref}:=|\operatorname{supp}(q^{\rm ref})|$ denote the number of assortments used by the reference design.

For a fixed $q\in\mathcal D_\epsilon^{\rm ref}$ feasible at budget $\eta$
under a reference parameter $\bar\theta$, assume
$M(q^{\rm ref},\bar\theta)\succ0$. When $m\ge s_{\rm ref}$, reference-first
rounding produces integer counts $(m_S)_{S\in\St}$ that sum to $m$ and retain
every reference-support assortment. The resulting information matrix
$M_m:=\sum_{S\in\St}m_S\mathcal I_S(\bar\theta)$ is full rank, and
\[
\sum_{S\in\St}m_Sd_S(\bar\theta)\le m\eta+O_q(1).
\]
For all sufficiently large $m$, the largest first-order variance benchmark
across the target contrasts satisfies
\[
\max_{k\in\Kcal}g_k(\bar\theta)^\top M_m^{-1}g_k(\bar\theta)
\le\frac{V_{\bar\theta}(q)}{m-O_q(1)}.
\]
For a fixed allocation, the rounding allowance is bounded independently of the sample size. Theorem~\ref{thm:design-rounding} in Appendix~\ref{app:rounding-details} gives exact constants and a certificate computed from the executed schedule for smaller $m$.

Algorithm~\ref{alg:sdp-overlay} uses a rounded reference fallback if a feasibility or rank check fails. Its cost can exceed the requested budget. The budget guarantee below applies when the pilot certificate is valid and the tightened design is selected. Let $m_{T,S}$ denote the number of post-pilot offers of assortment $S$.

Under a cap on total expected opportunity cost, a conservative deployment can reserve an upper bound for the pilot cost before sampling. Execute the rounded post-pilot schedule only if its certified cost fits within the remaining allowance. Otherwise, stop the planned experiment. This rule respects the cap on the certificate event when the reserve covers the pilot cost.

The post-pilot estimator fits the choice model by minimizing negative log likelihood plus a fixed quadratic penalty. The penalty stabilizes the fit, and its relative weight decreases as observations accumulate. Substituting the fitted parameter into each revenue difference gives the plug-in contrast estimates.

\begin{algorithm}[H]
\caption{Pilot-stabilized SDP overlay}
\label{alg:sdp-overlay}
\begin{algorithmic}[1]
\Require Base policy $\Bcal$, pilot size $p_T$, post-pilot design size $m_T\ge s_{\rm ref}$, budget $\eta$, stabilization $\epsilon$, full-rank reference design $q^{\rm ref}$, pilot-tightening mechanism, and fixed penalty $\lambda>0$.
\Ensure Contrast estimate $\widehat{\boldsymbol\Delta}_T$ based on the post-pilot sample and integer post-pilot offer counts $(m_{T,S})_{S\in\St}$.
\State Run the fixed pilot on planned rounds while $\Bcal$ runs only on base-policy rounds.
\State Fit $\widehat\theta_{p_T}$, construct $\gamma_T$, and set $\eta_T^{\rm tight}=\eta-\gamma_T$.
\If{$\eta_T^{\rm tight}\ge0$, the reference matrix is full rank at $\widehat\theta_{p_T}$, and the tightened stabilized SDP is feasible}
  \State Select $\widehat q_T$ and apply reference-first rounding.
\Else
  \State Execute a fixed rounded reference-design fallback.
\EndIf
\State Execute the post-pilot counts and continue $\Bcal$ only on the remaining rounds.
\State Compute the regularized maximum-likelihood estimate $\thetatilde_T$ from the $m_T$ post-pilot design observations using \eqref{eq:estimators} and report $\widehat{\boldsymbol\Delta}_T=\boldsymbol{\Delta}(\thetatilde_T)$.
\end{algorithmic}
\end{algorithm}

\long\def\EndToEndGuarantees{%
\subsection{End-to-End Guarantees}\label{subsec:end-to-end-guarantees}
   
The next theorem bounds revenue loss and shows that the implemented allocation approaches the stabilized optimum. The subsequent corollary connects this design criterion to the sampling variability of the post-pilot estimates.

The design-consistency conditions require the pilot estimate to converge to the true parameter, the tightening margin to vanish, and a stabilized allocation to lie strictly within the budget.

\begin{theorem}[Operational regret and design consistency]\label{thm:end-to-end-overlay}
Run Algorithm~\ref{alg:sdp-overlay}. Let $\mathcal G_T$ be the event on
which $\mathcal E_{p,T}$ holds and Algorithm~\ref{alg:sdp-overlay} selects
the tightened stabilized design. On this event and
on the base policy's guarantee event,
\begin{equation*}
\Reg_{\rm overlay}(T,\thetastar)
\le
B_{\Bcal}^{\rm hp}(T-n_T,\thetastar;\delta_{\Bcal,T})
+p_T+m_T\eta+O(1).
\end{equation*}
If the design and base-policy events fail with probabilities
$\delta_{{\rm des},T}$ and $\delta_{\Bcal,T}$, respectively, the expected
bound adds $O\{T(\delta_{{\rm des},T}+\delta_{\Bcal,T})\}$. Appendix~\ref{app:end-to-end-proof}
gives the exact rounding constants and explains how the failure probabilities
affect expected regret.

For design consistency, suppose that the consistency conditions in
Proposition~\ref{prop:pilot-design} hold and that $m_T\to\infty$.
The tightened stabilized problem is then feasible with probability tending
to one, so Algorithm~\ref{alg:sdp-overlay} selects it with probability tending to one.
With $\overline q_{T,S}=m_{T,S}/m_T$,
\begin{equation}
\operatorname{dist}\!\left(\overline q_T,\mathcal Q_\epsilon^\star(\thetastar,\eta)\right)\to_p0,
\qquad
V_{\thetastar}(\overline q_T)\to_p\tau_\epsilon^\star(\thetastar,\eta).
\label{eq:end-to-end-oracle}
\end{equation}
\end{theorem}

\begin{proof}
See Appendix~\ref{app:end-to-end-proof}.
\end{proof}

The regret bound separates the pilot cost $p_T$, post-pilot cost $m_T\eta+O(1)$, and base-policy regret. Sublinear planned costs and base-policy regret, together with vanishing design and base-policy failure probabilities, yield sublinear expected regret.

The rounded allocation approaches the set of stabilized oracle designs, and its variance criterion converges to their common value at a fixed budget and reference share.

Conditional on the pilot and schedule randomization, the post-pilot offers form a fixed design. The following Gaussian approximation describes the post-pilot contrast estimator using the covariance of the rounded allocation.

\begin{corollary}[Conditional post-pilot inference]\label{cor:end-to-end-inference}
Under the bounded MNL model, run Algorithm~\ref{alg:sdp-overlay} with fixed
$\epsilon\in(0,1)$ and $\lambda>0$, and let $m_T\to\infty$. Assume that
$\thetastar$ is an interior point of $\Theta$ and
that $M(q^{\rm ref},\thetastar)\succ0$. Let $\mathcal P_T$ be the
sigma-field generated by the pilot history and by all tie-breaking,
rounding, and schedule-order randomization, excluding every post-pilot
design outcome. Assume that, conditional on $\mathcal P_T$, customer choices
on the post-pilot design rounds are independent and follow the MNL choice law
for their scheduled assortments. Let $d_{\rm BL}$ denote the bounded-Lipschitz distance
between probability laws. Define the contrast-gradient matrix
$G_\Delta(\thetastar):=(g_k(\thetastar)^\top)_{k\in\Kcal}$ and the covariance map
$\Omega_\Delta(q):=G_\Delta(\thetastar)M(q,\thetastar)^{-1}
G_\Delta(\thetastar)^\top$. Then
\begin{equation}
d_{\rm BL}\!\left(
\mathcal L_{\thetastar}\!\left[
\sqrt{m_T}\{\widehat{\boldsymbol\Delta}_T-\boldsymbol{\Delta}(\thetastar)\}
\mid\mathcal P_T\right],
\mathcal N\{0,\Omega_\Delta(\overline q_T)\}
\right)\to_p0,
\label{eq:end-to-end-inference}
\end{equation}
If, in addition, the design-consistency conditions in
Proposition~\ref{prop:pilot-design} hold, then
\[
\max_{k\in\Kcal}[\Omega_\Delta(\overline q_T)]_{kk}
=
V_{\thetastar}(\overline q_T)
\to_p\tau_\epsilon^\star(\thetastar,\eta).
\]
Thus the largest diagonal entry of the approximating Gaussian covariance
converges to the stabilized-oracle value. Different optimal allocations can
attain this same value while yielding different covariance matrices.
A sufficient condition for a single
unconditional Gaussian limit is given in
Appendix~\ref{app:covariance-uniqueness}.
\end{corollary}

\begin{proof}
See Appendix~\ref{app:end-to-end-proof}.
\end{proof}

For a large post-pilot sample, each diagonal entry of $\Omega_\Delta(\overline q_T)$ divided by $m_T$ gives a contrast's approximate variance. Wald intervals use estimated standard errors and normal critical values around the plug-in estimate. Appendix~\ref{app:design-efficiency} gives the local information bound and attainability conditions. Corollary~\ref{cor:sdp-finite-sample-bridge} compares the rounded Fisher information with conservative likelihood curvature. Finite-sample bounds and practical intervals are developed in Appendices~\ref{sec:general}--\ref{subsec:mnl-proxies} and~\ref{app:interval-statements}.
}%

\long\def\GeneralInferenceAppendix{%
\section{Finite-Sample Inference for Predictable Planned Designs}\label{sec:general}
This appendix establishes the finite-sample likelihood guarantee for the
post-pilot estimator in Algorithm~\ref{alg:sdp-overlay}. The bound controls estimation error using likelihood curvature throughout the parameter set. It complements the local Gaussian approximation used to interpret the allocation criterion.

\subsection{Planned Sample and Estimator}\label{subsec:algorithm}
Here $I_T\subseteq[T]$ denotes the planned rounds used for inference. In
Algorithm~\ref{alg:sdp-overlay}, these are $I_T^{\rm post}$. Put
$\chi_t:=\one{t\in I_T}$. The assignment may depend on the observed past,
but $\chi_t$ and $S_t$ must be $\Fcal_{t-1}^{+}$-measurable: both are fixed
before $Y_t$ is observed.

For observation $t$, write
$\ell_t(\theta):=-\log p_\theta(Y_t\mid S_t)$. For a fixed
$\lambda>0$, define the planned-sample estimator by
\begin{equation}
\thetatilde_T\in\argmin_{\theta\in\Theta}
\left\{\frac{\lambda}{2}\|\theta\|_2^2+
\sum_{t\in I_T}\ell_t(\theta)\right\}.
\label{eq:estimators}
\end{equation}
In Algorithm~\ref{alg:sdp-overlay}, the pilot observations determine
$\widehat q_T$ and are excluded from $\thetatilde_T$. For rounded counts
$(m_S)_S$, the SDP uses
$M_{m_T}(\bar\theta)=\sum_Sm_S\mathcal I_S(\bar\theta)$. The finite-sample
bound below uses a predictable lower-Hessian matrix.
Corollary~\ref{cor:sdp-finite-sample-bridge} compares the two matrices
for the bounded MNL model.

\subsection{Likelihood Conditions and Confidence Geometry}
\label{subsec:assumptions}\label{subsec:confidence}
The assumptions distinguish score variability, likelihood curvature, and
the translation from parameter error to revenue error.

\begin{assumption}[Bounded parameter set]\label{ass:bounded}
$\Theta\subset\R^d$ is nonempty, convex, and compact, contains $\thetastar$,
and satisfies $\|\theta\|_2\le B_\Theta$ for every $\theta\in\Theta$. For every
possible $(S,Y)$, the negative log likelihood is twice continuously
differentiable on an open neighborhood of $\Theta$.
\end{assumption}

\begin{assumption}[Score moment-generating-function proxy]\label{ass:mgf}
Define the score at the truth by
\[
\xi_t:=\nabla_\theta\log p_{\thetastar}(Y_t\mid S_t)
=-\nabla\ell_t(\thetastar).
\]
There exists a predictable sequence of positive semidefinite matrices
$\{\Sigma_t\}_{t\ge1}$ such that, for every $u\in\R^d$,
\[
\E\!\left[\exp(u^\top\xi_t)\mid\Fcal_{t-1}^{+}\right]
\le\exp\!\left(\frac12u^\top\Sigma_tu\right).
\]
In particular, $\E[\xi_t\mid\Fcal_{t-1}^{+}]=0$ under correct
specification.
\end{assumption}

\begin{assumption}[Per-round curvature domination and link]
\label{ass:curvature}
There exists a predictable sequence of positive semidefinite matrices
$\{\Gamma_t\}_{t\ge1}$ and a constant $\rho\in(0,1]$ such that
\begin{align}
\nabla^2\ell_t(\theta)&\succeq\Gamma_t,
&&\forall t,\ \forall\theta\in\Theta,\notag\\
\Gamma_t&\succeq\rho\Sigma_t,
&&\forall t.
\label{eq:link}
\end{align}
\end{assumption}

\begin{assumption}[Lipschitz revenue map]\label{ass:lipschitz}
There exists $L_R<\infty$ such that, for every $S\in\St$ and all
$\theta,\theta'\in\Theta$,
\[
|R_\theta(S)-R_{\theta'}(S)|
\le L_R\|\theta-\theta'\|_2.
\]
\end{assumption}

For a positive definite matrix $A$, write
$\|x\|_A:=\sqrt{x^\top A x}$, and let $I_d$ denote the $d$-dimensional
identity matrix. Define the planned-sample matrices
\[
W_T^{\rm exp}:=\lambda I_d+\sum_{t\in I_T}\Sigma_t,
\qquad
V_T^{\rm exp}:=\lambda I_d+\sum_{t\in I_T}\Gamma_t,
\]
and the confidence radius
\[
\beta_T^{\rm exp}(\delta):=
\sqrt\lambda B_\Theta+
\frac{1}{\sqrt\rho}
\sqrt{2\log\!\left(
\frac{\det(W_T^{\rm exp})^{1/2}}
{\det(\lambda I_d)^{1/2}\delta}
\right)}.
\]
The superscript $\mathrm{exp}$ denotes planned observations.
Assumption~\ref{ass:curvature}, including $\rho\le1$, gives
$V_T^{\rm exp}\succeq\rho W_T^{\rm exp}$.

\subsection{Finite-Sample Planned-Sample Guarantee}\label{subsec:mainthm}
\begin{theorem}[Predictable planned-sample inference]\label{thm:main}
Assume Assumptions~\ref{ass:bounded}--\ref{ass:lipschitz}, and assume
$\chi_t=\one{t\in I_T}$ is $\Fcal_{t-1}^{+}$-measurable. Then,
for every $\delta\in(0,1)$, with probability at least $1-\delta$,
\[
\|\thetatilde_T-\thetastar\|_{V_T^{\rm exp}}
\le\beta_T^{\rm exp}(\delta/2).
\]
Consequently, simultaneously for every $(a,b)\in\Kcal$,
\begin{equation}
\left|\widehat\Delta_T^{(a,b)}-\Delta_R^{(a,b)}\right|
\le
2L_R\frac{\beta_T^{\rm exp}(\delta/2)}
{\sqrt{\lambda_{\min}(V_T^{\rm exp})}},
\qquad
\widehat\Delta_T^{(a,b)}
:=R_{\thetatilde_T}(S_a)-R_{\thetatilde_T}(S_b).
\label{eq:contrast_general}
\end{equation}
\end{theorem}

\begin{proof}
See Appendix~\ref{app:mainproof}.
\end{proof}

\paragraph{Simultaneous intervals.}
On the event in Theorem~\ref{thm:main}, each target contrast belongs to
\begin{equation}
\mathsf{CI}_T^{(a,b)}(\delta)
:=
\left[
\widehat\Delta_T^{(a,b)}
\ \pm2L_R
\frac{\beta_T^{\rm exp}(\delta/2)}
{\sqrt{\lambda_{\min}(V_T^{\rm exp})}}
\right],
\qquad (a,b)\in\Kcal.
\label{eq:simultaneous-ci}
\end{equation}
A single parameter-containment event covers the entire contrast family, so
the width carries no separate $|\Kcal|$ union-bound factor.

}%

\long\def\MNLFiniteSampleAppendix{%
\section{Finite-Sample MNL Likelihood Bridge}\label{subsec:mnl-proxies}
The SDP evaluates designs using Fisher information, while the finite-sample likelihood bound uses a conservative lower bound on curvature. For bounded MNL demand, this appendix verifies the likelihood conditions in Appendix~\ref{sec:general} and compares these two matrices for the rounded schedule.

\begin{proposition}[MNL proxies]\label{prop:mnlproxies}
Under Assumption~\ref{ass:mnlbounded}, the likelihood conditions of
Theorem~\ref{thm:main} hold with dimension $d=N$.
Let $D_t$ be the diagonal matrix that selects the coordinates in $S_t$:
\begin{equation*}
D_t := \Diag\bigl(\one{i\in S_t}\bigr)_{i=1}^N.
\end{equation*}
Then one may take
\begin{equation*}
\Sigma_t := K D_t,
\qquad
\Gamma_t := m_{\rm MNL} D_t,
\qquad
m_{\rm MNL} := \frac{v_{\min}}{(1+K v_{\max})^2},
\qquad
\rho := \frac{m_{\rm MNL}}{K}.
\end{equation*}
Moreover, the revenue map is Lipschitz with
\begin{equation*}
L_R = \sqrt{K}\,v_{\max}.
\end{equation*}
\end{proposition}

\begin{proof}
See Appendix~\ref{app:proof-mnlproxies}.
\end{proof}

The score is supported on the offered coordinates. Bounded attractions
provide the lower bounds on inside and outside choice probabilities used for curvature.

\begin{corollary}[Finite-sample bridge and rounded-matrix certificate]
\label{cor:sdp-finite-sample-bridge}
Fix a pilot reference $\bar\theta$ and a feasible stabilized allocation
$q\in\mathcal D_\epsilon^{\rm ref}$. Assume $M(q^{\rm ref},\bar\theta)\succ0$. Let
$s_{\rm ref}:=\bigl|\operatorname{supp}(q^{\rm ref})\bigr|$.
Apply the reference-first rounding of Theorem~\ref{thm:design-rounding}
to $m_T\ge s_{\rm ref}$ post-pilot offers. Let
$(m_{T,S})_{S\in\St}$ be the resulting offer counts, and define
\[
M_{m_T}(\bar\theta)
:=
\sum_Sm_{T,S}\mathcal I_S(\bar\theta),
\qquad
\Xi_T(\bar\theta)
:=
\max_{k\in\Kcal}
g_k(\bar\theta)^\top
M_{m_T}(\bar\theta)^{-1}
g_k(\bar\theta).
\]
These are the rounded Fisher information and largest contrast variance
criterion at the pilot reference $\bar\theta$.
Then $M_{m_T}(\bar\theta)\succ0$ and
\begin{align}
V_T^{\rm exp}
&\succeq
\lambda I_N+m_{\rm MNL}M_{m_T}(\bar\theta),
\label{eq:sdp-curvature-bridge}\\
\max_{k\in\Kcal}
g_k(\bar\theta)^\top
(V_T^{\rm exp})^{-1}
g_k(\bar\theta)
&\le
\frac{\Xi_T(\bar\theta)}{m_{\rm MNL}}.
\label{eq:sdp-ex-post-bridge}
\end{align}
Let
\[
\mathcal H:=\operatorname{supp}(q),\qquad
\kappa:=\lambda_{\min}\{M(q,\bar\theta)\},
\]
and define the rounding perturbation constant
\[
C_{\mathcal H}:=
\bigl(|\mathcal H|+2s_{\rm ref}\bigr)
\max_{S\in\mathcal H}
\|\mathcal I_S(\bar\theta)\|_{\rm op}.
\]
If $m_T>C_{\mathcal H}/\kappa$, then
\begin{equation}
\Xi_T(\bar\theta)
\le
\frac{V_{\bar\theta}(q)}
     {m_T-C_{\mathcal H}/\kappa},
\qquad
\max_{k\in\Kcal}
g_k(\bar\theta)^\top
(V_T^{\rm exp})^{-1}
g_k(\bar\theta)
\le
\frac{V_{\bar\theta}(q)}
{m_{\rm MNL}\{m_T-C_{\mathcal H}/\kappa\}}.
\label{eq:sdp-directional-bridge}
\end{equation}
Equation~\eqref{eq:sdp-ex-post-bridge} certifies the executed schedule, while
\eqref{eq:sdp-directional-bridge} quantifies its rounding loss relative to the
fractional design. Proposition~\ref{prop:sdp-directional-consequence} adds the
nonlinear-contrast remainder.
\end{corollary}

\begin{proof}
See Appendix~\ref{app:rounded-directional}.
\end{proof}
}%

\EndToEndGuarantees

\section{A General Rate Template and MNL Minimax Sharpness}
\label{sec:tradeoff}

Two allocations can have the same error exponent and different variance constants. Sections~\ref{sec:overlay} and~\ref{sec:implementation} distinguish their precision at a given sample size. We now fix a simple allocation to study how the number of experimental offers affects worst-case regret and estimation error as the horizon grows.

We combine MNL-UCB with balanced singleton offers, distributing planned traffic evenly across products. Because this schedule is fixed before observing outcomes, the estimator uses the whole planned sample without a design-selecting pilot. Lower bounds on a two-product MNL benchmark show that the achievable polynomial exponents are sharp in the worst case.

\RateWiseCriteria

With problem parameters fixed, the balanced-singleton results in
Section~\ref{subsec:balanced-exploration} give upper bounds of the form
\[
\mathcal R_T=\Otilde(\sqrt T+n_T),
\qquad
\mathcal E_T=\Otilde(n_T^{-1/2}).
\]
The lower bound in Section~\ref{subsec:hard-subclass} takes the form
$\mathcal R_T\gtrsim\max\{\sqrt T,\mathcal E_T^{-2}\}$.
In the upper bound, the $\sqrt T$ term comes from the base policy, and $n_T$ bounds the additional regret from the planned offers. The estimation error decreases at an inverse square root rate in the number of these offers.

\subsection{Balanced-Singleton Achievability}
\label{subsec:balanced-exploration}\label{subsec:mnl_rates}
To obtain explicit rates, we use balanced singleton exploration. Each planned
round offers one product, and a fixed cycle distributes these offers evenly
across products. The resulting observations allow us to estimate the MNL
product parameters and hence the target revenue contrasts. We spread the
planned rounds across the selling horizon according to the following schedule.

\begin{definition}[Balanced spaced singleton exploration]\label{def:balanced}
Fix a planned-sample size $n_T\in\{1,\dots,T\}$. Let the planned rounds be
\begin{equation*}
\tau_k := \left\lceil \frac{kT}{n_T} \right\rceil, \qquad k=1,\dots,n_T,
\end{equation*}
and on the $k$-th planned round offer the singleton assortment
\begin{equation*}
S_{\tau_k} = \{i_k\}, \qquad i_k := 1 + ((k-1) \bmod N).
\end{equation*}
All other rounds are base-policy rounds.
\end{definition}

The cycle assigns every product either $\lfloor n_T/N\rfloor$ or
$\lceil n_T/N\rceil$ planned singleton offers. We estimate each target contrast from the regularized maximum likelihood fit to these planned singleton observations. Under
Assumption~\ref{ass:mnlbounded}, estimating the product parameters from these
observations gives the following contrast-error bound: with probability at
least $1-\delta$,
\[
\max_{(a,b)\in\Kcal}
\abs{\widehat\Delta_T^{(a,b)}-\Delta_R^{(a,b)}}
=
\Otilde\!\left(\sqrt{\frac{N}{n_T}}\right)
\qquad (n_T\ge 2N).
\]
Lemma~\ref{lem:singleton-coordinatewise} in
Appendix~\ref{app:mnlproofs} gives the explicit finite-sample constant.

Appendix~\ref{app:mnlproofs} discusses feasible cover designs with the same $T$-dependence.

\paragraph{Joint regret and inference.}
Pairing this schedule with epoch-based MNL-UCB yields a joint guarantee for
revenue regret and planned-sample contrast error.

\begin{theorem}[MNL overlay rate]\label{thm:mnloverlay}
Suppose Assumption~\ref{ass:mnlbounded} holds, every singleton is
feasible, and the base policy is the epoch-based MNL-UCB policy of
\citet{AgrawalAvadhanulaGoyalZeevi2019} on the log-attraction
instance class
$\Theta_0\subseteq[\log v_{\min},0]^N$. Equivalently, the corresponding
inside attractions satisfy $v_i\in[v_{\min},1]$. Assume $v_0=1$,
$r_i\in[0,1]$, and a finite downward-closed assortment family: if
$S\in\St$ and $Q\subseteq S$, then $Q\in\St$. The static optimistic
assortment problem is solved exactly. Then there are absolute constants
$C_1,C_2$ such that, for $m\ge1$,
\begin{equation}
\sup_{\theta\in\Theta_0}B_{\Bcal}(m,\theta)
\le
C_1\sqrt{Nm\log(m+1)}+C_2N\log^2(m+1).
\label{eq:base-mnl-regret}
\end{equation}
Run Algorithm~\ref{alg:overlay} with the balanced spaced singleton
design and compute the planned-sample regularized MLE in
\eqref{eq:estimators}. Then
\begin{align}
\sup_{\theta\in\Theta_0}
\E_\theta[\Reg_{\rm overlay}(T,\theta)]
&\le
C_1\sqrt{N(T-n_T)\log(T-n_T+1)}
\notag\\
&\quad+C_2N\log^2(T-n_T+1)+n_T
\notag\\
&=\Otilde\!\left(\sqrt{N(T-n_T)}+N+n_T\right).
\label{eq:mnl-overlay-regret}
\end{align}
Moreover, uniformly for every $\thetastar\in\Theta_0$, with probability
at least $1-\delta$,
\begin{equation*}
\max_{(a,b)\in\Kcal}
\abs{\widehat\Delta_T^{(a,b)}-\Delta_R^{(a,b)}}
=\Otilde\!\left(\sqrt{\frac{N}{n_T}}\right).
\end{equation*}
\end{theorem}

\begin{proof}
See Appendix~\ref{app:mnlproofs}.
\end{proof}

Under the assumptions of Theorem~\ref{thm:mnloverlay}, fix $N$ and choose
$n_T\asymp T^\alpha$ for $\alpha\in(0,1)$. The worst-case expected regret and
contrast-error criteria from Section~\ref{subsec:rate-criteria} then satisfy
\[
\mathcal R_T=\Otilde\!\left(T^{\max\{1/2,\alpha\}}\right),
\qquad
\mathcal E_T=\Otilde(T^{-\alpha/2}).
\]
In particular, $\alpha=1/2$ gives
\[
\mathcal R_T=\Otilde(T^{1/2}),
\qquad
\mathcal E_T=\Otilde(T^{-1/4}).
\]
Thus order $\sqrt T$ planned rounds give an estimation error rate of
$T^{-1/4}$ while keeping the regret bound at order $\sqrt T$, up to logarithmic
factors. Corollary~\ref{cor:alphafamily} in Appendix~\ref{app:mnlproofs} gives
the formal statement and derivation.

These exponent statements hold with dimension, attraction bounds,
regularization, and cover-design constants fixed. An unknown horizon is handled
by a doubling schedule with an additional logarithmic factor. The capacity-one analysis below establishes the corresponding minimax exponent frontier.

\subsection{Minimax Lower Bound and Rate Frontier}\label{subsec:hard-subclass}
Section~\ref{subsec:balanced-exploration} obtains smaller estimation error by increasing the number of planned offers. To establish the corresponding lower bounds, we consider an MNL model with two products and one product per offer. The target contrast is the difference between the expected revenues of the two singleton assortments. Specifically,
\[
N=2,\qquad
\St=\{\{1\},\{2\}\},\qquad
r_1=r_2=1,
\]
with the single target
$\Kcal=\{(\{1\},\{2\})\}$.
For this benchmark, the attraction parameters lie in the fixed box $[1/3,3]^2$, extending beyond the normalized range in Theorem~\ref{thm:mnloverlay}. Corollary~\ref{cor:capacity-one-overlay} supplies an upper bound on this same benchmark using a two-armed minimax base policy. We compare it with the lower bounds below to establish matching exponents.

For $i\in\{1,2\}$, let $\mu_i=v_i/(1+v_i)$ be the purchase probability when singleton $\{i\}$ is offered. Since $r_1=r_2=1$, $\mu_i$ also equals its expected revenue. The target contrast is the revenue gap $\Delta=\mu_1-\mu_2$. When $\mu_1>\mu_2$, expected regret equals $(\mu_1-\mu_2)\E[T_2(T)]$, where $T_2(T)$ is the number of offers of singleton $\{2\}$.

The hard family below keeps product 1 optimal and varies only product 2's purchase probability. Learning the revenue gap therefore requires offers of the lower-revenue product.

For $0\le f\le1/8$, let $\phi_f$ denote the MNL instance with singleton purchase probabilities $(\mu_1,\mu_2^{(f)})=(3/4,1/4+2f)$. These instances form the family $\Ecal_0=\{\phi_f:0\le f\le1/8\}$. Product 1 is optimal throughout this family, and the contrast equals $\Delta_f=1/2-2f$.

Define the objectives on this hard family
\begin{equation*}
\mathcal R_T^{\Ecal_0}(\pi):=\sup_{0\le f\le1/8}\E_{\phi_f}[\Reg_\pi(T,\phi_f)],\qquad
\mathcal E_T^{\Ecal_0}(\widehat\Delta):=\sup_{0\le f\le1/8}\E_{\phi_f}\abs{\widehat\Delta-\Delta_f}.
\end{equation*}

\begin{theorem}[Minimax product lower bound]\label{thm:lowerbound}
There exists a universal constant $c>0$ such that every design pair $(\pi,\widehat\Delta)$ satisfies
\begin{equation*}
\mathcal E_T^{\Ecal_0}(\widehat\Delta)\sqrt{1+\mathcal R_T^{\Ecal_0}(\pi)}\ge c.
\end{equation*}
\end{theorem}

Observations of product 1 cannot distinguish these instances. The theorem quantifies the cost of learning from product 2: a policy with low expected regret cannot produce uniformly precise estimates of the revenue gap.

\begin{proof}
See Appendix~\ref{app:lowerproof}.
\end{proof}
For estimators measurable with respect to the planned-design sample,
Lemma~\ref{lem:design-sample-lower} in Appendix~\ref{app:lowerproof} also gives
$\mathcal E_T^{\Ecal_0}(\widehat\Delta)\ge
c_0/\sqrt{1+n_T}$. Together with the matching capacity-one upper bound in
Corollary~\ref{cor:capacity-one-overlay}, this identifies the exact inference
exponent for the overlay family.
 
\paragraph{Exponent frontier.}\label{subsec:combined-lower-bound}
   
To combine Theorem~\ref{thm:lowerbound} with the usual minimax regret lower bound, evaluate the criteria $\mathcal R_T$ and $\mathcal E_T$ over the benchmark parameter class $\Theta_0:=\Theta_{\rm cap}=[\log(1/3),\log 3]^2$. This class contains $\Ecal_0$ and also instances where the two singleton revenues are arbitrarily close. These latter instances yield the $\sqrt T$ minimax regret lower bound.

\begin{corollary}[Minimax exponent-level frontier]\label{cor:frontier}
There are constants $c_1,c_2>0$ such that every design pair satisfies
\begin{equation}
\mathcal R_T\ge c_1\sqrt T,
\qquad
\mathcal R_T\ge c_2\mathcal E_T^{-2}-1,
\qquad
\mathcal R_T\gtrsim \max\{\sqrt T,\mathcal E_T^{-2}\}.
\label{eq:frontier-lower}
\end{equation}
   
Because this benchmark is included in the broader MNL problem
family, \eqref{eq:frontier-lower} is also a lower bound for every MNL
problem class containing it. The overlay in
Corollary~\ref{cor:capacity-one-overlay} matches these exponents on the
embedded family, while Corollary~\ref{cor:alphafamily} supplies the same
achievable polynomial rates under the bounded normalized MNL assumptions.

\end{corollary}

\begin{proof}
See Appendix~\ref{app:lowerproof}.
\end{proof}

Writing $\mathcal R_T\asymp T^r$ and
$\mathcal E_T\asymp T^{-p}$ with $p\ge0$, the exponent frontier in Figure~\ref{fig:minimax-frontier} is $r=\max\{1/2,2p\}$.

\begingroup
\setlength{\intextsep}{8pt plus 2pt minus 2pt}
\begin{figure}[!htbp]
\centering
\input{figures/minimax_frontier_tikz}
\caption{The minimax frontier in exponent coordinates. Feasible operating
points satisfy $r\ge\max\{1/2,2p\}$, and the segment
$p\in[1/4,1/2)$, $r=2p$ is pairwise rate-wise nondominated.}
\label{fig:minimax-frontier}
\end{figure}
\endgroup
For the overlay family, $p=\alpha/2$ and
$r(\alpha)=\max\{1/2,\alpha\}$.
Combining the matching upper bound with
Corollary~\ref{cor:frontier} gives, for every
$\alpha\in[1/2,1)$,
\[
\mathcal R_T=\widetilde{\Theta}(T^\alpha),
\qquad
\mathcal E_T=\widetilde{\Theta}(T^{-\alpha/2}).
\]
The formal two-sided statement is
Corollary~\ref{cor:overlay-tightness} in
Appendix~\ref{app:lowerproof}.

For $\alpha<1/2$, increasing the number of planned rounds to order $T^{1/2}$ improves the estimation error rate without changing the regret exponent. These schedules are therefore dominated. For $\alpha\in[1/2,1)$, the corresponding operating points are pairwise rate-wise nondominated. Proposition~\ref{prop:nondominated_interval} in Appendix~\ref{app:lowerproof} records the formal comparison.

\FloatBarrier

\section{Numerical Experiment}\label{sec:numerics}
A synthetic example illustrates how the assortment family affects joint precision. We then use MNL demand calibrated to four Expedia markets to examine the oracle frontiers and evaluate the precision, opportunity cost, and budget protection of the pilot-stabilized design.

\subsection{Synthetic Experiment}\label{subsec:synthetic-experiment}
Consider five products with attractions $v=(1.000,0.788,0.606,0.436,0.315)$ and normalized revenues $r=(1.000,0.840,0.752,0.656,0.536)$. The targets are $R(\{j,5\})-R(\{5\})$, $j=1,2,3$. At the true parameter, we minimize the largest variance constant under $\eta=0.327628$, reserving a uniform-singleton reference share of $\epsilon=0.05$. Table~\ref{tab:synthetic-three-targets} compares an optimized singleton design, a design adding the three target pairs, and the full SDP over all 25 feasible offers of size at most three. The full SDP lowers the criterion by a further $30.3\%$ relative to target-plus-reference while using less opportunity cost.

\begingroup
\setlength{\intextsep}{8pt plus 2pt minus 2pt}
\setlength{\textfloatsep}{10pt plus 2pt minus 2pt}
\begin{table}[!htbp]
\centering\small
\caption{Three-target synthetic illustration under a common budget ceiling. Oracle designs use the true parameter. $V_{\thetastar}(q)$ is the largest first-order variance constant across the three targets. $C_{\thetastar}(q)$ is expected opportunity cost per offer.}
\label{tab:synthetic-three-targets}
\begin{tabular}{lrr}\toprule
Design & $V_{\thetastar}(q)$ & $C_{\thetastar}(q)$\\\midrule
Balanced singletons & 1.1543 & 0.3276\\
Optimized singletons & 0.6582 & 0.2747\\
Target-plus-reference & 0.4109 & 0.1968\\
Full SDP & 0.2864 & 0.06649\\
\bottomrule\end{tabular}
\end{table}

\endgroup

\subsection{Case Study: Expedia-Calibrated Markets}\label{subsec:expedia-case-study}
\subsubsection{Markets, Targets, and Evaluation}\label{subsec:expedia-sdp-evidence}
We calibrate an intercept MNL model from historical choices and displayed sets in Expedia destinations 4562, 8192, 9402, and 13870 \citep{ExpediaKaggle2013}. We use a fixed identifier hash for an approximately 60/20/20 calibration, validation, and test split and select five hotels per market by calibration bookings. We retain searches displaying a selected hotel and treat bookings of other hotels as outside choices. Positive calibration-set median prices, normalized by the largest selected-property median, proxy for room value. Appendix~\ref{app:expedia-calibration} gives the calibration details.

Let $a$ denote the property with the largest normalized price proxy in a market. We evaluate every marginal addition
$\Delta_j(\theta)=R_\theta(\{a,j\})-R_\theta(\{a\})$ for $j\ne a$.
There are four targets per market and sixteen in total. Joint experiments use one design and one fitted parameter vector for all four targets. Single-target benchmarks optimize each contrast separately. The feasible family contains all 25 nonempty subsets of at most three of the five properties. The reference design is uniform over the five singleton offers. The finite-sample design comparisons use a reserved reference share of $\epsilon=0.05$.

The initial comparisons use 500 replications per market, with a shared balanced-singleton pilot of $p_T=20{,}000$ and $m_T=5{,}000$ post-pilot offers per design. We simulate choices at the calibrated parameters. Pilot and main fits use $\lambda=1$. The post-pilot estimator uses only the $m_T$ observations, and the empirical all-planned refit adds the pilot.

For target $j$ and method $A$, $\operatorname{MAE}_{A,j}$ is the mean absolute error across the 500 replications. For a joint design, we report the worst-target MAE,
\[
E_\infty=\max_{j\ne a}\frac1{500}\sum_{b=1}^{500}
|\widehat\Delta_{b,j}-\Delta_j|.
\]
For a fixed stabilized design $q$, the Gaussian benchmark for worst-target MAE is $\sqrt{2V_{\thetastar}(q)/(\pi m_T)}$. Joint comparisons average market-relative reductions across the four markets. For each joint comparator and estimator, we use 5,000 paired replication-block resamples to form simultaneous confidence rectangles for target MAE means across the four markets and propagate them through the maxima and ratios. Single-target summaries average $1-\operatorname{MAE}_{A,j}/\operatorname{MAE}_{B,j}$ equally over the sixteen targets. Appendix~\ref{app:expedia-implementation} gives the paired tests and Holm correction. Marginal Wald intervals use nominal 95\% coverage, and joint coverage uses Bonferroni intervals for the four targets. Monte Carlo uncertainty is conditional on the calibrated parameters.

\subsubsection{Baselines and Cost Accounting}\label{subsec:expedia-baselines}
The tightened full-family SDP (Full) optimizes over all 25 offers. Restricted retains the reference singletons, the four target pairs, and the offer with the highest estimated revenue from the pilot. Equal and optimized pooling combine the same four separately optimized full-family designs, using equal weights or weights that minimize the joint criterion. Both use all resulting observations for each target. D-optimal targets information about the full parameter vector.

Comparisons under a common budget ceiling share the pilot, sample size, reference share, and ceiling $\eta=C_{\thetastar}(q^{\rm ref})$. Restricted and pooled designs use Full's estimated-cost tightening. The initial 20k/5k D-optimal ceiling comparator uses true costs offline. Actual-cost matching uses true costs to match Full's executed post-pilot cost within relative tolerance $10^{-6}$, with no estimated-cost constraint. The shared pilot makes this a match on total design cost as well. Restricted matching uses the same 1,998 feasible market-replications for both methods. Appendix~\ref{app:joint-results} reports the results under each cost constraint.

Post-pilot cost is $\sum_S m_{T,S}d_S(\thetastar)$. Total design cost adds the pilot. Both are expected revenue losses conditional on the executed experimental schedule. Tables report replication means.

\subsubsection{Oracle Budget--Precision Frontiers}\label{subsec:expedia-oracle-frontiers}
We solve the oracle program in Section~\ref{sec:overlay} separately for each of the sixteen targets using all 25 feasible assortments and a zero reference floor. With $\eta_{\mathrm{ref}}=C_{\thetastar}(q^{\rm ref})$, Figure~\ref{fig:expedia-oracle-frontiers} plots $F_{\thetastar}(\eta)/F_{\thetastar}(\eta_{\mathrm{ref}})$ against $\eta/\eta_{\mathrm{ref}}$. Lower values indicate greater precision since $F_{\thetastar}(\eta)/m$ is the local variance benchmark.

Each market has two free targets whose frontiers approach finite zero-cost values and two costly targets that follow the inverse-budget law. For the eight costly targets at $\eta/\eta_{\mathrm{ref}}=10^{-6}$, feasible primal designs and the analytic lower bound certify $1\le\eta F_{\thetastar}(\eta)/L_{\thetastar}\le1+3.04\times10^{-6}$. Appendix~\ref{app:expedia-oracle-frontiers} gives target-level coefficients, numerical checks, and reference-floor comparisons.

All four zero-cost information matrices have rank three. Each costly target needs one of the two missing directions, and its best offer per unit cost isolates that direction as in Corollary~\ref{cor:cost-information-rule}. In market 9402, both optimal paid offers combine the added hotel with hotels 34151 and 25159, omitting anchor 35679. Appendix~\ref{app:expedia-oracle-frontiers} gives a joint example that divides spending between these offers.

\begingroup
\setlength{\intextsep}{8pt plus 2pt minus 2pt}
\setlength{\textfloatsep}{10pt plus 2pt minus 2pt}
\begin{figure}[!htbp]
\centering
\includegraphics[width=0.98\linewidth]{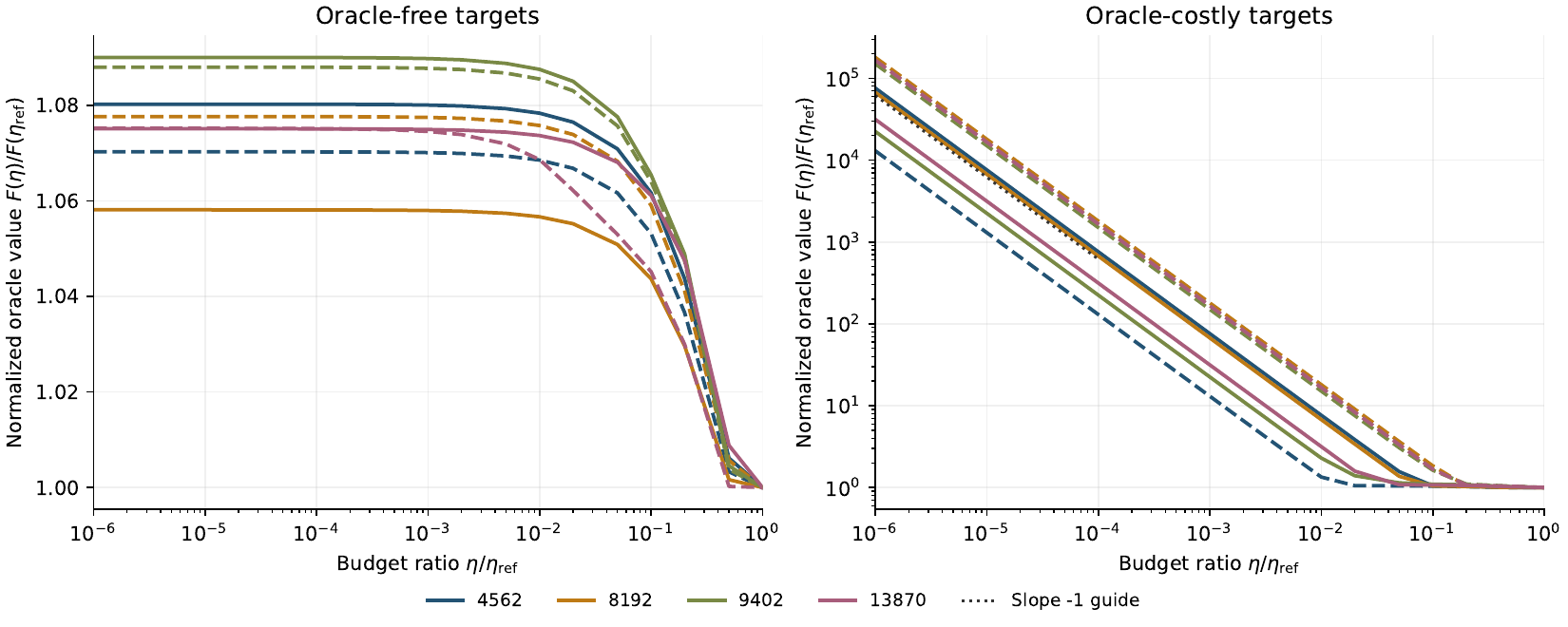}
\caption{Oracle opportunity-cost--precision frontiers for all sixteen Expedia targets. Each curve represents one separately optimized contrast. Colors identify markets. Table~\ref{tab:expedia-oracle-frontiers} maps solid and dashed lines to the hotel contrasts in each panel. Both panels use a logarithmic budget axis. The free-inference panel uses a linear vertical scale, and the costly-inference panel uses a logarithmic vertical scale. Each curve equals one at the reference budget by normalization. The dotted line has slope $-1$.}
\label{fig:expedia-oracle-frontiers}
\end{figure}
\endgroup

\subsubsection{Joint Precision and Single-Target Benchmarks}\label{subsec:expedia-cost-results}
Under the common budget ceiling, Full reduces post-pilot worst-target MAE by $26.84\%$ relative to Restricted, with a propagated 95\% interval of $[8.98\%,41.14\%]$. The shared pilot accounts for $93.62$--$96.82\%$ of Full's total opportunity cost. Full reduces total cost by $0.56$--$4.60\%$ relative to Restricted across markets (Table~\ref{tab:joint-core}).

\begingroup
\setlength{\intextsep}{8pt plus 2pt minus 2pt}
\setlength{\textfloatsep}{10pt plus 2pt minus 2pt}
\begin{table}[!htbp]\centering\small
\setlength{\tabcolsep}{3pt}
\caption{Joint common-ceiling results (20k/5k). Each method's cell gives $10^3E_\infty$ and cumulative post-pilot opportunity cost. Total cost adds the shared pilot. D-optimal uses true costs offline. 500 replications per cell.}
\label{tab:joint-core}
\begin{tabular}{rrrrrrr}\toprule
Market & Shared pilot & Full & Restricted & Equal pool & Optimized pool & D-optimal\\\midrule
4562 & 2,320.84 & 2.388 / 158.15 & 3.144 / 196.05 & 3.955 / 195.25 & 3.461 / 195.52 & 2.772 / 121.53\\
8192 & 2,065.57 & 2.266 / 129.23 & 2.917 / 141.48 & 3.274 / 140.98 & 2.893 / 141.35 & 2.813 / 153.77\\
9402 & 2,575.86 & 3.308 / 84.67 & 4.180 / 212.90 & 4.971 / 229.58 & 4.370 / 229.24 & 3.736 / 132.13\\
13870 & 2,021.69 & 2.697 / 72.62 & 4.506 / 123.35 & 3.864 / 121.57 & 3.386 / 122.05 & 3.468 / 126.20\\
\bottomrule\end{tabular}\end{table}
\endgroup

An offline comparison uses the calibrated true costs to match Full's executed post-pilot opportunity cost. Among the 1,998 matched market-replication pairs, Full reduces post-pilot $E_\infty$ relative to Restricted by $41.98\%$, with a propagated 95\% interval of $[27.71\%,53.36\%]$. Section~\ref{subsec:expedia-robustness} compares Full and Restricted under a common budget ceiling in the independent smaller-pilot cohort.

The pooling comparisons assess whether optimizing all four targets together improves on combining their individual designs. At the common ceiling, Full reduces post-pilot $E_\infty$ by $33.51\%$ relative to equal pooling and by $24.34\%$ relative to optimized pooling. The propagated 95\% interval for optimized pooling is $[6.14\%,38.99\%]$ at the common ceiling. It includes zero at matched cost and in the independent cohort. The joint D-optimal interval also includes zero at matched cost. Figure~\ref{fig:joint-information-comparison} shows the common-ceiling and matched-cost results. Appendix~\ref{app:joint-results} gives the full intervals.

Refitting Full with the pilot lowers its worst-target MAE in all four markets. At the common ceiling, Full's advantage over Restricted is $10.34\%$ for the all-planned refit, with a 95\% interval of $[-10.96\%,27.56\%]$.

Across the sixteen separately optimized contrasts at $\lambda=1$, the tightened SDP reduces post-pilot MAE by an average of $24.29\%$ relative to D-optimal at matched post-pilot opportunity cost, with a 95\% bootstrap interval of $[22.48\%,25.86\%]$. The post-pilot MAE reductions are significant for all sixteen targets after Holm correction. Appendix~\ref{app:expedia-additional-results} reports both sample conventions by market.

Refitting with all planned observations, including the pilot, gives an average MAE reduction of $11.37\%$, with a 95\% bootstrap interval of $[9.47\%,13.03\%]$. Mean MAE is lower for all sixteen targets, with reductions ranging from $4.71\%$ to $20.91\%$. Seven reductions are significant after Holm correction. Average all-planned Wald coverage is approximately $95.0\%$ for both methods.

We also cap D-optimal's post-pilot opportunity cost at the tightened SDP's actual cost in each replication. Under this cap, the tightened SDP reduces post-pilot MAE by an average of $23.41\%$ across the sixteen targets, with a 95\% bootstrap interval of $[21.50\%,25.18\%]$. All sixteen post-pilot comparisons are significant after Holm correction. The empirical all-planned refit gives a $12.29\%$ reduction, with interval $[10.38\%,13.93\%]$. D-optimal spends $21.24\%$ less post-pilot opportunity cost on average. Appendix~\ref{app:expedia-target-diagnostics} describes the target-level coverage, interval length, boundary rates, and cost records.

\begingroup
\setlength{\intextsep}{8pt plus 2pt minus 2pt}
\setlength{\textfloatsep}{10pt plus 2pt minus 2pt}
\begin{figure}[!htbp]\centering
\includegraphics[width=0.98\linewidth]{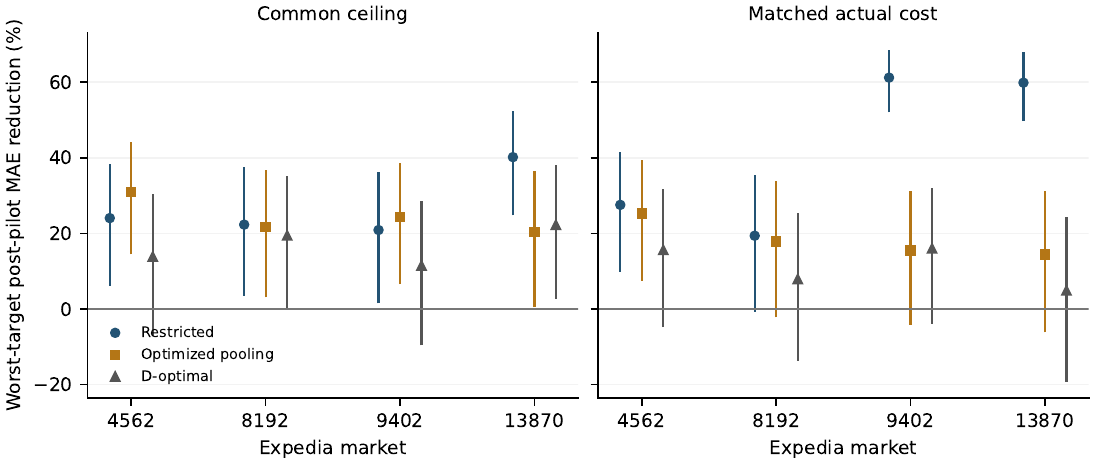}
\caption{Full joint SDP: reductions in post-pilot worst-target MAE under a common ceiling (left) and matched actual cost (right). Points show market-relative reductions. Bars show intervals propagated from simultaneous confidence bounds for target-level MAEs across the four markets. Matching and the common-ceiling D-optimal design use calibrated costs offline. Restricted matching uses 499 replications in markets 9402 and 13870 and 500 elsewhere. Other cells use 500.}
\label{fig:joint-information-comparison}\end{figure}
\endgroup

With allocations and samples fixed, the mean all-planned MAE reduction against D-optimal under the pilot-tightened budget ranges from $11.97\%$ to $12.26\%$ over inference penalties $\lambda\in\{0,0.1,1\}$. Gains against Pilot-best+reference and the pilot-matched mixture also persist at $\lambda=0$. Appendix~\ref{app:expedia-additional-results} gives the complete regularization comparison.

\subsubsection{Pilot Spending and Budget Certificates}\label{subsec:expedia-pilot}
\mbox{}\par\nobreak\vspace{-8pt}
We vary $p_T\in\{500,1000,2000,5000,10000,20000\}$ at fixed total traffic $p_T+m_T=25{,}000$, using one prespecified free and one costly target per market. Separate fixed-post-pilot diagnostics hold $m_T=5{,}000$. Both pilot-size grids retain all fallback runs. The Lipschitz certificate is compared with the box certificate in Proposition~\ref{prop:box-certificate}, using the same pilot confidence boxes.

When all runs fall back, both the pilot and post-pilot phases use balanced singleton offers. Their combined opportunity cost is therefore constant across these pilot sizes at fixed total traffic. At the reference budget and $p_T=5{,}000$, the box certificate leaves enough budget to select a stabilized design in every replication, while the Lipschitz certificate always triggers fallback (Figure~\ref{fig:pilot-total-cost}). For each feasible run, the post-pilot opportunity-cost bound in Theorem~\ref{thm:end-to-end-overlay} applies if the pilot confidence box contains the true demand parameter. Relative to the Lipschitz implementation, the box implementation reduces mean total cost from 2,807.5 to 1,006.3 and post-pilot MAE from 0.002075 to 0.000999. At this split, the nominal design has similar MAE and higher total cost. Larger pilots incur more pilot cost and leave fewer observations for post-pilot estimation. Among the tested box settings that are feasible in every replication, both total cost and post-pilot MAE are higher at 10,000 and 20,000 pilot observations than at 5,000.

\begingroup
\setlength{\intextsep}{8pt plus 2pt minus 2pt}
\setlength{\textfloatsep}{10pt plus 2pt minus 2pt}
\begin{figure}[!htbp]\centering
\includegraphics[width=0.98\linewidth]{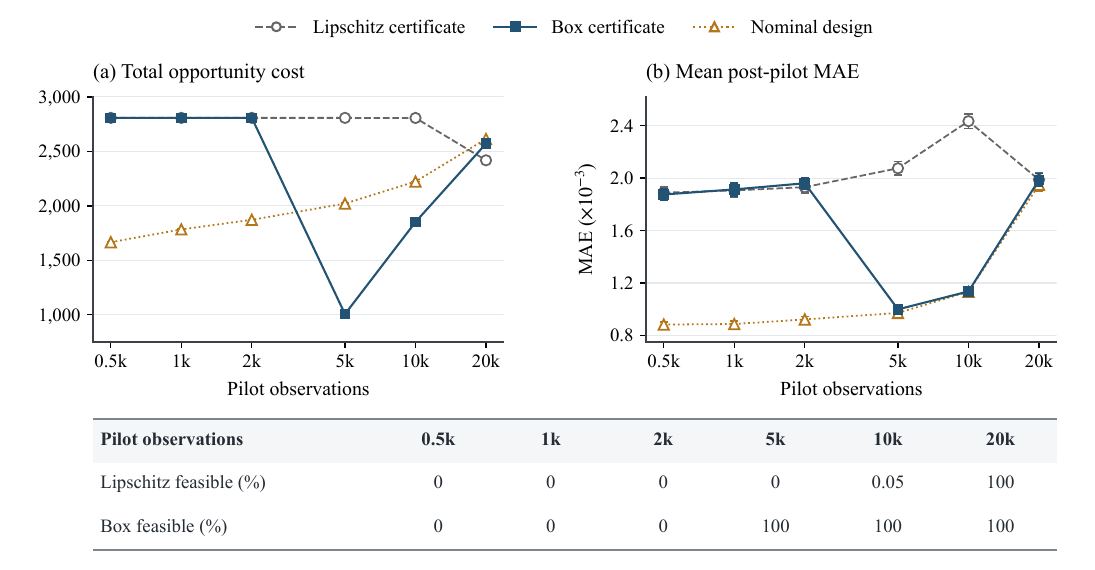}
\caption{Fixed total traffic of 25,000 at $\eta_{\rm ref}$. Total opportunity cost and mean post-pilot MAE are plotted against pilot size across four markets and eight prespecified targets. Points include all fallback runs. Error bars are 95\% Monte Carlo intervals for the means. Pilot-size axes use a logarithmic scale, and lines connect the tested settings. The table reports observed certificate feasibility, counting each shared pilot once per market. Each setting uses 500 unique pilot replications per market.}
\label{fig:pilot-total-cost}\end{figure}
\endgroup

Absolute-budget stress tests compare certificate protection with nominal allocation. At $p_T=20{,}000$ and $\eta=0.06$, the box-certified designs are all feasible with no observed budget violations. Appendix~\ref{app:pilot-results} summarizes the full stress grid and gives the location of the detailed results.

\subsubsection{Independent Small-Pilot Validation}\label{subsec:expedia-robustness}
We freeze $p_T=5{,}000$, $m_T=20{,}000$ after the pilot diagnostic and generate 500 fresh replications per market. Full, Restricted, optimized pooling and D-optimal all use the box certificate and estimated costs. Under correctly specified MNL demand, Full reduces post-pilot worst-target MAE by $30.25\%$ relative to Restricted, with interval $[13.36\%,43.68\%]$. The optimized-pooling and D-optimal intervals include zero. Appendix~\ref{app:misspecification-results} gives both endpoints for every comparator. All four methods execute without fallback or observed budget violation in this cohort.

Appendix~\ref{app:misspecification-results} separates post-pilot and all-planned coverage under MNL mixtures and compares both with direct target-offer estimation. Strong heterogeneity reduces model-based Wald coverage. Appendix~\ref{sec:extensions} gives simultaneous Hoeffding bounds for direct estimation under stationary choice laws. Appendix~\ref{app:scale-results} reports computation up to 1,350 offers.

\section{Conclusion}\label{sec:conclusion}
Our allocation problem asks which offers estimate specified revenue differences most precisely under an opportunity-cost budget. Revenue-maximizing assortments provide free first-order information when they cover the target directions.

When costly offers supply the missing information, the leading coefficient sets the scale of the inverse-budget variance law. The limiting spending shares identify which offers should receive the budget. Theorem~\ref{thm:end-to-end-overlay} gives conditions for convergence of the implemented design's variance criterion to the stabilized oracle value at a fixed budget and reference share. Section~\ref{sec:tradeoff} establishes matching regret and estimation-error exponents on the two-product MNL benchmark.

A three-target synthetic illustration shows how additional assortments can improve joint precision. In the Expedia-calibrated simulations, Full reduces post-pilot worst-target MAE relative to Restricted under the common budget ceiling in both cohorts. For the sixteen separately optimized contrasts, mean target-relative MAE reductions against D-optimal at matched post-pilot opportunity cost are $24.29\%$ for post-pilot estimation and $11.37\%$ for the empirical all-planned refit.

The information-sharing design relies on a correctly specified choice model. Under general stationary choice laws, target-assortment revenues can instead be estimated directly. Appendix~\ref{sec:extensions} gives simultaneous contrast bounds for this direct estimator.

\section*{Code and Data Disclosure}
The replication archive prepared for this paper contains the experiment code, fixed Expedia calibration parameters, aggregated menu counts, and a README with instructions for reproducing the experimental figures and tables. Running the code generates the experimental results. The README also explains how to obtain the original Expedia data from Kaggle and rebuild the calibration. The supplied calibration parameters can be used to run the reported simulations.

\bibliographystyle{ormsv080}
\bibliography{references}

\StartElectronicCompanion
\section*{Electronic Companion}

Appendices~\ref{sec:general} and~\ref{subsec:mnl-proxies} give finite-sample likelihood bounds and their connection to MNL information. Appendix~\ref{app:sdp-technical} supplies the design and implementation arguments. Appendices~\ref{app:design-efficiency} and~\ref{app:conditional-mle} justify the local variance benchmark and the conditional Gaussian approximation, respectively. Appendix~\ref{app:finite-sample-consequences} derives contrast bounds and practical intervals. Appendix~\ref{app:sdp-numerics} records the Expedia calibration, implementation, and empirical Wald diagnostics.

For compactness in the technical derivations, the electronic companion writes
$\Phi(q,\theta):=V_\theta(q)$ and
$\tau^\star(\theta,\eta):=F_\theta(\eta)$.

\section{Overlay Algorithm}\label{app:overlay-algorithm}
\begin{algorithm}[H]
\caption{Interleaving planned offers with a base policy}
\label{alg:overlay}
\begin{algorithmic}[1]
\Require Horizon $T$, base policy $\Bcal$, planned rounds $I_T$ with $|I_T|=n_T$, planned design rule $\pi_{\rm des}$, and contrast class $\Kcal$.
\Ensure Contrast estimates from the planned sample $\{\widehat\Delta_T^{(a,b)}:(a,b)\in\Kcal\}$.
\State Initialize the base-policy history, the planned sample, and the base-policy round counter.
\For{$t=1,\ldots,T$}
    \If{$t\in I_T$}
        \State Select $S_t$ according to $\pi_{\rm des}$, observe $Y_t$, and append $(S_t,Y_t)$ to the planned sample.
    \Else
        \State Advance the base-policy round counter. Select $S_t$ using $\Bcal$ and its own history, observe $Y_t$, and update that history.
    \EndIf
\EndFor
\State Estimate the contrasts using only the planned sample.
\end{algorithmic}
\end{algorithm}

\begingroup
\setlength{\abovedisplayskip}{4.5pt plus 2pt minus 2pt}
\setlength{\belowdisplayskip}{4.5pt plus 2pt minus 2pt}
\setlength{\abovedisplayshortskip}{0pt plus 2pt}
\setlength{\belowdisplayshortskip}{3pt plus 2pt minus 1pt}
\GeneralInferenceAppendix

\MNLFiniteSampleAppendix

\section{Technical Details for the Contrast-Optimal SDP Overlay}\label{app:sdp-technical}

\subsection{SDP Representation}\label{app:sdp-representation}
\begin{proposition}[SDP representation of the ideal design]\label{prop:design-sdp}
Program~\eqref{eq:c-optimal-sdp} is an SDP and its value equals
\begin{equation*}
\min_{q\in\Delta(\St):\,\sum_Sq_Sd_S(\bar\theta)\le\eta}
V_{\bar\theta}(q).
\end{equation*}
\end{proposition}

\begin{proof}[Proof of Proposition~\ref{prop:design-sdp}]
The constraints are affine apart from membership in the positive-semidefinite cone. For $M\succeq0$, the generalized Schur-complement identity
\[
\begin{bmatrix}M&g\\g^\top&\tau\end{bmatrix}\succeq0
\quad\Longleftrightarrow\quad
g\in\operatorname{Range}(M)
\text{ and }g^\top M^\dagger g\le\tau
\]
proves the equivalence.
\end{proof}

\subsection{Zero-Regret Information}\label{app:zero-regret-proof}
\begin{proof}[Proof of Theorem~\ref{thm:free-inference}]
Because $d_S(\theta)\ge0$, every design feasible at budget zero is
supported on $\mathcal O(\theta)$. Hence its information range is contained
in $\operatorname{Range}\{M_{\mathcal O}(\theta)\}$. Conversely, the uniform
distribution $q^0$ on the nonempty set $\mathcal O(\theta)$ satisfies
$M(q^0,\theta)=\frac{1}{|\mathcal O(\theta)|}M_{\mathcal O}(\theta)$
and has exactly that range, proving \eqref{eq:zero-cost-characterization}.
\end{proof}

\subsection{Exact Frontier and Limiting Cost Shares}\label{app:exact-frontier-proof}
\begin{proof}[Proof of Theorem~\ref{thm:exact-frontier}]
Suppress $\theta$ and write $\mathcal D=\{S:d_S>0\}$,
$U=\operatorname{Range}(M_{\mathcal O})$, and $W=U^\perp$.
For a positive semidefinite matrix, use the range-extended identity
\begin{equation}
b^\top A^\dagger b=\sup_z\{2b^\top z-z^\top Az\}.
\label{eq:extended-variational}
\end{equation}
It also shows lower semicontinuity in $A$. Thus $\Psi$ attains its minimum
on the compact cost-share simplex. To see that the minimum is finite,
choose an estimable positive-budget design $q^+$. If $z\in W$ annihilates
its compressed information, positive semidefiniteness implies
$\mathcal I_Sz=0$ for every supported offer. Estimability gives
$h_k^\top z=g_k^\top z=0$. Consequently every $h_k$ belongs to the
compressed information range. Rescaling the costly weights of $q^+$
provides a finite feasible projected criterion.
Some $h_k\ne0$, and $\|\sum y_SA_S\|$ is uniformly bounded on the simplex,
so the minimum $L$ is positive. Homogeneity,
$\Psi(cy)=\Psi(y)/c$, implies that every minimizer spends one.

For any original feasible $q$, put $y_S=d_Sq_S/\eta$. Restrict
\eqref{eq:extended-variational} to $z\in W$. Zero-cost information vanishes
on this space. Hence
\[
\eta\Phi(q,\theta)\ge\Psi(y)\ge L.
\]
This proves $\eta F_\theta(\eta)\ge L$.

For the reverse bound, take any projected minimizer $y^\star$, put
$x_S=y_S^\star/d_S$ and $t=\sum_{\mathcal D}x_S$, and let $q^0$ be uniform
on $\mathcal O$. The finite family has $d_{\min,+}>0$, so $t<\infty$.
For sufficiently small $\eta$, the design
\[
q_\eta=(1-\eta t)q^0+\eta x
\]
is feasible and has cost $\eta$. In $U\oplus W$ coordinates its information is
\[
M(q_\eta)=
\begin{pmatrix}B_\eta&\eta E\\\eta E^\top&\eta D\end{pmatrix},
\quad
B_\eta\to B_0\succ0\text{ on }U,
\quad D=\sum_{\mathcal D}x_S\mathcal I_{S,WW}.
\]
The costly information is positive semidefinite, so $Ez=0$ whenever
$z\in\operatorname{Null}(D)$. Also every $h_k\in\operatorname{Range}(D)$.
The null space therefore decouples exactly. On
$U\oplus\operatorname{Range}(D)$, Schur inversion gives
\begin{equation*}
\begin{aligned}
\eta g_k^\top M(q_\eta)^\dagger g_k
={}&\eta g_{k,U}^\top B_\eta^{-1}g_{k,U}\\
&+(h_k-\eta E^\top B_\eta^{-1}g_{k,U})^\top
(D-\eta E^\top B_\eta^{-1}E)^{-1}
(h_k-\eta E^\top B_\eta^{-1}g_{k,U})\\
\longrightarrow{}&h_k^\top D^\dagger h_k.
\end{aligned}
\end{equation*}
All inverses in the second term are restricted to $\operatorname{Range}(D)$.
If $U=\{0\}$ the terms involving $B_\eta$ are absent. Taking the maximum
over the finite target set gives $\eta\Phi(q_\eta,\theta)\to L$.
Together with the lower bound this proves~\eqref{eq:exact-small-budget}.
The construction works for every projected minimizer and establishes
first-order attainability of its cost shares.

Original optimizers exist by lower semicontinuity on the feasible simplex.
For any selection of them,
\[
L\le\Psi(y_\eta)\le\eta F_\theta(\eta)\to L.
\]
Every convergent subsequence in the compact cost-share simplex has an
optimal limit by lower semicontinuity. Compactness then gives convergence
of the distance to $\mathcal Y_\theta^\star$. Every such limit spends one,
which proves~\eqref{eq:cost-allocation-limit} and the uniqueness assertion.
\end{proof}

\begin{proof}[Proof of Corollary~\ref{cor:one-dimensional-cost}]
In one dimension the denominator is $\sum y_Sa_S\le\max_Sa_S$,
with equality precisely when the cost shares sum to one and are supported
on maximizers.
\end{proof}

\begin{proof}[Proof of Corollary~\ref{cor:cost-information-rule}]
Restrict
\eqref{eq:extended-variational} further to the target coordinate to obtain
\[
\Psi(y)\ge
\frac{\alpha^2}{\sum y_S(A_S)_{jj}}
\ge\frac{\alpha^2}{\max_S(A_S)_{jj}}.
\]
An offer containing $j$ and no other missing product has compressed
information $p_j(S)\{1-p_j(S)\}e_je_j^\top$. If it maximizes the
unrestricted information-per-cost ratio, its unit cost share attains the bound.
For the exchange rule, uniqueness implies $r_i>R^\star$ for every $i\in B$:
otherwise removing $i$ would improve revenue or preserve its optimum.
For $T\subseteq B$, the ratio of the isolating offer $T\cup\{j\}$ equals
\[
\frac{v_j(1+V_T)}
{(1+V_T+v_j)\{R^\star(1+V_T)-\sum_{i\in T}r_iv_i
                         +v_j(R^\star-r_j)\}}.
\]
Every denominator is positive because the offer differs from the unique
revenue maximizer. Adding an available product $i\in B\setminus T$
increases $(1+V_T)/(1+V_T+v_j)$ and decreases the second denominator
by $v_i(r_i-R^\star)>0$. Thus an optimal isolating offer uses all of $B$
when $|B|<K$ and $K-1$ products from $B$ when $|B|=K$.
If $j$ is the only missing product, every informative offer isolates it.
For $T=B\setminus\{i\}$,
$R^\star(1+V_T)-\sum_{\ell\in T}r_\ell v_\ell=v_i(r_i-R^\star)$,
which gives~\eqref{eq:mnl-exchange-score}.
\end{proof}

\subsection{Small-Budget Design Pseudo-Regret}\label{app:small-budget-regret}
\begin{corollary}[Total revenue loss controls local precision]\label{cor:small-budget-sacrifice}
On a fixed costly instance, let $q_\eta^\star$ be any ideal optimizer and
$D_m^\star(\theta,\eta)=mC_\theta(q_\eta^\star)$ for a positive integer
$m$, which may depend on $\eta$. As $\eta\downarrow0$,
\[
\frac{D_m^\star(\theta,\eta)F_\theta(\eta)}{m}\longrightarrow L_\theta,
\]
and hence
\begin{equation}
\sqrt{\frac{F_\theta(\eta)}{m}}
=(1+o(1))\sqrt{\frac{L_\theta}{D_m^\star(\theta,\eta)}}.
\label{eq:small-budget-sacrifice-frontier}
\end{equation}
\end{corollary}
\begin{proof}
Theorem~\ref{thm:exact-frontier} gives
$\eta F_\theta(\eta)\to L_\theta$ and
$C_\theta(q_\eta^\star)/\eta\to1$. Multiplying these limits proves
$C_\theta(q_\eta^\star)F_\theta(\eta)\to L_\theta$.
Substitute $D_m^\star=mC_\theta(q_\eta^\star)$.
\end{proof}

\subsection{Stabilization and Pilot Guarantees}\label{app:stabilization}
Stabilization ensures full-rank information and convergence of the
pilot-selected allocation to the stabilized oracle under the conditions below.

Let $M(q^{\rm ref},\theta)\succ0$, write
$\eta_{\rm ref}=C_\theta(q^{\rm ref})$, and take
$\eta>\epsilon\eta_{\rm ref}$ and
$\eta'=(\eta-\epsilon\eta_{\rm ref})/(1-\epsilon)$.
For $q^\star\in\mathcal Q^\star(\theta,\eta')$, the mixture
$q_\epsilon^{\rm mix}=(1-\epsilon)q^\star+\epsilon q^{\rm ref}$
is full rank and has cost at most $\eta$. Information domination gives
\[
\Phi(q_\epsilon^{\rm mix},\theta)
\le\frac{\tau^\star(\theta,\eta')}{1-\epsilon}.
\]
Restricting the original feasible set gives
$\tau^\star(\theta,\eta)\le\tau_\epsilon^\star(\theta,\eta)$,
proving~\eqref{eq:stabilized-oracle-comparison}. Every stabilized allocation
has cost at least $\epsilon\eta_{\rm ref}$. When $\eta_{\rm ref}\le\eta$,
monotonicity also gives the simpler upper bound
$\tau^\star(\theta,\eta)/(1-\epsilon)$.

\begin{proposition}[Finite-family MNL budget certificate]\label{prop:box-certificate}
Let $[\underline v,\overline v]\subset(0,\infty)^N$ contain the true
attractions on a pilot event of probability at least $1-\delta$.
Let $\widehat\theta_p$ be the pilot estimate, and write $[x]_+:=\max\{x,0\}$.
For each offer, let $\underline R_S$ and $\overline R_S$ be the minimum
and maximum of its MNL revenue over the vertices of its attraction box.
Then the pilot-measurable margin
\begin{equation}
\gamma_{\rm box}=
\max_{S\in\St}\left[
\max_{A\in\St}\overline R_A-\underline R_S
-d_S(\widehat\theta_p)\right]_+
\label{eq:box-cost-certificate}
\end{equation}
satisfies~\eqref{eq:generic-pilot-certificate} on this event.
If another margin $\gamma_{\rm Lip}$ is valid on the same event, their
minimum is also valid. If the box shrinks to the true attractions and
$\widehat\theta_p\to_p\thetastar$, this minimum converges to zero.
\end{proposition}
\begin{proof}
Holding the other attractions fixed, MNL revenue is a linear-fractional
function of one attraction with positive denominator. Its derivative has
constant sign. Successively moving coordinates to endpoints proves that
both extrema over a box occur at vertices. Thus, simultaneously for all $S$,
\[
d_S(\thetastar)=\max_A R_{\thetastar}(A)-R_{\thetastar}(S)
\le\max_A\overline R_A-\underline R_S
\le d_S(\widehat\theta_p)+\gamma_{\rm box}.
\]
Taking the minimum of two valid uniform margins preserves each inequality.
For a finite family, continuity gives uniform convergence of the revenue
bounds and estimated gaps as the box shrinks and the estimate converges.
Consequently $\gamma_{\rm box}\to_p0$ and so does the minimum.
\end{proof}

With at most $K$ products per offer, this calculation evaluates at most
$2^K$ vertices per offer. The simulations use the same simultaneous
Clopper--Pearson attraction box for both margin constructions.

\begin{proposition}[Pilot robustness and stabilized-oracle consistency]\label{prop:pilot-design}
\emph{Consistency.}
Fix $\epsilon\in(0,1)$ and $\eta>0$, and let $\mathcal U\subseteq\Theta$
be a compact neighborhood of $\thetastar$. Assume that
\begin{enumerate}
\item[(i)] $d_S(\vartheta)$ and $\mathcal I_S(\vartheta)$ are continuous on
$\mathcal U$ for every $S\in\St$, and $g_k(\vartheta)$ is continuous on
$\mathcal U$ for every $k\in\Kcal$.
\item[(ii)] for some $\underline\kappa>0$,
$M(q^{\rm ref},\vartheta)\succeq\underline\kappa I_d$ for all
$\vartheta\in\mathcal U$.
\item[(iii)] there are $q^{\rm sl}\in\mathcal D_\epsilon^{\rm ref}$ and
$\xi>0$ such that $\sum_{S\in\St}q_S^{\rm sl}d_S(\thetastar)\le\eta-\xi$.
\end{enumerate}
Let $\widehat\theta_{p_T}\to_p\thetastar$, let $\gamma_T\ge0$ be
pilot measurable with $\gamma_T\to_p0$, and, whenever the tightened oracle
is feasible, take a measurable selection
$\widehat q_T\in\mathcal Q_\epsilon^\star
\!\left(\widehat\theta_{p_T},\eta-\gamma_T\right)$.
Define $\widehat q_T$ arbitrarily in $\mathcal D_\epsilon^{\rm ref}$ on the
infeasibility event. Then the tightened oracle is feasible with probability
tending to one, and
\begin{equation}
\operatorname{dist}\!\left(
\widehat q_T,\mathcal Q_\epsilon^\star(\thetastar,\eta)
\right)\to_p0,
\qquad
\Phi(\widehat q_T,\thetastar)
\to_p\tau_\epsilon^\star(\thetastar,\eta).
\label{eq:pilot-oracle-consistency}
\end{equation}

\emph{Finite-sample certificate.}
Let $\mathcal C_{p,T}\subseteq\Theta$ be a nonempty compact
pilot-measurable set and define
\begin{align*}
\mathcal V_{p,T}&:=\operatorname{conv}
\!\left(\mathcal C_{p,T}\cup\{\widehat\theta_{p_T}\}\right),
\qquad
\rho_{p,T}:=\sup_{\vartheta\in\mathcal C_{p,T}}
\|\vartheta-\widehat\theta_{p_T}\|_2,\\
L_{p,T}&:=\max_{S\in\St}\sup_{\vartheta\in\mathcal V_{p,T}}
\|\nabla_\vartheta R_\vartheta(S)\|_2.
\end{align*}
Set $\gamma_T:=2L_{p,T}\rho_{p,T}$. If $R_\vartheta(S)$ is continuously differentiable on
$\mathcal V_{p,T}$ for every $S\in\St$, then on
$\mathcal E_{p,T}:=\{\thetastar\in\mathcal C_{p,T}\}$
the bound \eqref{eq:generic-pilot-certificate} holds simultaneously for
all $S\in\St$. In particular, every design feasible for the tightened
budget $\eta-\gamma_T$ is feasible at budget $\eta$ under
$\thetastar$.
\end{proposition}

\begin{proof}
Let $b(q,\vartheta)=\sum_Sq_Sd_S(\vartheta)$. Finiteness and continuity,
together with $\widehat\theta_{p_T}\to_p\thetastar$, give
\begin{align*}
\sup_{q\in\mathcal D_\epsilon^{\rm ref}}
|b(q,\widehat\theta_{p_T})-b(q,\thetastar)|&\to_p0,\\
\sup_{q\in\mathcal D_\epsilon^{\rm ref}}
|\Phi(q,\widehat\theta_{p_T})-\Phi(q,\thetastar)|&\to_p0.
\end{align*}
For the second convergence, assumption (ii) gives the uniform eigenvalue
floor $M(q,\vartheta)\succeq\epsilon\underline\kappa I_d$ and hence uniform
continuity of matrix inversion. The strict Slater design is therefore feasible
for the tightened pilot problem with probability tending to one.
From any subsequence, select a further subsequence along which the two
uniform errors and $\gamma_T$ converge to zero almost surely. On each such
path, for any $q^\star\in\mathcal Q_\epsilon^\star(\thetastar,\eta)$ and
fixed $\alpha\in(0,1)$, the mixture
$q^\alpha=(1-\alpha)q^\star+\alpha q^{\rm sl}$ has true slack at least
$\alpha\xi$ and is eventually feasible for the tightened pilot problem.
Optimality, compactness, and uniform convergence imply that every limit
point $\bar q$ of $\widehat q_T$ is feasible and satisfies
$\Phi(\bar q,\thetastar)\le\Phi(q^\alpha,\thetastar)$. Letting
$\alpha\downarrow0$ shows that every limit point belongs to
$\mathcal Q_\epsilon^\star(\thetastar,\eta)$. Compactness and continuity
then give both convergences in \eqref{eq:pilot-oracle-consistency} almost
surely along the selected subsequence. The subsequence characterization
of convergence in probability proves the result for the original sequence.

On $\mathcal E_{p,T}$, the mean-value inequality on
$\mathcal V_{p,T}$ gives
$|R_{\thetastar}(S)-R_{\widehat\theta_{p_T}}(S)|
\le L_{p,T}\rho_{p,T}$ for every $S$. Comparing the two revenue-maximizing
assortments then yields
$d_S(\thetastar)\le d_S(\widehat\theta_{p_T})+\gamma_T$.
Averaging with weights $q_S$ proves the budget certificate.
\end{proof}

\subsection{Reference-First Rounding Details}\label{app:rounding-details}
Write $\mathcal H_{\rm ref}:=\operatorname{supp}(q^{\rm ref})$ and
$s_{\rm ref}:=|\mathcal H_{\rm ref}|$.

\begin{theorem}[Reference-first support-preserving rounding]\label{thm:design-rounding}
Fix a reference parameter $\bar\theta$, a budget $\eta\in[0,1]$, and a
stabilized design $q\in\mathcal D_\epsilon^{\rm ref}$ satisfying
$\sum_{S\in\St}q_Sd_S(\bar\theta)\le\eta$, and assume
$M(q^{\rm ref},\bar\theta)\succ0$. For every integer $m$ at least
the size of the reference support, reference-first rounding produces counts
$(m_S)_{S\in\St}$ that sum to $m$, retain every reference-support action, and
remain within $O_q(1)$ of the fractional counts $mq$. Writing
$M_m:=\sum_{S\in\St}m_S\mathcal I_S(\bar\theta)$, the rounded design is full
rank and satisfies $\sum_{S\in\St}m_Sd_S(\bar\theta)\le m\eta+O_q(1)$.
For the fixed design $q$, the same rounding also satisfies, for all
sufficiently large $m$,
\[
\max_{k\in\Kcal}g_k(\bar\theta)^\top M_m^{-1}g_k(\bar\theta)
\le\frac{V_{\bar\theta}(q)}{m-O_q(1)}.
\]
The proof below gives the threshold, exact constants, and ex post certificate
for smaller $m$.
\end{theorem}

Let $\mathcal H=\operatorname{supp}(q)$, $s=|\mathcal H|$, and $A_{\mathcal H}=s+2s_{\rm ref}$. Reserve one count for each reference-support action. Apply largest-remainder rounding to the fractional counts $((m-s_{\rm ref})q_S)_{S\in\mathcal H}$ and then add the reserved counts. The resulting counts obey
\begin{align*}
\|(m_S)_S-mq\|_1&<A_{\mathcal H},
\\
M_m&\succeq\sum_{S\in\mathcal H_{\rm ref}}\mathcal I_S(\bar\theta)\succ0,
\\
\sum_Sm_Sd_S(\bar\theta)&\le m\eta+A_{\mathcal H}.
\end{align*}
Writing $I_{\max,\mathcal H}=\max_{S\in\mathcal H}\|\mathcal I_S(\bar\theta)\|_{\rm op}$ and $C_{\mathcal H}=I_{\max,\mathcal H}A_{\mathcal H}$ gives
\[
\|M_m-mM(q,\bar\theta)\|_{\rm op}<C_{\mathcal H}.
\]
If $\kappa=\lambda_{\min}\{M(q,\bar\theta)\}$ and $m>C_{\mathcal H}/\kappa$,
matrix domination and inversion give
\begin{align*}
M_m&\succeq\left(m-\frac{C_{\mathcal H}}{\kappa}\right)M(q,\bar\theta),\\
\max_k g_k(\bar\theta)^\top M_m^{-1}g_k(\bar\theta)
&\le\frac{\Phi(q,\bar\theta)}{m-C_{\mathcal H}/\kappa}.
\end{align*}
The regret bound follows from $d_S\in[0,1]$, feasibility of $q$, and the $\ell_1$ discrepancy. These displays prove Theorem~\ref{thm:design-rounding} and make its $O_q(1)$ term explicit.

\subsection{Equivalence, Shadow Price, and Finite-Family Pricing}\label{app:equivalence-pricing}
The budget multiplier $\lambda^\star$ describes how the optimal variance
criterion changes when the opportunity-cost budget is relaxed. The pricing
value bounds the possible improvement from assortments outside the family
used to optimize the design.

\begin{corollary}[Shadow prices and finite-family pricing]\label{cor:stabilized-shadow-price}
For fixed $\theta$, the ideal and stabilized value functions are nonincreasing
and convex in $\eta$. Under Slater, the budget multiplier $\lambda^\star$
satisfies $-\lambda^\star\in\partial_\eta\tau^\star(\theta,\eta)$, with the
analogous statement for $\tau_\epsilon^\star$. For a full-rank design $q$ and
weights $w\in\Delta(\Kcal)$, define
\begin{align*}
G(w)&:=\sum_k w_k g_k(\theta)g_k(\theta)^\top,\\
a_S(q,w)&:=\operatorname{tr}\!\left\{
M(q,\theta)^{-1}\mathcal I_S(\theta)M(q,\theta)^{-1}G(w)
\right\}.
\end{align*}
At a full-rank ideal optimum
$q^\star\in\mathcal Q^\star(\theta,\eta)$, let $w^\star$ be supported on the
active contrasts and let $\lambda^\star$ be the budget multiplier. The ideal
KKT conditions give
\[
a_S(q^\star,w^\star)-\lambda^\star d_S
\le \tau^\star-\lambda^\star\eta,
\qquad S\in\St.
\]
For a full-rank ideal restricted-master optimum $(q^R,\tau^R)$ satisfying
Slater, let $w^R$ and $\lambda^R$ denote its active-contrast and budget
multipliers. Its full-family pricing value is
$U^R:=\max_{S\in\St}\{a_S(q^R,w^R)-\lambda^R d_S\}$.
For $\zeta\ge0$, if a certified upper bound satisfies
$U^R\le\overline U^R\le\tau^R-\lambda^R\eta+\zeta$,
then convexity and the active-component subgradient give
\[
0\le\tau^R-\tau^\star(\theta,\eta)\le\zeta.
\]
Thus exact pricing or a valid upper bound yields a quantitative certificate. The
score of an assortment returned by a heuristic is a lower bound on $U^R$.
\end{corollary}

For the stabilized numerical programs, fix $0\le\epsilon<1$ and an allowed
assortment family $\mathcal A\subseteq\St$ with
$q^{\rm ref}\in\Delta(\mathcal A)$. Write
$q=\epsilon q^{\rm ref}+(1-\epsilon)\widetilde q$, where
$\widetilde q\in\Delta(\mathcal A)$. At a full-rank optimum satisfying Slater,
let $w\in\Delta(\Kcal)$ be supported on the active contrasts and let
$\lambda\ge0$ be the budget multiplier. Set $s_S:=a_S(q,w)-\lambda d_S$,
and let $(1-\epsilon)U$ denote the multiplier for
$\sum_{S\in\mathcal A}\widetilde q_S=1$. The KKT conditions are
\[
\begin{aligned}
s_S&\le U, &
\widetilde q_S(U-s_S)&=0, && S\in\mathcal A,\\
\lambda\{C_\theta(q)-\eta\}&=0, &
w_k\{g_k(\theta)^\top M(q,\theta)^{-1}g_k(\theta)-\tau\}&=0, && k\in\Kcal,
\end{aligned}
\]
with $\sum_k w_k=1$. For a primal-feasible $(q,\tau)$ with $M(q,\theta)\succ0$ and multipliers
$w\in\Delta(\Kcal)$, $\lambda\ge0$, and $U$ satisfying the score
inequalities, an upper bound on suboptimality over the allowed family is
\[
\tau-\sum_k w_k g_k(\theta)^\top M(q,\theta)^{-1}g_k(\theta)
+(1-\epsilon)\left\{U-\sum_{S\in\mathcal A}\widetilde q_Ss_S\right\}
+\lambda\{\eta-C_\theta(q)\}.
\]
The reported KKT residual is the maximum violation of primal feasibility, the
score inequalities, support stationarity, complementarity, and
$\sum_k w_k=1$. At an exact KKT point with $\epsilon=0$,
$U=\tau-\lambda\eta$, which recovers the ideal display above.

\subsection{End-to-End Operational Guarantee}\label{app:end-to-end-proof}
\begin{proof}[Proof of Theorem~\ref{thm:end-to-end-overlay}]
On $\mathcal G_T$, the pilot certificate makes $\widehat q_T$ feasible
at the truth. The rounding bound gives post-pilot design pseudo-regret at
most $m_T\eta+O(1)$, and the pilot costs at most $p_T$.
Theorem~\ref{thm:overlay} supplies the base-policy term. On a failure
event, total regret is at most $T$.

For design consistency, Proposition~\ref{prop:pilot-design} makes the
tightened problem feasible with probability tending to one and gives
\eqref{eq:pilot-oracle-consistency}. Its assumptions also make every fallback
check pass with probability tending to one: $\eta-\gamma_T\ge0$, the reference
matrix is full rank at the pilot estimate, and eventually $m_T\ge s_{\rm ref}$.
On this branch, Theorem~\ref{thm:design-rounding} gives the uniform bound
\[
\|\overline q_T-\widehat q_T\|_1
\le \frac{|\St|+2s_{\rm ref}}{m_T}\to0.
\]
Continuity on the stabilized design set then yields
\eqref{eq:end-to-end-oracle}.
\end{proof}

\begin{proof}[Proof of Corollary~\ref{cor:end-to-end-inference}]
Conditional on $\mathcal P_T$, the post-pilot schedule is fixed before
its outcomes are revealed. In either algorithm branch,
Theorem~\ref{thm:design-rounding} gives
\[
\operatorname{dist}(\overline q_T,\mathcal D_\epsilon^{\rm ref})
\le \frac{|\St|+2s_{\rm ref}}{m_T}\to0.
\]
The uniform conditional MNL expansion in
Appendix~\ref{app:conditional-mle} and the delta method therefore give
\eqref{eq:end-to-end-inference}. Under the additional design-consistency
conditions, \eqref{eq:end-to-end-oracle} gives convergence of the largest
marginal variance to $\tau_\epsilon^\star(\thetastar,\eta)$ through the identity
$\max_k[\Omega_\Delta(\overline q_T)]_{kk}
=V_{\thetastar}(\overline q_T)$.
\end{proof}

\subsection{Ordinary Gaussian Limit Under Covariance Convergence}
\label{app:covariance-uniqueness}
Suppose, in addition to the conditional approximation in
Corollary~\ref{cor:end-to-end-inference}, that
$\Omega_\Delta(\overline q_T)\to_p\Omega_0$
for a deterministic positive semidefinite matrix $\Omega_0$. Then
\[
\sqrt{m_T}\{
\widehat{\boldsymbol\Delta}_T-
\boldsymbol{\Delta}(\thetastar)
\}
\Rightarrow
\mathcal N(0,\Omega_0).
\]
Indeed, continuity of Gaussian laws in their covariance matrices gives
$d_{\rm BL}\!\left(\mathcal N\{0,\Omega_\Delta(\overline q_T)\},\allowbreak
\mathcal N(0,\Omega_0)\right)\to_p0$.
Both this distance and the conditional approximation error are bounded,
so convergence in probability upgrades to convergence in $L^1$.
Conditioning and the triangle inequality then yield the stated
unconditional limit.

Under the design-consistency conditions in
Proposition~\ref{prop:pilot-design}, the covariance-convergence
condition holds if all allocations in
$\mathcal Q_\epsilon^\star(\thetastar,\eta)$ induce the same covariance
matrix. It also holds when the secondary selection rule makes
$\overline q_T\to_p q^\star$ for a deterministic
$q^\star\in\mathcal Q_\epsilon^\star(\thetastar,\eta)$, in which case
$\Omega_0=\Omega_\Delta(q^\star)$.

\par\endgroup

\addvspace{8pt}
\noindent\begin{minipage}{\linewidth}
\setlength{\parindent}{1em}
\subsection{Oracle Frontier Illustration and Additional MNL Examples}
\label{app:oracle-illustrations}
Figure~\ref{fig:oracle-frontier-schematic} illustrates the two oracle regimes in
Section~\ref{sec:overlay}. The two examples below illustrate how the choice of
experiment depends on information per unit of opportunity cost.

\begingroup
\setlength{\intextsep}{8pt plus 2pt minus 2pt}
\setlength{\textfloatsep}{10pt plus 2pt minus 2pt}
\begin{figure}[H]
\centering
\input{figures/oracle_frontier_schematic_tikz}
\caption{Schematic oracle frontiers between revenue and precision. The vertical axis measures
the largest local variance constant across the target contrasts under the best
allocation for each budget. Lower values indicate greater precision. Free
instances have a finite zero-budget intercept. When first-order inference is
feasible at a positive budget, costly instances have $F_\theta(0)=+\infty$ and
$F_\theta(\eta)=\Theta(1/\eta)$ as $\eta\downarrow0$.}
\label{fig:oracle-frontier-schematic}
\end{figure}
\endgroup

A unique non-target experiment can be optimal as well. With attractions $(1/10,1,1/20)$, revenues $(1,4/5,2/5)$ and capacity two, the optimal operating offer is $\{1,2\}$. To estimate $R(\{1,3\})-R(\{1\})$, the optimal small-budget costly offer is $\{2,3\}$: it omits the product with the highest per-sale revenue and belongs to neither target offer. Its information per cost is $1400/1681$, compared with approximately $0.1283$ for $\{1,3\}$.

Cost alone can also misrank experiments. With $v=(1,6,0.4)$, $r=(0.9,0.55,0.58)$ and capacity two, $B=\{1,2\}$ remains optimal. Offer $\{2,3\}$ costs less than $\{1,3\}$ ($0.04770$ versus $0.05333$), but the latter supplies more missing-coordinate information per cost ($2.604$ versus $1.072$).
\end{minipage}\par

\FloatBarrier

\begingroup
\setlength{\abovedisplayskip}{6pt plus 2pt minus 2pt}
\setlength{\belowdisplayskip}{6pt plus 2pt minus 2pt}
\setlength{\abovedisplayshortskip}{0pt plus 2pt}
\setlength{\belowdisplayshortskip}{4pt plus 2pt minus 1pt}
\section{Formal Local Benchmark and Full-Rank Likelihood Attainability}
\label{app:design-efficiency}

For the criterion in Appendix~\ref{app:sdp-technical}, we establish the local
information lower bound and full-rank likelihood attainability for contrast estimation.

\begin{theorem}[Local benchmark and full-rank attainability]\label{thm:local-design-bound}
Fix an interior parameter $\theta$ and deterministic design proportions $n_S/n\to q_S$.
\begin{enumerate}
\item[(A)] Under differentiability in quadratic mean (DQM) and local asymptotic normality (LAN), a contrast is pathwise regular at the root-$n$ scale only if $g_k(\theta)\in\operatorname{Range}\{M(q,\theta)\}$. For such a contrast, the local asymptotic minimax squared-error risk is at least $g_k(\theta)^\top M(q,\theta)^\dagger g_k(\theta)$, which defines the local information benchmark.
\item[(B)] Suppose $M(q,\theta)\succ0$ and the normalized negative log likelihood converges uniformly to a population criterion with a separated minimum at $\theta$. Suppose also that the normalized score satisfies a central limit theorem with covariance $M(q,\theta)$, and the normalized Hessian converges locally uniformly to a continuous population Hessian whose value at $\theta$ is $M(q,\theta)$. Then the fixed-penalty maximum-likelihood estimator and plug-in contrasts are asymptotically normal with covariances $M(q,\theta)^{-1}$ and $G_\Delta(\theta)M(q,\theta)^{-1}G_\Delta(\theta)^\top$, respectively. Each contrast-specific Gaussian limit has variance equal to the information benchmark in Part~(A).
\end{enumerate}
\end{theorem}

Let $\widehat\Delta_{k,n}$ be any estimator of $\Delta_k$ based on the
$n$ deterministic-design observations. Part~(A) of
Theorem~\ref{thm:local-design-bound} gives the local asymptotic minimax bound
\begin{equation}
\lim_{c\to\infty}\liminf_{n\to\infty}
\sup_{\substack{h\in\operatorname{Range}\{M(q,\theta)\}\\\|h\|_2\le c}}
\E_{\theta+h/\sqrt n}\!\left[n\{\widehat\Delta_{k,n}-\Delta_k(\theta+h/\sqrt n)\}^2\right]
\ge g_k(\theta)^\top M(q,\theta)^\dagger g_k(\theta).
\label{eq:local-info-lower}
\end{equation}
Let $\widehat\theta_{n,\lambda}$ be the fixed-penalty estimator in
\eqref{eq:estimators}, applied to these $n$ observations with fixed
$\lambda>0$. Under the conditions in Part~(B), its Gaussian limit is
\begin{align}
\sqrt n\{\widehat\theta_{n,\lambda}-\theta\}
&\Rightarrow\mathcal N\{0,M(q,\theta)^{-1}\},
\label{eq:full-rank-mle-clt}\\
\sqrt n\{\boldsymbol{\Delta}(\widehat\theta_{n,\lambda})
-\boldsymbol{\Delta}(\theta)\}
&\Rightarrow
\mathcal N\!\left(0,G_\Delta(\theta)M(q,\theta)^{-1}G_\Delta(\theta)^\top\right).
\label{eq:full-rank-contrast-clt}
\end{align}

\subsection{Possibly Singular Local-Information Lower Bound}\label{app:singular-local-bound}
Let $\Lambda_{S,j}(\theta)$ denote the log likelihood from the $j$th
observation collected under assortment $S$, and define the normalized
assortment-specific score sum
\begin{equation*}
\Xi_{n,S}
:=\frac1{\sqrt n}\sum_{j=1}^{n_S}
\nabla_\theta\Lambda_{S,j}(\theta).
\end{equation*}
For a fixed local direction $h$ and
$\theta_n=\theta+h/\sqrt n$, differentiability in quadratic mean gives
\[
\sum_{j=1}^{n_S}
\{\Lambda_{S,j}(\theta_n)-\Lambda_{S,j}(\theta)\}
=h^\top \Xi_{n,S}
-\frac12\frac{n_S}{n}
 h^\top\mathcal I_S(\theta)h+o_{\Prob_\theta}(1).
\]
The covariance of $\Xi_{n,S}$ is
$(n_S/n)\mathcal I_S(\theta)$, and independence across deterministic
design cells gives
$\sum_S\Xi_{n,S}\Rightarrow\mathcal N\{0,M(q,\theta)\}$.
Summing over $S$ and using $n_S/n\to q_S$ establishes LAN with
information $M(q,\theta)$ on the tangent space
$\mathcal T_q=\operatorname{Range}\{M(q,\theta)\}$.

Restricting local directions to $\mathcal T_q$ gives a reduced experiment
whose information is positive definite on $\mathcal T_q$. The local asymptotic
minimax theorem on this experiment gives
\eqref{eq:local-info-lower} for quadratic loss
\citep[Chapters~7--8]{vanDerVaart1998}. The convolution theorem also
implies that the limiting law of any regular estimator is the
convolution of a centered Gaussian component with variance
$g_k(\theta)^\top M(q,\theta)^\dagger g_k(\theta)$ and an independent
remainder. Whenever that limiting law has a finite second moment, its
variance is therefore no smaller than the displayed Gaussian variance.

If $g_k(\theta)\notin\mathcal T_q$, choose
$v\in\operatorname{Null}\{M(q,\theta)\}$ with
$g_k(\theta)^\top v\ne0$. A root-$n$ local perturbation along $v$ has a
degenerate first-order likelihood ratio but changes the contrast at the
root-$n$ scale. Hence the functional is not pathwise regular over the
full local model. This proves Part~(A) of
Theorem~\ref{thm:local-design-bound}.

\subsection{Full-Rank Attainability with a Fixed Penalty}\label{app:full-rank-attainability}
Assume Part~(B) of Theorem~\ref{thm:local-design-bound}, and write
$\ell_{S,j}=-\Lambda_{S,j}$ for the negative log likelihood of each
observation. Uniform convergence of the normalized negative log likelihood,
separation of its population minimum, and the vanishing normalized penalty
imply consistency. Since $\theta$ is interior, consistency makes the empirical
minimizer interior with probability approaching one. Let $\overline H_n$ be
the integrated normalized Hessian along the segment from $\theta$ to
$\widehat\theta_{n,\lambda}$. On that event, the score equation and the
integral mean-value identity give
\begin{align*}
\sqrt n\{\widehat\theta_{n,\lambda}-\theta\}
={}&-
\left\{\overline H_n+\frac{\lambda}{n}I_d\right\}^{-1}
\left\{
 \frac1{\sqrt n}\sum_{S,j}\nabla\ell_{S,j}(\theta)
 +\frac{\lambda\theta}{\sqrt n}
\right\}.
\end{align*}
The score central limit theorem and local uniform Hessian convergence
give, respectively, a
$\mathcal N\{0,M(q,\theta)\}$ limit for the normalized score and
$\overline H_n\to_pM(q,\theta)$. Full rank makes the limiting Hessian
invertible. The fixed-penalty terms vanish, so Slutsky's
theorem proves \eqref{eq:full-rank-mle-clt}. The delta method proves
\eqref{eq:full-rank-contrast-clt}.

\section{Estimator Conventions and Uniform Conditional MNL Expansion}
\label{app:conditional-mle}

Conditional on the pilot information that selects the post-pilot schedule,
the subsequent observations form a fixed-design MNL experiment. We establish
its likelihood expansion uniformly over schedules whose allocations approach
the stabilized design set, as required by Corollary~\ref{cor:end-to-end-inference}.

\begin{remark}[Estimator and boundary conventions]
\label{rem:estimator-convention}
In Algorithm~\ref{alg:sdp-overlay}, the estimator in
\eqref{eq:estimators} uses the post-pilot sample $I_T^{\rm post}$.
The pilot constructs $\widehat q_T$. Both the pilot and
post-pilot estimators use a fixed penalty $\lambda>0$. The penalty makes the
matrices in the finite-sample self-normalized bound invertible. For a sample of size $n$, the fixed penalty contributes $O(1/n)$ to the
normalized first-order condition of the summed likelihood and does not
change the full-rank first-order covariance.

The finite-sample containment result permits boundary solutions. The likelihood
CLTs, Wald approximation, and profile-likelihood limit use an interior truth
and an optimizer that is interior with probability approaching one.
\end{remark}
   
\begin{lemma}[Uniform conditional MNL expansion over stabilized schedules]
\label{lem:uniform-conditional-mle}
Let $\St$ be finite, let $\thetastar$ be an interior point of the bounded
log-attraction parameter box, suppose
$M(q^{\rm ref},\thetastar)\succ0$, and let the finite contrast vector
$\boldsymbol{\Delta}(\theta)$ be continuously differentiable near $\thetastar$. Fix
$\epsilon\in(0,1)$ and $\lambda>0$. For each $n$, let $\mathcal P_n$ be a sigma-field such that
the schedule $S_{n,1},\ldots,S_{n,n}$ is $\mathcal P_n$-measurable and
$\mathcal P_n$ contains none of the corresponding outcomes. Assume that,
conditionally on $\mathcal P_n$,
\[
\Prob_{\thetastar}
(Y_{n,j}=y\mid\mathcal P_n,Y_{n,1},\ldots,Y_{n,j-1})
=p_{\thetastar}(y\mid S_{n,j}).
\]
Write $\ell(\theta;y,S):=-\log p_\theta(y\mid S)$ for the negative
log likelihood. Let
\[
q_{n,S}:=\frac1n\sum_{j=1}^n\one{S_{n,j}=S},
\qquad
\widehat\theta_{n,\lambda}
\in\argmin_{\theta\in\Theta}
\left\{
\frac{\lambda}{2}\|\theta\|_2^2
+\sum_{j=1}^n\ell(\theta;Y_{n,j},S_{n,j})
\right\}.
\]
Here $q_n=(q_{n,S})_{S\in\St}$ is the empirical allocation of the
schedule. Suppose
$\operatorname{dist}(q_n,\mathcal D_\epsilon^{\rm ref})\to0$ in
probability. The information matrix $M(q_n,\thetastar)$ is then
nonsingular with probability approaching one. All inverse-based displays
below are understood on that event and may be defined arbitrarily on its
vanishing-probability complement. Then, for every $\zeta>0$,
\begin{equation}
\Prob_{\thetastar}\!\left(
\|\widehat\theta_{n,\lambda}-\thetastar\|_2>\zeta
\,\middle|\,\mathcal P_n
\right)\xrightarrow{p}0.
\label{eq:conditional-consistency}
\end{equation}
Moreover, with
$\xi_{n,j}:=\nabla_\theta\log
p_{\thetastar}(Y_{n,j}\mid S_{n,j})$,
\begin{equation}
\sqrt n\{\widehat\theta_{n,\lambda}-\thetastar\}
=
M(q_n,\thetastar)^{-1}
\frac1{\sqrt n}\sum_{j=1}^n\xi_{n,j}
+r_n,
\label{eq:uniform-conditional-expansion}
\end{equation}
where the remainder is conditionally negligible in the sense that, for
every $\zeta>0$,
\begin{equation}
\Prob_{\thetastar}\!\left[
\Prob_{\thetastar}
(\|r_n\|_2>\zeta\mid\mathcal P_n)>\zeta
\right]\longrightarrow0.
\label{eq:conditional-op}
\end{equation}
Define the contrast covariance map
$\Omega_\Delta(q):=G_\Delta(\thetastar)M(q,\thetastar)^{-1}
G_\Delta(\thetastar)^\top$. Write $d_{\rm BL}$ for the bounded-Lipschitz
distance between probability laws and
$\mathcal L_{\thetastar}[\cdot\mid\mathcal P_n]$ for the conditional law
under $\thetastar$. Then
\begin{align}
d_{\rm BL}\!\left(
\mathcal L_{\thetastar}\!\left[
\sqrt n\{\widehat\theta_{n,\lambda}-\thetastar\}
\,\middle|\,\mathcal P_n
\right],
\mathcal N\{0,M(q_n,\thetastar)^{-1}\}
\right)&\xrightarrow{p}0,
\label{eq:conditional-theta-kernel}\\
d_{\rm BL}\!\left(
\mathcal L_{\thetastar}\!\left[
\sqrt n\{\boldsymbol{\Delta}(\widehat\theta_{n,\lambda})
-\boldsymbol{\Delta}(\thetastar)\}
\,\middle|\,\mathcal P_n
\right],
\mathcal N\{0,\Omega_\Delta(q_n)\}
\right)&\xrightarrow{p}0.
\label{eq:conditional-delta-kernel}
\end{align}
\end{lemma}

\subsection{Proof}\label{app:conditional-mle-proof}

For $q\in\Delta(\St)$ define the conditional population criterion
$Q_q(\vartheta):=\sum_{S\in\St}q_S\E_{\thetastar}[\ell(\vartheta;Y,S)]$.
The Kullback--Leibler identity gives
\[
Q_q(\vartheta)-Q_q(\thetastar)
=\sum_{S\in\St}q_S
\KL\bigl(p_{\thetastar}(\cdot\mid S),p_\vartheta(\cdot\mid S)\bigr).
\]
Because $M(q^{\rm ref},\thetastar)\succ0$, every coordinate appears in
at least one assortment in $\operatorname{supp}(q^{\rm ref})$. Equality
of the MNL distributions on such an assortment identifies each active
coordinate from
$p_\vartheta(i\mid S)/p_\vartheta(0\mid S)=e^{\vartheta_i}$.
Consequently, $Q_{q^{\rm ref}}$ has the unique global minimizer
$\thetastar$. For every $q=(1-\epsilon)\widetilde q+
\epsilon q^{\rm ref}$,
\begin{equation}
Q_q(\vartheta)-Q_q(\thetastar)
\ge
\epsilon\{Q_{q^{\rm ref}}(\vartheta)-
Q_{q^{\rm ref}}(\thetastar)\}.
\label{eq:uniform-global-separation}
\end{equation}
Compactness and continuity therefore imply uniform separation away from
$\thetastar$ over $q\in\mathcal D_\epsilon^{\rm ref}$.

Conditional on $\mathcal P_n$, the post-pilot observations have the
product MNL law for the displayed deterministic schedule. Because the
action and outcome spaces are finite and the parameter box is compact,
the negative log likelihood is uniformly bounded and Lipschitz in the
parameter. A conditional uniform law of
large numbers therefore holds uniformly over all deterministic schedules,
while
$\lambda\|\vartheta\|_2^2/(2n)$ converges uniformly to zero. Together
with \eqref{eq:uniform-global-separation} and
$\operatorname{dist}(q_n,\mathcal D_\epsilon^{\rm ref})\to0$, the
argmin theorem gives \eqref{eq:conditional-consistency}.

For MNL, the negative-log-likelihood Hessian depends on the offered
assortment and parameter. It is the same for every realized choice. Hence, writing
\[
H_n(\vartheta)
:=\frac1n\sum_{j=1}^n
\nabla^2_\theta\ell(\vartheta;Y_{n,j},S_{n,j})
=\sum_{S\in\St}q_{n,S}\mathcal I_S(\vartheta),
\]
continuity, compactness, and
$M(q^{\rm ref},\thetastar)\succ0$ imply the uniform eigenvalue floor
\[
\inf_{q\in\mathcal D_\epsilon^{\rm ref}}
\lambda_{\min}\{M(q,\thetastar)\}
\ge
\epsilon\lambda_{\min}\{M(q^{\rm ref},\thetastar)\}>0.
\]
The same lower bound, up to a factor $1/2$, holds for $q_n$ and parameters
in a shrinking neighborhood of $\thetastar$ with probability approaching
one. Conditional consistency and the interiority of $\thetastar$ imply
that the conditional probability of an interior estimator converges to
one in probability. Define the integrated Hessian
$\overline H_n:=\int_0^1
H_n\!\left(\thetastar+u\{\widehat\theta_{n,\lambda}-\thetastar\}\right)\,du$.
On the event that the estimator is interior, the score equation and the
integral mean-value identity give
\begin{align*}
\sqrt n\{\widehat\theta_{n,\lambda}-\thetastar\}
={}&
\left\{\overline H_n+\frac{\lambda}{n}I_d\right\}^{-1}
\left\{
\frac1{\sqrt n}\sum_{j=1}^n\xi_{n,j}
-\frac{\lambda\thetastar}{\sqrt n}
\right\}.
\end{align*}
Uniform continuity of the information matrices gives
$\overline H_n-M(q_n,\thetastar)=o_p(1)$ conditionally in probability.
The normalized score sum has a uniformly bounded conditional second
moment, so it is conditionally tight. The eigenvalue floor permits
replacing the inverse Hessian in the score equation by
$M(q_n,\thetastar)^{-1}$, with a conditionally negligible error.
The two penalty terms vanish. This proves
\eqref{eq:uniform-conditional-expansion}--\eqref{eq:conditional-op}.

Conditionally on $\mathcal P_n$, the normalized score sum
$n^{-1/2}\sum_{j=1}^n\xi_{n,j}$ has mean zero and covariance
$M(q_n,\thetastar)$. The individual score vectors are uniformly bounded,
so the Lindeberg condition holds uniformly over deterministic schedules.
If the Gaussian approximation were not uniform over schedules whose
allocations approach $\mathcal D_\epsilon^{\rm ref}$, there would be a
sequence of deterministic schedules with approximation errors bounded
away from zero. Compactness of the allocation simplex gives a subsequence
with $q_n\to q\in\mathcal D_\epsilon^{\rm ref}$. The multivariate
triangular-array CLT applies along this subsequence, and continuity of
Gaussian laws in their covariance matrices contradicts the violation.
The distance condition lets us apply this uniform approximation to the
$\mathcal P_n$-measurable random schedule. Combining it with
\eqref{eq:uniform-conditional-expansion} proves
\eqref{eq:conditional-theta-kernel}.

The score expansion and eigenvalue floor also make the scaled estimation
error conditionally tight. Continuous differentiability of the finite
contrast vector then makes the scaled Taylor remainder conditionally
negligible. The delta method gives \eqref{eq:conditional-delta-kernel}.

\section{Finite-Sample Consequences and Practical Interval Constructions}\label{app:finite-sample-consequences}
\label{app:interval-proofs}

This appendix uses the general finite-sample theorem to derive bounds for
revenue contrast estimates. It then applies these bounds to rounded MNL
designs and presents two interval constructions: finite-sample intervals
for singleton observations and asymptotic profile-likelihood intervals
for general assortments.

\subsection{Consequences of the General Finite-Sample Theorem}\label{app:general-consequences}
\begin{proposition}[From high-probability bounds to expected objectives]\label{prop:hp2exp}
Assume that each reported contrast estimate takes values in $[-1,1]$.
Suppose a design pair $(\pi,\widehat\Delta)$ satisfies, uniformly over $\theta\in\Theta_0$,
\[
\Prob_\theta\!\left(\Reg_\pi(T,\theta) \le B_R(T,\delta)\right) \ge 1-\delta,
\qquad
\Prob_\theta\!\left(\max_{(a,b)\in\Kcal}\abs{\widehat\Delta^{(a,b)}-\Delta_{R,\theta}^{(a,b)}} \le B_E(T,\delta)\right) \ge 1-\delta.
\]
Then the worst-case expected objectives defined in
Section~\ref{subsec:rate-criteria} satisfy
\[
\mathcal{R}_T(\pi) \le B_R(T,\delta) + \delta T,
\qquad
\mathcal{E}_T(\widehat\Delta) \le B_E(T,\delta) + 2\delta.
\]
In particular, taking $\delta=T^{-2}$ converts the high-probability bound of
Theorem~\ref{thm:main} into an expected-objective statement without changing
the polynomial rate in $T$.
\end{proposition}

\begin{proof}
Regret is always bounded by $T$ because each round contributes at most $1$. Also, each revenue contrast lies in $[-1,1]$, so the absolute error of any contrast estimator taking values in $[-1,1]$ is at most $2$. Split each expectation over the high-probability event and its complement.
\end{proof}

\subsection{Directional Consequence for Rounded MNL Designs}\label{app:rounded-directional}
The linear approximation converts parameter error into contrast error through the gradient at the pilot reference. The next bound also accounts for error in that reference and for the nonlinear remainder. These terms explain how the local design criterion enters a finite-sample contrast bound.

\begin{proposition}[Directional consequence for rounded MNL designs]
\label{prop:sdp-directional-consequence}
Assume the conditions of Corollary~\ref{cor:sdp-finite-sample-bridge}
and use its rounded schedule and notation.
Here $\Xi_T(\bar\theta)$ is the rounded contrast variance criterion and
$m_{\rm MNL}$ is the curvature constant from Proposition~\ref{prop:mnlproxies}.
If $\bar\theta\in\Theta$, $\|\bar\theta-\thetastar\|_2\le r_p$ for some $r_p\ge0$ and
all target gradients are $L_g$-Lipschitz on $\Theta$, then on the post-pilot inference event in
Theorem~\ref{thm:main}, with
$\widehat\Delta_{T,k}:=\Delta_k(\thetatilde_T)$ and
$r_T^{\rm exp}:=\beta_T^{\mathrm{exp}}(\delta/2)/
\sqrt{\lambda_{\min}(V_T^{\mathrm{exp}})}$,
\begin{align}
\max_{k\in\Kcal}
\left|\widehat\Delta_{T,k}-\Delta_k(\thetastar)\right|
&\le
\beta_T^{\mathrm{exp}}(\delta/2)
\sqrt{\frac{\Xi_T(\bar\theta)}{m_{\rm MNL}}}
\notag\\
&\quad
+L_g\left(r_p r_T^{\rm exp}+\frac12(r_T^{\rm exp})^2\right).
\label{eq:sdp-finite-sample-contrast}
\end{align}
If $m_T>C_{\mathcal H}/\kappa$, \eqref{eq:sdp-directional-bridge}
bounds the variance factor in the first term.
\end{proposition}
\begin{proof}[Proof of Corollary~\ref{cor:sdp-finite-sample-bridge}
and Proposition~\ref{prop:sdp-directional-consequence}]
For every assortment, define
$D_S:=\Diag(\one{i\in S})_{i=1}^N$. Then
$\mathcal I_S(\bar\theta)\preceq D_S$, while
Proposition~\ref{prop:mnlproxies} uses the corresponding curvature proxy
$m_{\rm MNL}D_S$. Summing over the post-pilot rounded schedule
proves \eqref{eq:sdp-curvature-bridge}. Inversion reverses Loewner order,
which gives \eqref{eq:sdp-ex-post-bridge}. The multiplicative statement
then follows from Theorem~\ref{thm:design-rounding}.

Put $e=\thetatilde_T-\thetastar$. Theorem~\ref{thm:main} gives
$\norm{e}_{V_T^{\rm exp}}\le\beta_T^{\rm exp}(\delta/2)$ and
$\norm{e}_2\le r_T^{\rm exp}$. Cauchy--Schwarz and
\eqref{eq:sdp-ex-post-bridge} bound $|g_k(\bar\theta)^\top e|$ by
the first term in \eqref{eq:sdp-finite-sample-contrast}. The integral
mean-value identity and gradient Lipschitzness give, for every $k$,
\[
\abs{\Delta_k(\thetastar+e)-\Delta_k(\thetastar)-g_k(\bar\theta)^\top e}
\le L_g\left(r_p\norm{e}_2+\frac12\norm{e}_2^2\right).
\]
Combining these bounds proves \eqref{eq:sdp-finite-sample-contrast}.
\end{proof}

For each executed schedule we record the rank and minimum eigenvalue of
$M_{m_T}(\bar\theta)$, the threshold $C_{\mathcal H}/\kappa$, and the
exact ex post certificate $\Xi_T(\bar\theta)$.
\subsection{Practical Interval Statements}\label{app:interval-statements}
With fixed singleton offer counts, each product's purchase count is
binomial. We construct a simultaneous attraction box from binomial
confidence intervals and obtain contrast intervals by optimizing over
that box.

\begin{corollary}[Exact finite-sample singleton interval]
\label{cor:exact-singleton-ci}
Write $v^\star=(e^{\theta_1^\star},\ldots,e^{\theta_N^\star})$ for the
true attraction vector. For any attraction vector $v$, write
$R_v(S):=\sum_{i\in S}r_iv_i/(1+\sum_{j\in S}v_j)$ and
$\Delta_{R,v}^{(a,b)}:=R_v(S_a)-R_v(S_b)$.
Let $C_i$ be the number of purchases among the $n_i\ge1$ planned
offers of singleton $\{i\}$. The counts are fixed before the outcomes entering
$C_i$ are observed. They may be selected using an initial pilot that excludes
those outcomes. Construct a two-sided
Clopper--Pearson interval
$[\underline p_i,\overline p_i]$ for the singleton purchase
probability $p_i^\star=v_i^\star/(1+v_i^\star)$ with marginal error probability
$\delta/N$, and define
\begin{equation*}
\mathcal V_i(\delta):=
\left[\frac{\underline p_i}{1-\underline p_i},
      \frac{\overline p_i}{1-\overline p_i}\right]
\cap[v_{\min},v_{\max}],
\end{equation*}
with the upper endpoint interpreted as $+\infty$ if
$\overline p_i=1$.
   
If every coordinate intersection is nonempty, let
$\mathcal V(\delta):=\prod_{i=1}^N\mathcal V_i(\delta)$ and report the
contrast interval in \eqref{eq:exact-contrast-ci}. If any coordinate
intersection is empty, define the reported interval for every contrast to
be the fallback interval $[-1,1]$. Then the procedure is defined for every
sample realization and, with probability at least $1-\delta$,
$v^\star\in\mathcal V(\delta)$, no fallback is triggered, and,
simultaneously for every $k=(a,b)\in\Kcal$,
\begin{equation}
\Delta_{R,v^\star}^{(a,b)}\in
\left[
\min_{v\in\mathcal V(\delta)}\Delta_{R,v}^{(a,b)},
\max_{v\in\mathcal V(\delta)}\Delta_{R,v}^{(a,b)}
\right].
\label{eq:exact-contrast-ci}
\end{equation}
   
\end{corollary}

\begin{proof}\label{app:proof-exact-singleton}
Condition on any initial pilot used to select the fixed counts. On a singleton
offer, the purchase indicator is Bernoulli with
probability $p_i^\star=v_i^\star/(1+v_i^\star)$. Exact binomial coverage and a union
bound give simultaneous containment of all $p_i^\star$ with probability at
least $1-\delta$. The map $p\mapsto p/(1-p)$ is increasing, and
Assumption~\ref{ass:mnlbounded} supplies the intersection with the
prespecified attraction box.
   
On the simultaneous coverage event, every intersection is nonempty and
contains $v_i^\star$, so the fallback is not triggered. Optimizing each
contrast over the resulting simultaneous box proves
\eqref{eq:exact-contrast-ci}. On any realization for which a coordinate
intersection is empty, the fallback $[-1,1]$ is well defined and contains
every feasible revenue contrast.
   
\end{proof}

The endpoints in \eqref{eq:exact-contrast-ci} solve box-constrained
optimization problems for the contrast. For a coordinatewise monotone
contrast, the extrema occur at the appropriate box corners. Otherwise,
certified branch-and-bound can provide bounds on the global minimum and
maximum. Using certified outer bounds preserves finite-sample coverage.

For general assortments, we use the likelihood of the observed choices
to construct an interval for each contrast. The following result gives
its asymptotic coverage.

\begin{proposition}[Profile-likelihood interval for an arbitrary design]
\label{prop:profile-likelihood}
Consider any deterministic assortment schedule with
\(n_S/n\to q_S\) and \(M(q,\thetastar)\succ0\). Let
\[
L_n(\theta):=-\sum_{t=1}^n\log p_\theta(Y_t\mid S_t),
\qquad
\widehat\theta_n\in\argmin_{\theta\in\Theta}L_n(\theta).
\]
For a scalar contrast \(\Delta_k\) and
\(\delta_{\rm PL}\in(0,1)\), define
\begin{equation*}
\mathcal C_{k,1-\delta_{\rm PL}}^{\rm PL}:=
\left\{z:
2\!\left[
\inf_{\theta\in\Theta:\,\Delta_k(\theta)=z}L_n(\theta)
-L_n(\widehat\theta_n)
\right]
\le\chi^2_{1,1-\delta_{\rm PL}}
\right\}.
\end{equation*}
   
Here $\inf\varnothing=+\infty$. Assume additionally that $\thetastar$
is interior, the local MNL model is identifiable, $g_k(\thetastar)\ne0$,
and the unconstrained optimizer and the constrained optimizer at
$z=\Delta_k(\thetastar)$ are interior with probability approaching one.
The likelihood-ratio calibration then gives
\(\Prob_{\thetastar}\{\Delta_k(\thetastar)\in
\mathcal C_{k,1-\delta_{\rm PL}}^{\rm PL}\}
\to1-\delta_{\rm PL}\). Replacing \(\delta_{\rm PL}\)
by \(\delta_{\rm PL}/|\Kcal|\) gives asymptotic simultaneous coverage
of at least $1-\delta_{\rm PL}$ by Bonferroni. If $g_k(\thetastar)=0$ or a boundary constraint is active, the
ordinary $\chi_1^2$ calibration is not asserted.
   
\end{proposition}

\begin{proof}\label{app:proof-profile-likelihood}
The deterministic proportions imply an independent triangular-array
likelihood with normalized information converging to
\(M(q,\thetastar)\). Full rank, local identifiability, interiority, and the nonzero
constraint gradient give the usual local quadratic likelihood expansion.
The likelihood-ratio statistic for the regular one-dimensional constraint
\(\Delta_k(\theta)=\Delta_k(\thetastar)\) therefore converges to
\(\chi_1^2\) by Wilks' theorem
\citep[Chapter~16]{vanDerVaart1998}. Inverting the test proves marginal
coverage, and the union bound proves the simultaneous statement.
\end{proof}

For numerical inversion, the profile interval can equivalently be computed as
the projection of the likelihood-ratio region. Its lower and upper endpoints
solve
\[
\min_{\theta\in\Theta}\Delta_k(\theta)
\quad\text{and}\quad
\max_{\theta\in\Theta}\Delta_k(\theta)
\quad\text{subject to}\quad
2\{L_n(\theta)-L_n(\widehat\theta_n)\}
\le\chi^2_{1,1-\delta_{\rm PL}}.
\]
Global endpoint optimality requires a separate optimization certificate.

\par\endgroup

\section{Expedia Calibration, Implementation, and Additional Results}\label{app:sdp-numerics}
\begingroup
\setlength{\floatsep}{6pt plus 2pt minus 1pt}
\setlength{\textfloatsep}{6pt plus 2pt minus 1pt}
\setlength{\intextsep}{6pt plus 2pt minus 1pt}
\subsection{Calibration and Fixed Simulation Inputs}\label{app:expedia-calibration}
The 9,917,530 displayed-property rows form 399,344 searches with at most one booking each. SplitMix64 residues modulo five give calibration (0--2), validation (3), and test (4) splits. We screen up to thirty destinations by descending calibration-query count. Eligibility requires five hotels with at least ten calibration bookings each, at least 1,000 retained calibration queries, and at least 100 retained validation queries. All four eligible markets are included. We order hotels by calibration bookings, breaking ties by identifier. The booking-enriched sample defines the calibration population. Table~\ref{tab:expedia-calibration-samples} compares test losses on identical menus for models trained on the same split. The supplementary manifest indexes full-precision inputs, code, and replication records.

\begin{table}[!htbp]\SingleSpacedXI\centering\small
\caption{Historical searches and mean test log loss. The anchor is the selected hotel with the highest normalized calibration median price. The loss difference is the shared-attraction loss minus the hotel-specific loss on the displayed test menus.}
\label{tab:expedia-calibration-samples}
\begin{tabular}{rrrrrrrrr}\toprule
& & & & & & \multicolumn{3}{c}{Test log loss}\\\cmidrule(l){7-9}
Market & Anchor & Retained & Calib. & Valid. & Test & Hotel-specific & Shared & Difference\\\midrule
4562 & 53494 & 2501 & 1539 & 497 & 465 & 0.5820 & 0.5879 & 0.0059\\
8192 & 116942 & 4691 & 2870 & 916 & 905 & 0.8759 & 0.8812 & 0.0053\\
9402 & 35679 & 1976 & 1172 & 409 & 395 & 0.6366 & 0.6462 & 0.0095\\
13870 & 106230 & 1720 & 1055 & 323 & 342 & 0.7132 & 0.7156 & 0.0024\\
\bottomrule\end{tabular}\end{table}

Before simulation, we fix the shared attraction box $[v_{\min},v_{\max}]^5$ for each case, with $v_{\min}=\min_i v_i/5$ and $v_{\max}=\max\{1.2,5\max_i v_i\}$.

\subsection{Pilot Certificates, Designs, and Intervals}\label{app:expedia-implementation}
For each singleton success probability, a Clopper--Pearson interval with tail probability $0.005$ per side is transformed to attractions and intersected with the fixed box. The coordinatewise hull with the bounded ridge pilot estimate gives the certificate box. Its farthest-corner radius is $\rho_{p,T}$, and a certified upper bound $\overline L_{p,T}$ on the revenue-gradient norm gives $\gamma_{p,T}=2\overline L_{p,T}\rho_{p,T}$. We also use these boxes for the certificate in Proposition~\ref{prop:box-certificate}. Empty certificates, infeasible tightened families, and failed information checks trigger the reference fallback. At the reference ceiling, the single-target comparison has no observed fallback or fractional or rounding-adjusted violation for the tightened SDP and no observed violation for the nominal SDP.

We use reference-first rounding. Single-target integer matches retain 5,000 offers and at least 50 per singleton. Independent linear-oracle checks verify single-target D-optimal gaps below $2.0\times10^{-5}$.

Untruncated Wald intervals use the unpenalized likelihood information and contrast gradient at the fit. Post-pilot fits condition on the pilot-selected design. All-planned refits and their coverage are empirical diagnostics. We retain all observations and boundary fits.

The single-target simulations use seed 20261107. The original protocol designated all-planned MAE as the primary outcome. The common-budget and actual-cost analyses use 5,000 replication-block draws with seeds 20261108 and 20261109, respectively.

For the single-target comparisons, individual tests use two-sided paired $t$-tests of replication-level absolute errors. Holm correction is applied across the sixteen targets separately for each comparator, sample convention, and inference penalty.

For each joint comparator and estimator, 5,000 paired block resamples calibrate simultaneous bounds for 32 target MAE means across four markets and two methods. We propagate them through the worst-target maxima and market-relative reductions. Joint Wald intervals use $z_{1-0.05/8}$.

\subsection{Separately Optimized Single-Target Results}\label{app:expedia-additional-results}
Let $\bar\theta_b$ be the pilot estimate in replication $b$. The six single-target baselines are balanced singletons, the full oracle SDP at the calibrated parameter with a reference floor, nominal SDP with estimated costs and no tightening, Pilot-best+reference with remaining traffic on the pilot revenue maximizer, a pilot-matched mixture with the reference share set to match the tightened design's estimated cost when attainable, and D-optimal maximizing $\log\det M(q,\bar\theta_b)$ over the tightened family.

The supplementary common-budget results cover all six baselines. All-planned MAE intervals against the oracle and nominal SDPs include zero. Against nominal SDP, tightening raises post-pilot MAE by $2.89\%$ and lowers total cost by $7.53\%$. Against Pilot-best+reference, it lowers all-planned MAE by $16.45\%$ at $6.08\%$ higher total cost.

Table~\ref{tab:expedia-cost-matched} gives the matched-cost results by market. The largest single-target integer cost discrepancy is $9.9930\times10^{-7}$ relative to the target cost.

\begin{table}[!htbp]\SingleSpacedXI
\centering\small
\setlength{\tabcolsep}{3pt}
\caption{Tightened pilot SDP relative to D-optimal at matched post-pilot opportunity cost ($\lambda=1$). Entries are equal-weight mean target-relative MAE reductions, in percent. 95\% intervals use joint replication-block resampling within fixed markets. Holm counts adjust sixteen comparisons separately for each sample convention. Both methods share the pilot and each uses 5,000 post-pilot observations.}
\label{tab:expedia-cost-matched}
\begin{tabular}{lrrrrrr}\toprule
& \multicolumn{3}{c}{\textit{Post-pilot estimation}} & \multicolumn{3}{c}{\textit{Empirical all-planned refit}}\\
\cmidrule(lr){2-4}\cmidrule(lr){5-7}
Market & Reduction & 95\% interval & Holm & Reduction & 95\% interval & Holm\\\midrule
4562 & 22.19 & [18.43, 25.58] & 4/4 & 7.29 & [3.56, 10.59] & 0/4\\
8192 & 24.96 & [21.37, 28.21] & 4/4 & 12.72 & [8.61, 16.31] & 1/4\\
9402 & 28.18 & [24.65, 31.46] & 4/4 & 14.28 & [10.66, 17.65] & 3/4\\
13870 & 21.85 & [18.17, 25.26] & 4/4 & 11.19 & [7.35, 14.65] & 3/4\\
\midrule
All sixteen targets & 24.29 & [22.48, 25.86] & 16/16 & 11.37 & [9.47, 13.03] & 7/16\\
\bottomrule\end{tabular}
\end{table}

\FloatBarrier
\paragraph{Sensitivity to inference regularization.}\label{subsec:expedia-regularization}
With pilot estimates, allocations, integer counts, and choices fixed, we refit over the same bounds using negative log likelihood plus $\lambda\|\theta\|_2^2/2$ for $\lambda\in\{0,0.1,1\}$. The refits in Table~\ref{tab:expedia-regularization} use the original common-ceiling samples, including D-optimal under the pilot-tightened budget. The separately reoptimized cost-matched comparison uses $\lambda=1$.

\begin{table}[!htbp]\SingleSpacedXI
\centering\small
\setlength{\tabcolsep}{4pt}
\caption{Sensitivity to inference regularization with allocations and samples fixed.
Entries are mean target-relative reductions in all-planned MAE, in percent.
The D-optimal comparator uses the pilot-tightened budget.
Pilot fits retain $\lambda=1$. The last column gives the stratified paired bootstrap interval for the D-optimal comparison.}
\label{tab:expedia-regularization}
\begin{tabular}{rrrrr}\toprule
$\lambda$ & \shortstack{Pilot-best\\+reference} & \shortstack{Pilot-matched\\mixture} & \shortstack{D-optimal\\(pilot-tightened)} & \shortstack{D-optimal\\95\% interval}\\\midrule
0 & 17.35 & 18.10 & 12.26 & [10.38, 13.98]\\
0.1 & 17.25 & 18.02 & 12.22 & [10.34, 13.94]\\
1 & 16.45 & 17.41 & 11.97 & [10.10, 13.69]\\
\bottomrule\end{tabular}
\end{table}

\FloatBarrier
\subsection{Target-Level Coverage and Cost Accounting}\label{app:expedia-target-diagnostics}
Table~\ref{tab:joint-core} gives the shared pilot costs. The tightened SDP and cost-matched D-optimal have zero boundary rates for all sixteen targets with the post-pilot estimator and the all-planned refit. Supplementary files give target errors, exact coverage intervals, interval lengths, paired tests, and matched costs. At $\lambda=1$, mean all-planned MAE reductions against cost-matched D-optimal are $10.97\%$ across eight oracle-free targets and $11.77\%$ across eight costly targets, with 95\% bootstrap intervals $[8.34\%,13.38\%]$ and $[9.12\%,14.22\%]$, respectively.

\subsection{Oracle Frontier Computation and Complete Results}\label{app:expedia-oracle-frontiers}
The zero-floor oracle uses the same 25 assortments for all sixteen targets. The grid is zero and $\eta/\eta_{\rm ref}\in\{a10^k:a\in\{1,2,5\},\ k=-6,\ldots,-1\}\cup\{1\}$. Eight zero-cost values are finite. The costly slopes range from $-0.9999998$ to $-0.9999924$. Table~\ref{tab:expedia-oracle-frontiers} maps Figure~\ref{fig:expedia-oracle-frontiers} and compares reference floors for the eight prespecified pilot targets in Appendix~\ref{app:pilot-results}. Only the costly targets have the positive coefficient in Theorem~\ref{thm:exact-frontier}.

The maximum relative gap across 304 zero-floor solves is $7.10\times10^{-6}$. Enumeration verifies Panel A's paid offers to $2.60\times10^{-12}$ in the projected coefficient. Of 288 positive-floor cells, 96 are infeasible, 16 assign all unreserved offers to the unique revenue maximizer at $x=\epsilon$, and 176 pass independent numerical checks. We rescale missing information directions. Supplementary certificates give bounds, condition numbers, and allocations.

\begin{table}[!htbp]\SingleSpacedXI\centering\small
\setlength{\tabcolsep}{3pt}
\caption{Oracle targets and the cost of reference stabilization. Panel A gives the line mapping, exact leading coefficient and paid offer for the zero-floor frontier. Panel B gives envelopes of certified $\tau_\epsilon^\star/F_\theta$ intervals across four targets per regime. Here $x=\eta/\eta_{\rm ref}$, and the gap is the largest relative primal--dual gap across both oracle values in each feasible row. The constraint is infeasible when $x<\epsilon$.}
\label{tab:expedia-oracle-frontiers}\label{tab:expedia-exact-coefficients}
\begin{tabular}{rrllrrl}\toprule
\multicolumn{7}{l}{\textit{Panel A. Zero-floor frontier}}\\
Market & Hotel & Regime & Line & $F(0)$ & $10^4L_\theta$ & Optimal paid offer\\\midrule
4562 & 61529 & Free & Solid & 0.035327 & --- & ---\\
4562 & 79740 & Free & Dashed & 0.033463 & --- & ---\\
4562 & 21155 & Costly & Dashed & $\infty$ & 0.313835 & $\{53494,79740,21155\}$\\
4562 & 14536 & Costly & Solid & $\infty$ & 1.356274 & $\{53494,79740,14536\}$\\
8192 & 77089 & Costly & Solid & $\infty$ & 1.017735 & $\{116942,77089,104517\}$\\
8192 & 37818 & Free & Solid & 0.025058 & --- & ---\\
8192 & 134154 & Costly & Dashed & $\infty$ & 1.160663 & $\{116942,134154,104517\}$\\
8192 & 104517 & Free & Dashed & 0.032628 & --- & ---\\
9402 & 34151 & Free & Dashed & 0.060385 & --- & ---\\
9402 & 25159 & Free & Solid & 0.060192 & --- & ---\\
9402 & 67964 & Costly & Dashed & $\infty$ & 4.411856 & $\{34151,25159,67964\}$\\
9402 & 8864 & Costly & Solid & $\infty$ & 1.379202 & $\{34151,25159,8864\}$\\
13870 & 87490 & Free & Solid & 0.026821 & --- & ---\\
13870 & 120248 & Free & Dashed & 0.052799 & --- & ---\\
13870 & 7377 & Costly & Solid & $\infty$ & 0.885132 & $\{120248,7377,106230\}$\\
13870 & 57125 & Costly & Dashed & $\infty$ & 2.143384 & $\{120248,106230,57125\}$\\
\bottomrule\end{tabular}
\par\smallskip
\begin{tabular}{rrrrrr}\toprule
\multicolumn{6}{l}{\textit{Panel B. Stabilized-to-unrestricted oracle ratios}}\\
$x$ & $\epsilon$ & $\epsilon/x$ & Free & Costly & Max. rel. gap\\\midrule
$10^{-6}$ & $0.05x^2$ & $5\times10^{-8}$ & $[1,1+1.24\times10^{-8}]$ & $[1,1+1.06\times10^{-7}]$ & $8.60\times10^{-8}$\\
$0.01$ & $0.05$ & $5$ & Infeasible & Infeasible & ---\\
$0.01$ & $0.01$ & $1$ & $1.009$--$1.011$ & $58.063$--$207.276$ & $1.04\times10^{-6}$\\
$0.05$ & $0.05$ & $1$ & $1.047$--$1.053$ & $56.802$--$88.751$ & $4.53\times10^{-8}$\\
$0.10$ & $0.01$ & $0.1$ & $1.009$--$1.010$ & $1.009$--$1.011$ & $7.99\times10^{-9}$\\
$0.10$ & $0.05$ & $0.5$ & $1.046$--$1.052$ & $1.046$--$1.467$ & $4.43\times10^{-9}$\\
\bottomrule\end{tabular}\end{table}

\paragraph{A joint two-direction limit.}\label{app:joint-projected-limit}
In market 9402, the joint optimum equalizes the two costly projected variances at $L_\theta=5.791058\times10^{-4}$. The limiting cost shares on $\{34151,25159,67964\}$ and $\{34151,25159,8864\}$ are $0.7618394$ and $0.2381606$, respectively. Pricing equality holds to relative tolerance $3.5\times10^{-9}$.

\FloatBarrier
\subsection{Joint Four-Target Designs}\label{app:joint-results}
At true-cost ratio one, the stabilized joint oracle lowers the variance criterion by $34.62$--$40.23\%$ against Restricted. All available joint simulation cells have zero boundary, fallback and strict-violation frequencies. Matching uses calibrated costs and relative tolerance $10^{-6}$, with maximum discrepancy $9.01\times10^{-7}$. The Full--Restricted matched-cost comparison uses 1,998 pairs. Restricted could not match Full's cost in replication 442 in market 9402 and replication 224 in market 13870. The matched pools reoptimize components at cost equality. Spend-cap D-optimal can spend less than Full. After refinement, independent linear-oracle checks certify concavity gaps below $5\times10^{-5}$. Supplementary records give feasibility, pricing results, scores and weights at all three budgets.

Post-pilot joint Wald coverage for Full ranges from $95.6\%$ to $96.6\%$ across the four markets at the reference ceiling. Table~\ref{tab:joint-contracts} gives the MAE reductions and intervals. Supplementary results give market-level errors, costs, and coverage.
\begin{table}[!htbp]\SingleSpacedXI\centering\small
\setlength{\tabcolsep}{3.5pt}
\caption{Worst-target MAE reductions for Full relative to each comparator, averaged across the four markets, with propagated 95\% intervals (percent). $^a$ Restricted matching uses 1,998 available market-replications. Other comparisons use 2,000. The spend cap equals Full's actual post-pilot opportunity cost.}
\label{tab:joint-contracts}
\begin{tabular}{llrrrr}\toprule
Contract & Comparator & Post gain & 95\% interval & All gain & 95\% interval\\\midrule
Ceiling & Restricted & 26.84 & [8.98, 41.14] & 10.34 & [-10.96, 27.56]\\
Ceiling & Equal pool & 33.51 & [17.12, 46.61] & 10.85 & [-9.97, 27.76]\\
Ceiling & Optimized pool & 24.34 & [6.14, 38.99] & 10.03 & [-11.74, 27.51]\\
Ceiling & D-optimal & 16.74 & [-3.58, 33.06] & 11.30 & [-9.74, 28.30]\\
Match & Restricted$^a$ & 41.98 & [27.71, 53.36] & 8.04 & [-13.63, 25.58]\\
Match & Equal pool & 26.96 & [8.64, 41.48] & 7.08 & [-14.89, 24.84]\\
Match & Optimized pool & 18.23 & [-1.29, 33.95] & 4.58 & [-17.96, 22.79]\\
Match & D-optimal & 11.11 & [-10.43, 28.41] & 5.25 & [-16.99, 23.25]\\
Spend cap & D-optimal: spend cap & 14.54 & [-5.96, 31.06] & 6.45 & [-16.52, 24.82]\\
\bottomrule\end{tabular}\end{table}
\FloatBarrier

\subsection{Pilot Size, Total Cost, and Budget Stress}\label{app:pilot-results}
The grids in Section~\ref{subsec:expedia-pilot} fix total or post-pilot traffic. Relative ceilings are $0.5,1,1.5$ times $\eta_{\rm ref}$. Absolute ceilings are $0.03,0.06,0.12$ at $p_T\in\{1{,}000,20{,}000\}$. Free/costly pairs are $(61529,14536)$, $(37818,77089)$, $(25159,8864)$, $(87490,7377)$. We retain all fallbacks and count each shared pilot once per market for feasibility. Box alone is feasible at $0.5\eta_{\rm ref}$ with 20,000 pilot offers, and is $99.9\%$ feasible at $1.5\eta_{\rm ref}$ with 2,000.

At $p_T=5{,}000$ with fixed total traffic and ceiling $\eta_{\rm ref}$, Box lowers mean target-relative total cost by $49.75\%$ against nominal SDP, with $2.68\%$ higher post-pilot MAE. At $p_T=1{,}000$, both certificates fall back at all three absolute ceilings, with strict violation rates of $100\%$, $100\%$, and $25\%$. At $p_T=20{,}000$, both certificates fall back and overspend at $\eta=0.03$, only Box is feasible at $\eta=0.06$, and both are feasible at $\eta=0.12$. At $\eta=0.06$, nominal strict and rounding-adjusted violation rates are $31.93\%$ and $7.03\%$. Box has zero observed violations.

Figure~\ref{fig:pilot-total-cost} shows fixed-total results at the reference ceiling. The supplementary pilot tables give the means, fixed-post-pilot results, coverage and cost accounting. The absolute grid is in \emph{pilot\_absolute\_summary.csv}.

\subsection{Independent Validation and Choice-Model Misspecification}\label{app:misspecification-results}
We mix MNL parameters $\theta\pm\delta u$ equally, with $u=(-2,-1,0,1,2)/\sqrt{10}$ in hotel identifier order, and fix prices and the absolute ceiling. The model-based estimators fit a single MNL model to the simulated choices. The 20k/5k cohort uses the Lipschitz certificate and reuses correct-model Full and Restricted records. The independent 5k/20k cohort uses the box certificate and shared estimated costs. Direct uses 5,000 observations on each target pair and the anchor singleton, with no pilot. Table~\ref{tab:small-pilot-endpoints} reports the independent correct-model comparisons. Table~\ref{tab:mixture-diagnostics} measures errors and costs at true mixture revenues. The pilot budget guarantees use the single-MNL revenue mapping. Appendix~\ref{sec:extensions} gives Hoeffding bounds for Direct under stationary choice laws.

\begin{table}[!htbp]\SingleSpacedXI\centering\small
\caption{Independent 5k/20k validation under correct MNL specification. Worst-target MAE reductions for Full relative to each comparator, averaged across the four markets (percent). All methods share the box certificate and estimated costs. Actual costs can differ across methods. Each market uses 500 fresh replications.}
\label{tab:small-pilot-endpoints}
\begin{tabular}{lrrrr}\toprule
Comparator & Post gain & 95\% interval & All gain & 95\% interval\\\midrule
Restricted & 30.25 & [13.36, 43.68] & 18.98 & [0.12, 34.28]\\
Optimized pool & 14.74 & [-5.30, 30.84] & 12.05 & [-8.26, 28.49]\\
D-optimal & 7.10 & [-14.93, 24.76] & 5.99 & [-15.61, 23.54]\\
\bottomrule\end{tabular}\end{table}

\begin{table}[!htbp]\SingleSpacedXI\centering\small
\setlength{\tabcolsep}{1.7pt}
\caption{Misspecification at 25,000 total offers. In each market, $E_\infty$ takes the maximum across targets of the mean absolute error over 500 replications. Bias is the largest absolute target mean error. The table averages $E_\infty$, bias, and total opportunity cost over the four markets. Coverage is the minimum across markets of empirical simultaneous Bonferroni Wald coverage. Full refits share the executed schedule and cost.}
\label{tab:mixture-diagnostics}
\begin{tabular}{rrrrrrrrr}\toprule
 & \multicolumn{2}{c}{Full post-pilot} & \multicolumn{2}{c}{Full all-planned} & \multicolumn{2}{c}{Direct} & \multicolumn{2}{c}{Total cost}\\
\cmidrule(lr){2-3}\cmidrule(lr){4-5}\cmidrule(lr){6-7}\cmidrule(l){8-9}
$\delta$ & MAE / bias$^a$ & Coverage (\%) & MAE / bias$^a$ & Coverage (\%) & MAE / bias$^a$ & Coverage (\%) & Full & Direct\\\midrule
\multicolumn{9}{l}{\textit{Panel A. 20k/5k, Lipschitz certificate}}\\
0 & 2.665 / 0.937 & 95.6 & 1.803 / 0.338 & 94.6 & 4.570 / 0.424 & 95.0 & 2,357.16 & 1,528.43\\
0.5 & 2.771 / 1.099 & 95.8 & 1.833 / 0.460 & 92.8 & 4.732 / 0.380 & 94.6 & 2,372.63 & 1,546.08\\
1.5 & 4.371 / 3.834 & 84.0 & 3.032 / 2.614 & 56.0 & 5.091 / 0.361 & 94.0 & 2,655.79 & 1,852.29\\
\midrule
\multicolumn{9}{l}{\textit{Panel B. 5k/20k, box certificate}}\\
0 & 1.364 / 0.327 & 93.4 & 1.314 / 0.215 & 94.4 & 4.890 / 0.322 & 95.4 & 941.30 & 1,528.43\\
0.5 & 1.415 / 0.577 & 94.0 & 1.324 / 0.487 & 93.4 & 4.730 / 0.303 & 94.6 & 929.18 & 1,546.08\\
1.5 & 3.647 / 3.584 & 45.4 & 3.408 / 3.329 & 42.6 & 4.902 / 0.330 & 94.8 & 973.16 & 1,852.29\\
\bottomrule
\multicolumn{9}{l}{$^a$ MAE and absolute bias are multiplied by $10^3$.}\\
\end{tabular}\end{table}

\FloatBarrier
\subsection{Finite-Family Computational Diagnostic}\label{app:scale-results}
Five fixed seeds at each $N=5,10,15,20$ use $K=3$, three targets, reference share $0.05$ and ceiling $0.5\eta_{\rm ref}$. Each solve runs in a fresh process with one numerical-library thread. All twenty solves satisfy their reported gap certificates. For 1,350 offers, the median solve and certification time is 10.06 seconds, and the largest relative gap is $4.4\times10^{-10}$. Full timing, process memory, and support sizes accompany the supplementary checks.

\subsection{Assortment Selection after the Experiment}\label{app:selection-loss}
We select the offer with the largest post-pilot estimated gain among the anchor singleton and its four pair extensions, with zero for the singleton. Loss is the true expected revenue gap between the best and selected candidates per future offer. In Table~\ref{tab:selection-loss}, Full lowers loss relative to Restricted under the common ceiling and at matched cost. The Full--Restricted interval includes zero in the independent cohort. All three Full--D-optimal intervals include zero.

\begin{table}[!htbp]\SingleSpacedXI\centering\small
\setlength{\tabcolsep}{4pt}
\caption{Selection loss differences (comparator minus Full) with paired 95\% intervals, scaled by $10^3$. Intervals are exploratory and unadjusted across comparisons. We use equal market weights and 20,000 within-market pair resamples. Offline true-cost matching has 1,998 Restricted pairs. Other cells have 2,000.}
\label{tab:selection-loss}
\begin{tabular}{lrrr}\toprule
Comparator & Common ceiling & Matched cost & Independent 5k/20k\\\midrule
Restricted & $0.061\ [0.015,\ 0.107]$ & $0.430\ [0.315,\ 0.551]$ & $0.013\ [-0.013,\ 0.039]$\\
Equal pool & $0.057\ [0.006,\ 0.110]$ & $0.058\ [0.010,\ 0.107]$ & ---\\
Optimized pool & $0.034\ [-0.012,\ 0.079]$ & $0.059\ [0.010,\ 0.108]$ & $0.024\ [-0.003,\ 0.050]$\\
D-optimal & $-0.028\ [-0.070,\ 0.012]$ & $-0.010\ [-0.052,\ 0.030]$ & $-0.005\ [-0.031,\ 0.022]$\\
\bottomrule\end{tabular}\end{table}

\FloatBarrier
\endgroup

\begingroup
\setlength{\abovedisplayskip}{6pt plus 2pt minus 2pt}
\setlength{\belowdisplayskip}{6pt plus 2pt minus 2pt}
\setlength{\abovedisplayshortskip}{0pt plus 2pt}
\setlength{\belowdisplayshortskip}{4pt plus 2pt minus 1pt}
\section{Predictable-Design Concentration and Proofs}
\label{app:predictable-design-proofs}

We prove Theorem~\ref{thm:main} by first controlling the accumulated score
and then bounding the parameter estimation error. We apply these bounds
to the planned observations to obtain the contrast guarantee.

For the score bound, set $W_t:=\lambda I_d+\sum_{s=1}^t\Sigma_s$,
$t\in[T]$.

\subsection{Self-Normalized Concentration}\label{app:selfnorm}
\begin{lemma}[Self-normalized concentration]\label{lem:selfnorm}
Under Assumptions~\ref{ass:bounded} and~\ref{ass:mgf}, for
$\lambda>0$ and any $\delta\in(0,1)$, with probability at least
$1-\delta$,
\[
\left\|\sum_{s=1}^t\xi_s\right\|_{W_t^{-1}}
\le
\sqrt{2\log\!\left(
\frac{\det(W_t)^{1/2}}
{\det(\lambda I_d)^{1/2}\delta}
\right)}
\qquad\text{for every }t\in[T].
\]
\end{lemma}

\begin{proof}
Put $\Xi_t=\sum_{s=1}^t\xi_s$ and
$\Sigma_{1:t}=\sum_{s=1}^t\Sigma_s$. For fixed $u\in\R^d$,
\[
\mathcal M_t(u)=
\exp\!\left\{u^\top\Xi_t-\tfrac12u^\top\Sigma_{1:t}u\right\}
\]
is a nonnegative supermartingale by Assumption~\ref{ass:mgf}.
Mixing over an independent
$u\sim\mathcal N(0,\lambda^{-1}I_d)$ preserves this property, and
the Gaussian integral equals
\[
\widetilde{\mathcal M}_t=
\frac{\det(\lambda I_d)^{1/2}}{\det(W_t)^{1/2}}
\exp\!\left\{\tfrac12\norm{\Xi_t}_{W_t^{-1}}^2\right\}.
\]
Ville's inequality bounds
$\sup_{t\le T}\widetilde{\mathcal M}_t$ by $1/\delta$ with
probability at least $1-\delta$. Rearranging proves the result
simultaneously in $t$.
\end{proof}

\subsection{Proof of the Planned-Sample Guarantee}\label{app:mainproof}
\begin{proof}[Proof of Theorem~\ref{thm:main}]
Put $L_T^{\rm exp}(\theta)=\lambda\|\theta\|_2^2/2+
\sum_{t=1}^T\chi_t\ell_t(\theta)$ and
$e_T=\thetatilde_T-\thetastar$. Predictability of $\chi_t\in\{0,1\}$
gives the masked score $\chi_t\xi_t$ the moment-generating-function proxy $\chi_t\Sigma_t$,
while the masked losses have curvature proxy $\chi_t\Gamma_t$.
Convex optimality, including boundary solutions, gives
\begin{equation}
\ip{\nabla L_T^{\rm exp}(\thetatilde_T)}{-e_T}\ge0.
\label{eq:fooc_appendix}
\end{equation}
The segment from $\thetastar$ to $\thetatilde_T$ lies in $\Theta$.
Its integrated Hessian
$\bar H_T=\int_0^1\nabla^2L_T^{\rm exp}(\thetastar+\tau e_T)\,d\tau$
satisfies $\bar H_T\succeq V_T^{\rm exp}$, so Taylor expansion gives
\begin{equation}
\|e_T\|_{V_T^{\rm exp}}^2
\le e_T^\top\bar H_Te_T
\le-\ip{\nabla L_T^{\rm exp}(\thetastar)}{e_T}.
\label{eq:basic_contain}
\end{equation}
Since $\nabla L_T^{\rm exp}(\thetastar)=\lambda\thetastar-\sum_t\chi_t\xi_t$
and $V_T^{\rm exp}\succeq\lambda I_d$, Cauchy--Schwarz yields
\[
\|e_T\|_{V_T^{\rm exp}}
\le\sqrt\lambda B_\Theta+
\left\|\sum_t\chi_t\xi_t\right\|_{(V_T^{\rm exp})^{-1}}.
\]
Apply Lemma~\ref{lem:selfnorm} to the masked score with failure probability
$\delta/2$. Since $V_T^{\rm exp}\succeq\rho W_T^{\rm exp}$, this bounds
the right-hand side by $\beta_T^{\rm exp}(\delta/2)$. Finally,
Assumption~\ref{ass:lipschitz} gives \eqref{eq:contrast_general} for every
target pair on the same event.
\end{proof}

\section{Additional Results and Proofs for the MNL Specialization}\label{app:mnlproofs}

\subsection{Proof of Proposition~\ref{prop:mnlproxies}}\label{app:proof-mnlproxies}
The conditionally centered MNL score is supported on $S_t$ and obeys
$\abs{u^\top\xi_t}\le\sum_{i\in S_t}\abs{u_i}
\le\sqrt K\norm{D_t^{1/2}u}_2$. Hoeffding's lemma thus gives
$\E[\exp(u^\top\xi_t)\mid\Fcal_{t-1}^+]
\le\exp\{K u^\top D_tu/2\}$, as required for $\Sigma_t=KD_t$.

For the curvature bound, write $p_i=p_\theta(i\mid S)$ and
$p_0=p_\theta(0\mid S)$. Weighted
Cauchy--Schwarz yields
$u^\top\{\diag(p_S)-p_Sp_S^\top\}u
\ge p_0\sum_{i\in S}p_i u_i^2$.
Since $p_0\ge(1+Kv_{\max})^{-1}$ and
$p_i\ge v_{\min}(1+Kv_{\max})^{-1}$, the Fisher information satisfies
$\mathcal I_S(\theta)\succeq m_{\rm MNL}D_S$, where
$D_S:=\Diag(\one{i\in S})_{i=1}^N$ and
$m_{\rm MNL}=v_{\min}/(1+Kv_{\max})^2$. Hence
$\nabla^2\ell_t(\theta)\succeq\Gamma_t=m_{\rm MNL}D_t
\succeq(m_{\rm MNL}/K)\Sigma_t$.

For the revenue Lipschitz bound, write
$R_v(S):=\sum_{j\in S}r_jv_j/(1+\sum_{j\in S}v_j)$. For $i\in S$, the derivative
$\partial R_v(S)/\partial v_i
=\{r_i-R_v(S)\}/\{1+\sum_{j\in S}v_j\}$ has magnitude at most
one. Summing over at most $K$ active coordinates and using
$\abs{e^x-e^y}\le v_{\max}\abs{x-y}$ on $\Theta$ gives
$\abs{R_\theta(S)-R_{\theta'}(S)}
\le\sqrt K v_{\max}\norm{\theta-\theta'}_2$.\Halmos

\subsection{Balanced-Singleton Auxiliary Results}

\begin{remark}[Singleton feasibility and a cover-design alternative]\label{rem:coverdesign}
For the explicit rate theorem, assume that every singleton assortment is
feasible. More generally, a finite feasible cover
$S^{(1)},\dots,S^{(L)}$ with positive weights $q_1,\dots,q_L$ summing to one and
$\sum_{\ell=1}^Lq_\ell D_{S^{(\ell)}}\succeq\kappa_{\rm cov}I_N$,
$\kappa_{\rm cov}>0$, gives the same $T$-dependence: rounding $n_Tq_\ell$
with error at most one offer per assortment yields
\[
\sum_{t\in I_T} \Gamma_t \succeq m_{\rm MNL}\bigl(\kappa_{\rm cov} n_T - O(1)\bigr) I_N.
\]
The fixed cover affects constants. Theorem~\ref{thm:mnloverlay} uses balanced singletons.
\end{remark}

\begin{lemma}[Coordinatewise inference from balanced singleton rounds]
\label{lem:singleton-coordinatewise}
Let
\[
n_{\min,T}:=\left\lfloor\frac{n_T}{N}\right\rfloor,\qquad
b_\theta:=\max\{|\log v_{\min}|,|\log v_{\max}|\},
\qquad
m_{\rm sing}:=\frac{v_{\min}}{(1+v_{\max})^2}.
\]
Here $n_{\min,T}$ is the minimum number of singleton offers per product.
The constant $b_\theta$ bounds the absolute log attraction, and
$m_{\rm sing}$ is a lower bound on singleton likelihood curvature.
Under Assumption~\ref{ass:mnlbounded}, the planned-sample regularized
MLE computed from the balanced singleton rounds satisfies, with probability
at least $1-\delta$,
\begin{equation}
\norm{\thetatilde_T-\thetastar}_\infty
\le
\frac{
\lambda b_\theta+
\sqrt{\frac{n_{\min,T}+1}{2}\log\frac{2N}{\delta}}
}{
\lambda+m_{\rm sing}n_{\min,T}
}.
\label{eq:coordinatewise-singleton}
\end{equation}
Consequently,
\begin{equation}
\max_{(a,b)\in\Kcal}
\abs{\widehat\Delta_T^{(a,b)}-\Delta_R^{(a,b)}}
\le
2
\frac{
\lambda b_\theta+
\sqrt{\frac{n_{\min,T}+1}{2}\log\frac{2N}{\delta}}
}{
\lambda+m_{\rm sing}n_{\min,T}
}.
\label{eq:coordinatewise-contrast}
\end{equation}
In particular, if $n_T\ge 2N$, the right-hand side is
\[
\Otilde\left(
\sqrt{\frac{N}{n_T}}+\frac{N}{n_T}
\right)
=
\Otilde\left(\sqrt{\frac{N}{n_T}}\right).
\]
\end{lemma}

\paragraph{Proof.}\label{app:proof-singleton-coordinatewise}
Let $n_i$ be the number of planned offers of singleton $\{i\}$.
Balanced cycling gives $n_i\in\{n_{\min,T},n_{\min,T}+1\}$. The planned-sample regularized objective
separates by coordinate and has $i$th-coordinate curvature at least
$\lambda+m_{\rm sing}n_i$. The constrained-optimality argument in
Appendix~\ref{app:mainproof}, Hoeffding's inequality, and a union bound
give, simultaneously over $i$,
\[
(\lambda+m_{\rm sing}n_i)
\abs{\widetilde\theta_{T,i}-\theta_i^\star}
\le\lambda b_\theta+
\sqrt{\frac{n_i}{2}\log\frac{2N}{\delta}}
\]
with probability $1-\delta$. Using
$n_{\min,T}\le n_i\le n_{\min,T}+1$ proves
\eqref{eq:coordinatewise-singleton}.

To bound the contrast error, use the revenue gradient. For $i\in S$,
$\partial R_\theta(S)/\partial\theta_i
=p_\theta(i\mid S)\{r_i-R_\theta(S)\}$, and the derivative is zero
off $S$. Since $r_i,R_\theta(S)\in[0,1]$,
\[
\norm{\nabla_\theta R_\theta(S)}_1
=\sum_{i\in S}p_\theta(i\mid S)\abs{r_i-R_\theta(S)}
\le\sum_{i\in S}p_\theta(i\mid S)\le1.
\]
Convexity of $\Theta$, the mean-value theorem, and
$\ell_1$--$\ell_\infty$ duality therefore imply
$\abs{R_\theta(S)-R_{\theta'}(S)}\le\norm{\theta-\theta'}_\infty$.
Applying this bound to both assortments proves
\eqref{eq:coordinatewise-contrast}.\Halmos

\subsection{Proof of Theorem~\ref{thm:mnloverlay}}\label{app:proof-mnloverlay}
\begin{proof}
MNL-UCB gives
$\sup_{\theta\in\Theta_0}\E_\theta[\Reg_{\Bcal}(m,\theta)]
\le c_1\sqrt{Nm\log(Nm)}+c_2N\log^2(Nm)$ for absolute constants,
while regret is also at most $m$. If $N\le m$, then
$\log(Nm)\le2\log(m+1)$. If $N>m$, $m<N$ is absorbed by
$N\log^2(m+1)$. These two cases establish
\eqref{eq:base-mnl-regret}, and Theorem~\ref{thm:overlay} then gives
\eqref{eq:mnl-overlay-regret}.

The inference statement follows from
Lemma~\ref{lem:singleton-coordinatewise} when $n_T\ge2N$. If
$1\le n_T<2N$, bounded contrasts give an error of at most
$2\le2\sqrt{2}\sqrt{N/n_T}$, which yields the same rate.
\end{proof}

\paragraph{Polynomial schedules.}
\begin{corollary}[Polynomial overlay family]
\label{cor:alphafamily}
Under the assumptions of Theorem~\ref{thm:mnloverlay}, let $N$ be fixed
and let $n_T\asymp T^\alpha$ for $\alpha\in(0,1)$. Here
$\mathcal R_T$ and $\mathcal E_T$ denote the worst-case expected regret
and contrast-error criteria defined in
Section~\ref{subsec:rate-criteria}. Then
\[
\mathcal R_T
=
\Otilde\!\left(T^{\max\{1/2,\alpha\}}\right),
\qquad
\mathcal E_T
=
\Otilde(T^{-\alpha/2}).
\]
\end{corollary}

\begin{proof}
Substitute $n_T\asymp T^\alpha$ into
Theorem~\ref{thm:mnloverlay} and apply
Proposition~\ref{prop:hp2exp}.
\end{proof}

\section{Additional Results and Proofs for the Minimax Frontier}\label{app:lowerproof}
We prove Theorem~\ref{thm:lowerbound} on the hard subclass
$\Ecal_0=\{\phi_f:0\le f\le1/8\}$ from Section~\ref{sec:tradeoff}.
Fix a policy $\pi$ and a gap estimator $\widehat\Delta$. Under
$\phi_f$, the two singleton rewards have means $(3/4,1/4+2f)$. Let $P_f$
be the adaptive-trajectory law and set
$R_0=\E_{\phi_0}[\Reg_\pi(T,\phi_0)]
=\E_{\phi_0}[T_2(T)]/2$.
Choose the comparison instance by setting
$f_\pi=\{16\sqrt{1+R_0}\}^{-1}$. Since $f_\pi\le1/16$,
$\phi_{f_\pi}$ belongs to $\Ecal_0$.

The two contrasts are
$\Delta_0=1/2$ and $\Delta_{f_\pi}=1/2-2f_\pi$. Thresholding
$\widehat\Delta$ at their midpoint $m_\pi=1/2-f_\pi$ defines a test $\psi$
between $P_0$ and $P_{f_\pi}$. A testing error requires an estimation error
of at least $f_\pi$, so Bretagnolle--Huber gives
\begin{align*}
&\E_{\phi_0}\abs{\widehat\Delta-\Delta_0}
+\E_{\phi_{f_\pi}}\abs{\widehat\Delta-\Delta_{f_\pi}}\notag\\*
&\qquad\ge f_\pi\{P_0(\psi\ne0)+P_{f_\pi}(\psi\ne f_\pi)\}
\ge \frac{f_\pi}{2}\exp\{-\KL(P_0,P_{f_\pi})\}.
\end{align*}
Only the choice law of arm 2 changes. Write $\mathrm{kl}(p,q)$ for
the KL divergence between Bernoulli distributions with means $p$ and $q$.
The adaptive KL chain rule and
$\mathrm{kl}(p,q)\le(p-q)^2/\{q(1-q)\}$ yield
\[
\KL(P_0,P_{f_\pi})
=2R_0\,\mathrm{kl}\!\left(\frac14,\frac14+2f_\pi\right)
\le64R_0f_\pi^2\le\frac14.
\]

Thus one of the two expected absolute errors is at least
$e^{-1/4}f_\pi/4=e^{-1/4}/\{64\sqrt{1+R_0}\}$. Taking the supremum over
$\Ecal_0$ and using $\mathcal R_T^{\Ecal_0}(\pi)\ge R_0$ proves
$\mathcal E_T^{\Ecal_0}(\widehat\Delta)
\sqrt{1+\mathcal R_T^{\Ecal_0}(\pi)}\ge e^{-1/4}/64$.

\subsection{Capacity-One Upper Bound}\label{app:capacity-one-upper}
   
Define the capacity-one classes $\mathcal V_{\rm cap}:=[1/3,3]^2$
and $\Theta_{\rm cap}:=[\log(1/3),\log 3]^2$.
\begin{corollary}[Capacity-one overlay bound]
\label{cor:capacity-one-overlay}
Consider the two-product capacity-one problem
$\St_0=\{\{1\},\{2\}\}$, $r_1=r_2=1$, and parameter class
$\Theta_{\rm cap}$. Assume $2\le n_T\le T$. Let the base policy on the remaining rounds be any two-armed minimax policy satisfying
\begin{equation}
\sup_{\theta\in\Theta_{\rm cap}}
B_{\Bcal}(m,\theta)\le C_{\rm cap}\sqrt m
\label{eq:capacity-one-base-bound}
\end{equation}
for a universal constant $C_{\rm cap}$. Standard minimax stochastic-bandit
policies provide such a guarantee, as summarized by
\citet{Agrawal2019}. Run Algorithm~\ref{alg:overlay} with the balanced
singleton design and estimate the revenue gap
$\Delta_\theta:=R_\theta(\{1\})-R_\theta(\{2\})$ using the plug-in
estimate $\widehat\Delta_T$ from the planned-sample regularized MLE in
\eqref{eq:estimators}, fitted over $\Theta_{\rm cap}$ with fixed
$\lambda>0$.
Then
\begin{align*}
\sup_{\theta\in\Theta_{\rm cap}}
\E_\theta[\Reg_{\rm overlay}(T,\theta)]
&\le C_{\rm cap}\sqrt{T-n_T}+n_T,
\\
\sup_{\theta\in\Theta_{\rm cap}}
\E_\theta|\widehat\Delta_T-\Delta_\theta|
&=\Otilde(n_T^{-1/2}).
\end{align*}
For $n_T\asymp T^\alpha$, these bounds give
$\mathcal R_T=\Otilde(T^{\max\{1/2,\alpha\}})$ and
$\mathcal E_T=\Otilde(T^{-\alpha/2})$ on the same fixed class.
\end{corollary}

\begin{proof}
Each singleton is a Bernoulli arm with
$\mu_i=v_i/(1+v_i)\in[1/4,3/4]$, so
Theorem~\ref{thm:overlay} and \eqref{eq:capacity-one-base-bound} give
the regret bound.

Balanced cycling supplies $\lfloor n_T/2\rfloor$ or
$\lceil n_T/2\rceil$ samples per arm. Apply
Lemma~\ref{lem:singleton-coordinatewise} with $N=2$,
$v_{\min}=1/3$, and $v_{\max}=3$ to obtain a uniform
$\Otilde(n_T^{-1/2})$ gap error. The plug-in gap lies in $[-1,1]$,
so Proposition~\ref{prop:hp2exp} gives the expected bound.
\end{proof}
   
\subsection{Proof of Corollary~\ref{cor:frontier}}\label{app:proof-frontier}
\begin{proof}
The class $\Theta_0$ contains two-armed Bernoulli instances with
arbitrarily close singleton revenues. The usual minimax regret lower bound \citep[Chapter~15]{LattimoreSzepesvari2020}
on these instances gives the first inequality. Since $\Theta_0$ also
contains $\Ecal_0$, Theorem~\ref{thm:lowerbound} implies
$1+\mathcal R_T\ge c^2/\mathcal E_T^2$, which gives the second inequality.

For the upper bound, $\Theta_{\rm cap}=\Theta_0$, so
Corollary~\ref{cor:capacity-one-overlay} gives
$\mathcal R_T=\Otilde(\sqrt T+n_T)$ and
$\mathcal E_T=\Otilde(n_T^{-1/2})$ on this benchmark class.
   
\end{proof}
\subsection{Auxiliary Frontier Results}

\begin{lemma}[Design-sample precision lower bound]
\label{lem:design-sample-lower}
For the balanced-singleton overlay on the fixed capacity-one problem,
any estimator $\widehat\Delta$ measurable with respect to the planned-design
sample only satisfies
$\mathcal E_T^{\Ecal_0}(\widehat\Delta)\ge c_0/\sqrt{1+n_T}$
for a universal $c_0>0$.
\end{lemma}

\paragraph{Proof.}\label{app:proof-design-sample-lower}
Let $m_2$ be the number of planned item-$2$ offers, so
$m_2\le\lceil n_T/2\rceil$, and set
$h=1/\{16\sqrt{1+m_2}\}$, $f_1=h/2$.
The instances $\phi_0,\phi_{f_1}\in\Ecal_0$ differ only in the
item-$2$ mean, from $1/4$ to $1/4+h$, and their targets differ by $h$.
Let $P_0^{\rm des}$ and $P_1^{\rm des}$ denote the corresponding
planned-sample laws under $\phi_0$ and $\phi_{f_1}$.
Since $q(1-q)\ge1/8$ on $[1/4,5/16]$,
\[
\KL(P_0^{\rm des},P_1^{\rm des})
=m_2\,\mathrm{kl}(1/4,1/4+h)
\le8m_2h^2\le1/32.
\]
The midpoint test and Bretagnolle--Huber inequality make the sum of
the two mean absolute errors at least $he^{-1/32}/4$. Thus their
maximum is at least
$he^{-1/32}/8\ge e^{-1/32}/\{128\sqrt{1+n_T}\}$, proving the
claim with $c_0=e^{-1/32}/128$.\Halmos

\begin{corollary}[Tightness along the non-dominated overlay family]
\label{cor:overlay-tightness}
   
For every $\alpha\in[1/2,1)$, choose $n_T\asymp T^\alpha$ in the
capacity-one overlay of Corollary~\ref{cor:capacity-one-overlay}.
Together with Corollary~\ref{cor:frontier}, this policy--estimator pair
satisfies, up to logarithmic factors,
\begin{equation}
\mathcal R_T=\widetilde{\Theta}(T^\alpha),
\qquad
\mathcal E_T=\widetilde{\Theta}(T^{-\alpha/2}).
\label{eq:overlay-tightness}
\end{equation}
\end{corollary}

\begin{proof}\label{app:proof-overlay-tightness}
For $\alpha\ge1/2$, Corollary~\ref{cor:capacity-one-overlay} gives
$\mathcal E_T=\Otilde(T^{-\alpha/2})$ and
$\mathcal R_T=\Otilde(T^\alpha)$. The first bound and
Corollary~\ref{cor:frontier} imply $\mathcal R_T=\Omegatilde(T^\alpha)$.
The second and Theorem~\ref{thm:lowerbound} imply
$\mathcal E_T=\Omegatilde(T^{-\alpha/2})$, proving \eqref{eq:overlay-tightness}.
\end{proof}
\begin{proposition}[Rate-wise non-dominated interval inside the
polynomial family]
\label{prop:nondominated_interval}
   
Within the capacity-one overlay family of
Corollary~\ref{cor:capacity-one-overlay}, with its planned-sample gap
estimator and $n_T\asymp T^\alpha$, $\alpha\in(0,1)$, compare polynomial orders.
For $\alpha<1/2$, increasing $\alpha$ improves inference without changing
the regret exponent. For $1/2\le\alpha<\alpha'<1$, $\alpha$ has the better
regret exponent and $\alpha'$ the better inference exponent.
Thus $[1/2,1)$ is a continuum of pairwise rate-wise non-dominated points.
\end{proposition}

\begin{proof}\label{app:proof-nondominated}
Corollary~\ref{cor:capacity-one-overlay} and
Lemma~\ref{lem:design-sample-lower} give error exponent
$-\alpha/2$. For $\alpha\le1/2$, the two-armed regret lower bound
and overlay upper bound give regret exponent $1/2$, so
$\alpha<1/2$ is dominated by $1/2$. For $\alpha\ge1/2$,
Corollary~\ref{cor:overlay-tightness} gives exponents $\alpha$ and
$-\alpha/2$, which move in opposite directions.
\end{proof}

\section{Model-Free Simultaneous Contrast Inference}\label{sec:extensions}\label{subsec:robust-inference}\label{app:proof-modelfree}

Let $\mathcal A_{\Kcal}$ be the distinct assortments appearing in $\Kcal$ and $J_{\Kcal}=|\mathcal A_{\Kcal}|$. Allocate $n_S\ge1$ planned observations to each $S\in\mathcal A_{\Kcal}$, with counts fixed before their outcomes, possibly using an initial pilot that excludes those outcomes. Let $\widehat R_{\rm des}(S)$ be their sample mean of realized revenues $Z_t=r_{Y_t}\in[0,1]$. Under a stationary choice law $p^\star$, write $R(S):=R^\star(S)=\sum_{i\in S}r_i p^\star(i\mid S)$.

\begin{theorem}[Model-free simultaneous contrast inference]
\label{thm:modelfree}
Under \eqref{eq:model}, with probability at least $1-\delta$,
simultaneously for every $S\in\mathcal A_{\Kcal}$ and
$k=(a,b)\in\Kcal$,
\begin{align}
\left|
\widehat R_{\rm des}(S)-R(S)
\right|
&\le
\sqrt{\frac{\log(2J_{\Kcal}/\delta)}{2n_S}},
\label{eq:modelfree-mean}\\
\left|
\{\widehat R_{\rm des}(S_a)-\widehat R_{\rm des}(S_b)\}
-\{R(S_a)-R(S_b)\}
\right|
&\le
\sqrt{\frac{\log(2J_{\Kcal}/\delta)}{2n_{S_a}}}
+
\sqrt{\frac{\log(2J_{\Kcal}/\delta)}{2n_{S_b}}}.\notag
\end{align}
\end{theorem}

\begin{proof}
Condition on any initial pilot used to select the fixed counts. Fix
$S\in\mathcal A_{\Kcal}$ and enumerate its planned rounds by
$\tau_{S,1},\ldots,\tau_{S,n_S}$. Equation~\eqref{eq:model} makes
$Z_{\tau_{S,k}}-R(S)$ a martingale-difference sequence with range
length one. Hence
\[
\Prob\!\left(
\abs{\widehat R_{\rm des}(S)-R(S)}>\varepsilon
\right)
\le2e^{-2n_S\varepsilon^2}.
\]
Set
$\varepsilon=\sqrt{\log(2J_{\Kcal}/\delta)/(2n_S)}$ and take a
union bound over the $J_{\Kcal}$ design assortments to obtain
\eqref{eq:modelfree-mean}. The contrast bound follows from the triangle
inequality.
\end{proof}

Direct estimation allocates planned traffic to target assortments. Under a correctly specified MNL model, observations of other assortments also inform the contrasts through shared product parameters; the SDP uses this information when choosing the allocation.

\par\endgroup

\end{document}

%% file: figures/expedia_preference_learning_tikz.tex
\begin{tikzpicture}[
  x=1cm,y=1cm,
  font=\fontsize{9}{11}\selectfont,
  line width=0.55pt,
  flow/.style={-{Latex[length=2mm]},draw=ORGray,line width=0.7pt},
  itemA/.style={draw=ORAccent,fill=ORAccentLight,minimum width=1.1cm,minimum height=0.39cm,inner sep=1pt},
  itemB/.style={draw=ORGray,fill=ORLightGray,minimum width=1.1cm,minimum height=0.39cm,inner sep=1pt},
  outcome/.style={draw=ORGray,fill=white,rounded corners=1.5pt,minimum width=1.15cm,minimum height=0.48cm,inner sep=2pt},
  paneltitle/.style={font=\fontsize{9}{11}\selectfont\bfseries,text=ORAccent}
]
\foreach \x in {0,5.1,10.2}{
  \draw[draw=ORGray!65,fill=white,rounded corners=2pt] (\x,0) rectangle +(4.65,3.6);
}
\node[paneltitle] at (2.325,3.2) {1. Observe choices};
\node[paneltitle] at (7.425,3.2) {2. Compare preferences};
\node[paneltitle] at (12.525,3.2) {3. Assess uncertainty};

\node[itemA] at (1.15,2.55) {$A_1$};
\node[itemB] at (1.15,2.05) {$B_1$};
\node[itemA] (chosen) at (1.15,1.55) {$A_2$};
\node[itemB] at (1.15,1.05) {$B_2$};
\draw[flow] (chosen.east) -- (2.45,1.55);
\node[align=center,text width=1.7cm] at (3.25,1.8) {Clicks or\\bookings};
\node[align=center,text=ORGray] at (2.325,0.4) {Items from rankings $A$ and $B$};

\node[outcome,fill=ORAccentLight] at (6.025,2.4) {$A$ wins};
\node[outcome] at (7.425,2.4) {Tie};
\node[outcome,fill=ORLightGray] at (8.825,2.4) {$B$ wins};
\draw[flow] (7.425,2.05) -- (7.425,1.65);
\node[align=center,text width=3.5cm] at (7.425,1.2) {Estimate relative\\preference};
\node[align=center,text=ORGray] at (7.425,0.4) {Combine user-level outcomes};

\node at (12.525,2.55) {Confidence interval};
\draw[draw=ORGray] (10.75,1.45) -- (14.3,1.45);
\draw[draw=ORGray,densely dashed] (12.525,1.25) -- (12.525,2.2);
\draw[draw=ORAccent,line width=1.1pt] (11.75,1.9) -- (13.75,1.9);
\draw[draw=ORAccent,line width=1.1pt] (11.75,1.78) -- (11.75,2.02);
\draw[draw=ORAccent,line width=1.1pt] (13.75,1.78) -- (13.75,2.02);
\fill[ORAccent] (12.95,1.9) circle (2pt);
\node at (12.525,1.04) {$0$};
\node[align=center] at (11.4,0.97) {Favors $B$};
\node[align=center] at (13.65,0.97) {Favors $A$};
\node[align=center,text=ORGray] at (12.525,0.4) {Report comparison uncertainty};

\draw[flow] (4.65,1.8) -- (5.1,1.8);
\draw[flow] (9.75,1.8) -- (10.2,1.8);
\draw[flow] (12.525,0) -- (12.525,-0.4);
\draw[draw=ORGray!65,fill=ORLightGray,rounded corners=2pt] (0,-1.05) rectangle (14.85,-0.4);
\node[align=center] at (7.425,-0.725) {Use the evidence to screen candidates for A/B tests of business outcomes};
\end{tikzpicture}

%% file: figures/pilot_safe_pipeline_tikz.tex
\begingroup
\linespread{1}\selectfont
\definecolor{ORFigInk}{HTML}{252C34}
\definecolor{ORFigGray}{HTML}{59636C}
\definecolor{ORFigBlue}{HTML}{28576D}
\definecolor{ORFigGrayFill}{HTML}{F6F7F8}
\definecolor{ORFigBlueFill}{HTML}{F0F5F7}
\definecolor{ORFigDivider}{HTML}{DEE3E6}
\resizebox{0.98\linewidth}{!}{%
\begin{tikzpicture}[
  x=1cm,y=1cm,
  font=\fontsize{10}{12}\selectfont,
  text=ORFigInk,
  stage/.style={draw=ORFigGray,fill=ORFigGrayFill,
    rounded corners=2pt,line width=0.75pt,inner sep=0pt},
  plannedstage/.style={stage,draw=ORFigBlue,fill=ORFigBlueFill},
  flow/.style={-{Latex[length=2mm,width=1.4mm]},
    line width=0.8pt,draw=ORFigGray},
  plannedflow/.style={flow,draw=ORFigBlue},
  sub/.style={font=\fontsize{8.5}{10.5}\selectfont,text=ORFigBlue,align=center},
  note/.style={font=\fontsize{9}{10.5}\selectfont,text=ORFigGray},
  mathnote/.style={font=\fontsize{10}{12}\selectfont}
]
  \path[use as bounding box] (0,0.35) rectangle (17.1,4.98);

  \node[note] at (8.6,4.78)
    {Interleaved rounds with separate observation histories};

  \node[anchor=west,align=left,font=\bfseries] at (0.05,3.36)
    {Base-policy\\rounds};
  \node[anchor=west,note,mathnote] at (0.05,2.67) {$T-n_T$ rounds};

  \node[stage,minimum width=3.33cm,minimum height=1.03cm] (select)
    at (4.665,3.335) {Select assortment};
  \node[stage,minimum width=3.33cm,minimum height=1.03cm] (observe)
    at (9.385,3.335) {Observe choice};
  \node[stage,minimum width=3.33cm,minimum height=1.03cm] (update)
    at (14.105,3.335) {Update base policy};
  \draw[flow] (select.east) -- (observe.west);
  \draw[flow] (observe.east) -- (update.west);
  \draw[flow] (update.north) -- (14.105,4.22) -- (4.665,4.22)
    -- (select.north);
  \node[note,fill=white,inner xsep=4pt,inner ysep=1pt] at (9.35,4.22)
    {next base-policy round};

  \draw[draw=ORFigDivider,line width=0.55pt] (0,2.48) -- (17.0,2.48);

  \node[anchor=west,align=left,font=\bfseries,text=ORFigBlue] at (0.05,1.52)
    {Planned\\rounds};
  \node[anchor=west,note,mathnote,text=ORFigBlue] at (0.05,0.88)
    {$n_T=p_T+m_T$};

  \foreach \name/\x in {pilot/4.6,deployment/11.8,inference/15.4}
    \node[plannedstage,minimum width=3.2cm,minimum height=1.55cm] (\name)
      at (\x,1.415) {};
  \node[plannedstage,minimum width=3.4cm,minimum height=1.55cm] (allocation)
    at (8.2,1.415) {};
  \node at (4.6,1.67) {Collect pilot};
  \node[sub,mathnote] at (4.6,1.03) {$p_T$ observations};
  \node at (8.2,1.67) {Select allocation};
  \node[sub] at (8.2,1.03) {Budget tightening\\Reference stabilization};
  \node at (11.8,1.67) {Round and deploy};
  \node[sub,mathnote] at (11.8,1.03) {$m_T$ post-pilot offers};
  \node at (15.4,1.67) {Estimate contrasts};
  \node[sub] at (15.4,1.03) {Post-pilot data};
  \draw[plannedflow] (pilot.east) -- (allocation.west);
  \draw[plannedflow] (allocation.east) -- (deployment.west);
  \draw[plannedflow] (deployment.east) -- (inference.west);
\end{tikzpicture}%
}
\endgroup

%% file: figures/free_costly_geometry_tikz.tex
\begingroup%
\definecolor{ORInk}{HTML}{30343B}%
\definecolor{ORWarm}{HTML}{B86F36}%
\begin{tikzpicture}[
  x=1cm,y=1cm,
  font=\fontsize{10}{12}\selectfont,
  text=ORInk,line cap=round,line join=round,
  target/.style={-{Stealth[length=2.1mm,width=1.4mm]},draw=ORAccent,line width=.9pt},
  component/.style={-{Stealth[length=2.0mm,width=1.3mm]},draw=ORWarm,line width=.85pt},
  plane/.style={fill=ORAccentLight,draw=ORAccent!48,line width=.6pt},
  condition/.style={align=center,font=\fontsize{9}{11}\selectfont,text width=6.9cm},
  sublabel/.style={align=center,font=\fontsize{9}{11}\selectfont,text=ORGray}
]
  \path[use as bounding box] (0,0) rectangle (14.8,5.15);
  \draw[ORGray!22,line width=.45pt] (7.40,.25) -- (7.40,4.95);
  \node[font=\fontsize{10}{12}\selectfont\bfseries] at (3.60,4.72)
    {(a) Free inference};
  \node[font=\fontsize{10}{12}\selectfont\bfseries] at (11.20,4.72)
    {(b) Costly inference};

  \path[plane] (.55,1.75) -- (5.70,1.75) -- (6.75,2.65) -- (1.60,2.65) -- cycle;
  \coordinate (free-origin) at (3.30,2.17);
  \draw[target] (free-origin) -- (5.72,2.40);
  \draw[target] (free-origin) -- (2.15,2.53);
  \fill[ORInk] (free-origin) circle[radius=1.4pt];
  \node[anchor=south,text=ORAccent] at (5.76,2.44) {$g_k(\theta)$};
  \node[anchor=south,text=ORAccent] at (2.05,2.58) {$g_j(\theta)$};
  \node[sublabel] at (3.60,1.18) {Zero-cost information span};
  \node[condition] at (3.60,.48)
    {$g_k(\theta)\in\operatorname{Range}\{M_{\mathcal O}(\theta)\}$\\[2pt]
     for every $k\in\mathcal K$};

  \path[plane] (8.15,1.75) -- (13.30,1.75) -- (14.35,2.65) -- (9.20,2.65) -- cycle;
  \coordinate (costly-origin) at (10.55,2.17);
  \coordinate (projection) at (12.50,2.17);
  \coordinate (target-tip) at (12.50,4.00);
  \draw[-{Stealth[length=1.8mm,width=1.15mm]},draw=ORGray!65,line width=.7pt]
    (costly-origin) -- (projection);
  \draw[component] (projection) -- (target-tip);
  \draw[target] (costly-origin) -- (target-tip);
  \draw[ORWarm!70,line width=.5pt]
    (12.31,2.17) -- (12.31,2.36) -- (12.50,2.36);
  \fill[ORInk] (costly-origin) circle[radius=1.4pt];
  \fill[ORGray!75] (projection) circle[radius=1.1pt];
  \node[anchor=south,text=ORAccent] at (12.50,4.06) {$g_k(\theta)$};
  \node[anchor=west,align=left,text=ORWarm,text width=1.75cm,
    font=\fontsize{9}{11}\selectfont] at (12.73,3.25)
    {Missing\\component};
  \node[anchor=north,sublabel,inner sep=0pt] at (11.55,2.11) {Projection};
  \node[sublabel] at (11.20,1.18) {Zero-cost information span};
  \node[condition] at (11.20,.48)
    {$g_k(\theta)\notin\operatorname{Range}\{M_{\mathcal O}(\theta)\}$\\[2pt]
     for some $k\in\mathcal K$};
\end{tikzpicture}%
\endgroup%

%% file: figures/minimax_frontier_tikz.tex
\begingroup
\begin{tikzpicture}[every node/.style={font=\fontsize{9}{11}\selectfont}]
\begin{axis}[
  width=12.3cm,
  height=6.6cm,
  xmin=0.42,
  xmax=1.025,
  ymin=0,
  ymax=0.565,
  axis lines=left,
  axis line style={ORGray, line width=0.45pt, -{Stealth[length=3pt]}},
  tick style={ORGray, line width=0.4pt},
  xlabel={Regret exponent $r$ (lower is better)},
  ylabel={Precision exponent $p$ (higher is better)},
  label style={font=\fontsize{9}{11}\selectfont},
  tick label style={font=\fontsize{9}{11}\selectfont},
  xtick={0.5,0.75,1.0},
  xticklabels={$1/2$,$3/4$,$1$},
  ytick={0,0.25,0.5},
  yticklabels={$0$,$1/4$,$1/2$},
  enlargelimits=false,
  clip=false,
]
  \addplot[draw=none, fill=black!5] coordinates {
    (0.42,0) (0.50,0) (0.50,0.25) (1,0.50)
    (1,0.565) (0.42,0.565)
  } \closedcycle;
  \node[anchor=center, align=center, text=ORGray]
    at (axis cs:0.69,0.535) {Excluded by the lower bounds};
  \addplot[black!24, line width=0.4pt, densely dotted]
    coordinates {(0.42,0.25) (0.50,0.25)};
  \addplot[black!24, line width=0.4pt, densely dotted]
    coordinates {(0.42,0.50) (1,0.50) (1,0)};
  \addplot[ORGray, line width=0.9pt, densely dashed]
    coordinates {(0.50,0) (0.50,0.25)};
  \node[anchor=west, align=left, text=ORGray]
    at (axis cs:0.535,0.085) {Same regret rate,\\lower precision};
  \draw[ORGray, line width=0.45pt]
    (axis cs:0.531,0.085) -- (axis cs:0.505,0.085);
  \addplot[ORAccent, line width=1.2pt]
    coordinates {(0.50,0.25) (1,0.50)};
  \addplot[only marks, mark=*, mark size=2.1pt,
    draw=ORAccent, fill=ORAccent] coordinates {(0.50,0.25)};
  \addplot[only marks, mark=*, mark size=2.4pt,
    draw=ORAccent, fill=white, line width=0.9pt]
    coordinates {(1,0.50)};
  \node[anchor=west, align=left, text=ORAccent]
    at (axis cs:0.705,0.235)
    {Nondominated frontier\\$r=2p,\quad 1/4\le p<1/2$};
  \draw[ORAccent, line width=0.5pt]
    (axis cs:0.77,0.29) -- (axis cs:0.80,0.393);
\end{axis}
\end{tikzpicture}
\endgroup

%% file: figures/oracle_frontier_schematic_tikz.tex
\begingroup
\definecolor{ORCostly}{HTML}{B86F36}
\begin{tikzpicture}[every node/.style={font=\fontsize{9}{11}\selectfont}]
\pgfplotsset{frontier panel/.style={
  width=7.0cm, height=5.3cm,
  xmin=0, xmax=1.04, ymin=0, ymax=1.35,
  axis lines=left,
  axis line style={ORGray, line width=0.45pt, -{Stealth[length=3pt]}},
  tick style={ORGray, line width=0.4pt},
  xlabel={Opportunity-cost budget $\eta$},
  ylabel={Variance constant $F_\theta(\eta)$},
  label style={font=\fontsize{9}{11}\selectfont},
  tick label style={font=\fontsize{9}{11}\selectfont},
  xtick={0,1}, xticklabels={$0$,$1$}, ytick=\empty,
  enlargelimits=false, clip=false,
}}
\begin{axis}[frontier panel, name=freepanel,
  title={\textbf{Free instance}},
  title style={font=\fontsize{10}{12}\selectfont, yshift=3pt},
]
  \addplot[ORAccent, line width=1.05pt, domain=0:1, samples=100]
    {0.22+0.45*exp(-4*x)};
  \addplot[only marks, mark=*, mark size=2pt,
    draw=ORAccent, fill=ORAccent] coordinates {(0,0.67)};
  \node[anchor=west, align=left, text=ORAccent]
    at (axis cs:0.18,1.04)
    {Finite at zero cost\\$F_\theta(0)<\infty$};
  \draw[ORAccent, line width=0.5pt, -{Stealth[length=3.4pt]}]
    (axis cs:0.23,0.85) -- (axis cs:0.015,0.695);
\end{axis}
\begin{axis}[frontier panel,
  at={(freepanel.south east)}, anchor=south west, xshift=1.0cm,
  title={\textbf{Costly instance}},
  title style={font=\fontsize{10}{12}\selectfont, yshift=3pt},
]
  \addplot[ORCostly, line width=1.05pt, domain=0.079:1, samples=150]
    {0.16+0.085/x};
  \node[anchor=west, text=ORCostly]
    at (axis cs:0.29,1.27) {$F_\theta(0)=+\infty$};
  \draw[ORCostly, line width=1.05pt, -{Stealth[length=4.1pt]}]
    (axis cs:0.089,1.115) -- (axis cs:0.079,1.24)
    -- (axis cs:0.073,1.32);
  \node[anchor=west, align=left, text=ORCostly]
    at (axis cs:0.29,1.03)
    {Inverse-budget growth\\$F_\theta(\eta)=\Theta(\eta^{-1})$ as $\eta\downarrow0$};
  \draw[ORCostly, line width=0.5pt, -{Stealth[length=3.4pt]}]
    (axis cs:0.32,0.84) -- (axis cs:0.14,0.79);
\end{axis}
\end{tikzpicture}
\endgroup